\pdfoutput=1
\def\arxivversion{}
\def\withsupplement{}
\documentclass[10pt,twocolumn,letterpaper]{article}

\usepackage[pagenumbers]{cvpr}

\usepackage{multirow}
\usepackage{tikz}
\usepackage{colortbl}
\newcommand{\xs}{\cellcolor{black!7}}  %
\newcommand{\xsswatch}{{\setlength{\fboxsep}{0pt}\setlength{\fboxrule}{0.4pt}\fcolorbox{black!45}{black!7}{\rule{0pt}{1.4ex}\hspace{1.1em}}}}  %

\usepackage[skins]{tcolorbox}
\definecolor{cvprblue}{rgb}{0.21,0.49,0.74}
\tcbset{cvprfinding/.style={enhanced,
  colback=cvprblue!4!white,colframe=cvprblue!78!black,
  boxrule=0.45pt,arc=1.2pt,left=4pt,right=4pt,top=6.5pt,bottom=2pt,
  before skip=8pt,after skip=6pt,
  fontupper=\normalsize,fonttitle=\normalsize\bfseries,coltitle=white,
  attach boxed title to top left={xshift=5pt,yshift=-7pt},
  boxed title style={colback=cvprblue!88!black,colframe=cvprblue!88!black,boxrule=0pt,arc=1pt,
    left=4pt,right=4pt,top=2pt,bottom=2pt}}}
\newtcolorbox{findingbox}[1]{cvprfinding,title={#1}}
\newtcolorbox{researchquestionbox}{cvprfinding,title=Research Question}
\newtcolorbox{keyfindingsbox}{cvprfinding,title=Key Findings}
\newtcolorbox{takeawaybox}{cvprfinding,title=Takeaway}

\ifdefined\withsupplement
  \newcommand{\suppref}[1]{\cref{#1}}
\else
  \newcommand{\suppref}[1]{the supplement}
\fi

\ifdefined\arxivversion
  \newcommand{\arxivor}[2]{#2}
\else
  \newcommand{\arxivor}[2]{#1}
\fi

\newcommand{\dl}{\Delta\ell}
\newcommand{\pos}[1]{\left[#1\right]_{+}}
\newcommand{\Hm}{\mathcal{H}}
\newcommand{\Gm}{\mathcal{G}}
\newcommand{\Ex}{\mathbb{E}}

\definecolor{cvprblue}{rgb}{0.21,0.49,0.74}
\usepackage[pagebackref,breaklinks,colorlinks,allcolors=cvprblue]{hyperref}

\def\confName{CVPR}
\def\confYear{2027}

\title{When Lower Harm Misleads:\\Auditing Concept Corrections}

\author{Tianyou Lai$^{1,*}$\qquad Wentao Yue$^{2,*}$\qquad Hongji Li$^{3}$\qquad Chaoyuan Hao$^{4}$\qquad Qilei Li$^{5}$\\[3pt]
{\normalsize $^{1}$Carnegie Mellon University\qquad $^{2}$The Hong Kong Polytechnic University}\\
{\normalsize $^{3}$Mohamed bin Zayed University of Artificial Intelligence}\\
{\normalsize $^{4}$University of Chinese Academy of Sciences\qquad $^{5}$Central China Normal University}\\[2pt]
{\small $^{*}$Equal contribution}
}

\begin{document}
\maketitle
\begin{abstract}
An expert who corrects a concept prediction, such as a BI-RADS descriptor in breast ultrasound, expects the diagnosis to improve.
Objectives that reduce the harm of such corrections are judged by harm and accuracy, yet harm also falls when a model responds less to every correction or starts from a worse uncorrected prediction, the reference for harm.
We propose a correction audit, a reproducible evaluation protocol that separates these explanations: it splits each correction into harm and useful gain on identical events, separates the starting prediction from the correction response, and compares objectives with amplitude baselines matched in source harm.
On three breast-ultrasound sources, worst-case correction training, evaluated from the mean-case model's start, lowers harm but gives up more useful gain in all 21 source comparisons pooled over six seeds, raising the corrected negative log-likelihood (NLL); from the same start, selecting between the objectives by harm picks the less helpful model on all three sources.
On CUB, a monotonic harm penalty at its reported weight lowers path-level harm in all 80 concept sets while raising the NLL in all 80, almost entirely before any correction.
At a fixed start, rescaling both responses to the same source harm reproduces the penalty's mean harm, useful gain and NLL on single-concept corrections within a prespecified margin of $0.0005$ nat, although the two responses point in different directions.
Lower harm therefore signals better corrections only when it is read together with useful gain, absolute risk and the start it is measured from.
\end{abstract}

\section{Introduction}
\label{sec:intro}

Medicine remembers ``\emph{do no harm}''; Hippocrates wrote ``\emph{to help, or at least to do no harm}''~\cite{hippocrates1923epidemics}.
Objectives that minimize the harm of concept corrections can keep the fallback and drop the help.
Concept-based models predict human-interpretable concepts and then a task label from them~\cite{koh2020concept,espinosa2022concept}.
Their distinctive use is test-time correction: in breast ultrasound, a radiologist who sees a mispredicted BI-RADS descriptor, such as margin or orientation, can set it right before the malignancy score is read.
Corrections do not always help: their effect depends on the training regime and intervention procedure~\cite{shin2023closer}, can vanish under distribution shift~\cite{espinosa2025leakage}, and correcting only some wrong concepts can leave a prediction worse than correcting none~\cite{verghese2026harmful}.

\begin{figure*}[t]
\centering
\includegraphics[width=\textwidth]{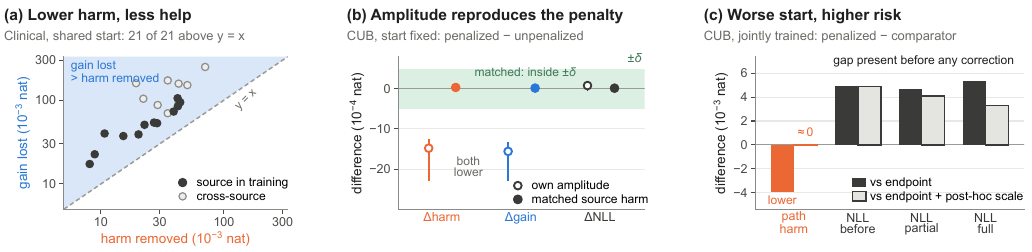}
\caption{Lower correction harm need not mean better corrections.
(a)~Clinical concept correction from the mean-case model's start, which compares correction responses rather than deployed models: worst-case correction training removes harm but loses more useful gain in all 21 comparisons, so corrected NLL rises (\cref{tab:ultrasound_gain,fig:clinical}).
(b,\,c)~CUB controls with a monotonic harm penalty.
(b)~Start fixed: at its own source-optimal amplitude the penalized model has lower harm and gain; a post-hoc amplitude that equalizes source harm brings all three mean differences inside the margin $\pm\delta$, $\delta=5\times10^{-4}$ nat (\cref{fig:contrasts}).
(c)~Jointly trained starts: a post-hoc scale gives the endpoint model the penalty's maximum sampled partial harm, and the penalty's NLL gap is present before any correction (\cref{sec:res_native,sec:res_native_frontier}).
Intervals: (a)~\cref{tab:ultrasound_gain}; (c)~\suppref{tab:native}. Units: $10^{-3}$ nat in (a) and (c), $10^{-4}$ nat in (b).}
\label{fig:teaser}
\end{figure*}

Training objectives therefore target the harm of corrections directly.
A monotonic intervention objective penalizes any increase of the task loss along partial correction paths and, trained jointly with an endpoint objective, sharply reduces path-level harm while keeping task accuracy~\cite{verghese2026harmful}.
Mean-case and worst-case correction objectives instead train on the task loss at the legal correction states of each image.
Lower harm invites the reading that such models are easier to correct and benefit more from it.

\begin{researchquestionbox}
When an objective lowers the harm of concept corrections, do corrected predictions improve, or does the model respond less and start from a worse prediction?
\end{researchquestionbox}

Harm can fall through two effects that improve no corrected prediction.
A model that responds weakly to every correction causes little harm but also discards the useful part of each correction, so harm alone rewards a model that ignores corrections.
And because harm is measured against the model's own uncorrected risk, a model that starts from worse predictions shows less harm without correcting any better.
Harm and accuracy alone reveal neither effect.

Our correction audit removes the two effects in turn: it splits each loss change on identical events into a harmful part $\Hm$ and a useful part $\Gm$, compares responses at a shared start or splits absolute risk at native starts into before- and after-correction parts, and compares objectives with amplitude baselines matched in source harm, which must reproduce the outcomes if amplitude is all an objective changes.

\Cref{fig:teaser} summarizes two lines of evidence.
(A)~In clinical correction, worst-case training lowers harm but loses more useful gain in all 21 comparisons, so the corrected NLL rises (\cref{fig:teaser}a); at the shared mean-case start, the gap lies in the correction response.
(B)~On CUB, a different objective, the monotonic penalty, shows each effect in isolation.
At a fixed start, an amplitude that equalizes source harm reproduces the penalty's mean harm, gain and NLL on single-concept corrections within $\pm0.0005$ nat (\cref{fig:teaser}b); with jointly trained starts, a post-hoc scale reaches the penalty's maximum sampled partial harm at lower NLL, and the penalty's higher risk is already present before any correction (\cref{fig:teaser}c).

Our contributions are:
\begin{itemize}
\item A correction audit that serves as a reproducible evaluation protocol: identical events, harm--gain and before/after splits, and amplitude baselines matched in source harm, with decision rules fixed in advance (\cref{sec:audit}). The splits are exact identities; the controls attribute a harm reduction to a smaller response, a worse start or better corrections, which accuracy and harm metrics, including those of~\cite{verghese2026harmful}, cannot (\cref{tab:reversals}).
\item Clinical evidence from three ultrasound sources, seven training-source combinations and six seeds that lower correction harm can mean less help, also when harm is used to select between training objectives (\cref{sec:res_ultrasound_gain}).
\item Controlled CUB evidence that a smaller response and a worse start each lower harm without better corrections, together with a candidate mechanism for the lost gain (\cref{sec:res_native,sec:res_frozen,sec:res_native_frontier}).
\item Expert BI-RADS concept annotations for the public BUSI647 and BUS-BRA images, \arxivor{released}{to be released} with their encoding, mapping and evaluation code (\suppref{sec:supp_compute}).
\end{itemize}

\section{Related Work}
\label{sec:related}

\paragraph{Concept interventions.}
Concept bottleneck models~\cite{koh2020concept} and concept embedding models~\cite{espinosa2022concept} expose concept predictions that users can overwrite at test time.
How much an intervention helps depends on the training regime, the selection and grouping of concepts, and the data~\cite{shin2023closer}.
Several lines of work make models more receptive to interventions: intervention-aware training~\cite{espinosa2023learning}, joint concept distributions that propagate a correction~\cite{vandenhirtz2024stochastic,hoffmann2025posthoc}, realignment after a correction~\cite{singhi2024improving}, causal feedback through the input-to-concept pathway~\cite{zhu2026causal}, intervenable black-box models~\cite{laguna2024beyond} and leakage-limiting mixtures~\cite{espinosa2025leakage}.
Leakage beyond the annotated concepts shapes what an intervention can achieve; it has been measured~\cite{parisini2025leakage}, removed or reduced~\cite{almudevar2026never,galliamov2026concepts,zabounidis2026disentangled}, and defended as benign~\cite{espinosa2026defense}.
An audit of concept-bottleneck accuracy on CUB studies the grounding and norm saturation of concept scores without evaluating interventions~\cite{saralajew2026auditing}.
Synthetic benchmarks show that a constraint can preserve training accuracy while removing a correction pathway~\cite{skirzynski2026measuring}.
Closest to us is concurrent work~\cite{verghese2026harmful}, which shows that these conditions do not guarantee helpful interventions, defines intervention harm as a difference of population risks, separates endpoint from partial harm, and trains with a monotonic intervention objective.
We keep its population-level path metrics, objective and CUB configuration and reproduce its lower path harm (\cref{sec:res_native}); where it measures and reduces harm, we ask whether lower harm means better corrections, which takes useful gain, absolute risk, the start and matched-harm amplitude baselines, on clinical data as well.
Concurrent work also models residuals jointly with the concepts during interventions~\cite{marcou2026residual}.

\paragraph{Concept models in medical imaging.}
Intervenable concept models have been built for pediatric appendicitis ultrasound~\cite{marcinkevics2024interpretable} and for breast-ultrasound lesion detection with BI-RADS concepts~\cite{bunnell2024learning}, where interventions are scored by AUROC; report-guided concept editing during training has also been proposed for breast ultrasound~\cite{yue2026trace}.
We use expert BI-RADS-based descriptor annotations to measure harm and useful gain on identical correction events.

\paragraph{Calibration and post-hoc rescaling.}
Temperature scaling is a standard one-parameter post-hoc calibration of classifier confidence~\cite{guo2017calibration}, and concept-probability calibration has been used inside concept models~\cite{espinosa2025leakage}.
Our post-hoc baselines act on a different object: they rescale the change of the task logits caused by a correction, leave the concept probabilities and the uncorrected prediction untouched, and are fitted on source data to a harm budget.
Bounded-strength interventions appear in~\cite{hoffmann2025posthoc}; we use response scaling, including coefficients above $1$, as a control fitted to a harm budget, never as a procedure for deployment.

\paragraph{Harm, benefit and baselines.}
Model updates that lower average error can introduce mistakes, motivating negative-flip tracking~\cite{yan2021positive}; fixed-reference intervention gains~\cite{skirzynski2026measuring} and the coupling of initial performance and intervention effectiveness~\cite{koh2020concept,shin2023closer,laguna2024beyond} have been studied.
We turn this coupling into controls for harm-reducing objectives.

\section{Controlled Correction Audit}
\label{sec:audit}

We compare models $A$ and $B$ on the same corrections and ask whether $A$'s lower harm reflects better corrected predictions, a weaker response to corrections, or a worse starting point.
Each control below removes one explanation.

\subsection{Harm and useful gain on identical events}
\label{sec:audit_hg}
Let $b(x)\in\mathbb{R}^K$ be the task logits of a model before intervention and $z_S(x)$ the logits after the concepts in a set $S$ are replaced by their true values.
For a correction event $(x,S)$ with label $y$ we write $\dl=\ell(z_S,y)-\ell(b,y)$, where $\ell$ is the cross-entropy, and report the mean positive and negative parts separately,
\begin{equation}
\Hm=\Ex\pos{\dl},\qquad \Gm=\Ex\pos{-\dl},
\label{eq:hg}
\end{equation}
so that the corrected risk satisfies $\Ex\,\ell(z_S,y)=\Ex\,\ell(b,y)+\Hm-\Gm$.
This accounting identity keeps three quantities visible that a single score mixes: the harm a correction causes, the useful change it brings, and the quality of the uncorrected prediction.
A lower $\Hm$ that comes with a larger loss of $\Gm$ means that a model simply corrects less, not better.

The two models are compared on \emph{identical} events.
Where both make the same concept error, the same strict partial corrections are enumerated for both, and we report the fraction of cases that this common-error support covers.
Where the design allows, the uncorrected logits $b$, the concept probabilities and the residual inputs are shared, so that a difference between models can only come from the correction response.
Such a fixed start shows whether lower harm is bought with lower gain; the amplitude controls then separate the size of the response from its direction.

\subsection{Absolute risk and the before/after split}
\label{sec:audit_path}
When models are trained jointly, their starts differ.
Following~\cite{verghese2026harmful}, the population risk after correcting $S$ is $R_S=\Ex\,\ell(z_S,y)$, and partial harm is $R_S-R_\varnothing$ for $\varnothing\neq S$ short of the full concept set.
Along 512 fixed random concept orders shared by both models, the maximum sampled partial harm is the largest $R_S-R_\varnothing$ over all prefixes, and the harmful-path prevalence is the fraction of orders with any prefix above $10^{-8}$ nat, as in~\cite{verghese2026harmful}.
For two models evaluated on the same images and orders, the contrast of partial harm splits exactly into an after- and a before-intervention part,
\begin{equation}
\begin{aligned}
&\big(R^A_S-R^A_\varnothing\big)-\big(R^B_S-R^B_\varnothing\big)\\
&\qquad=\underbrace{\big(R^A_S-R^B_S\big)}_{\text{after}}-\underbrace{\big(R^A_\varnothing-R^B_\varnothing\big)}_{\text{before}}.
\end{aligned}
\label{eq:split}
\end{equation}
A lower partial harm for $A$ can therefore come from better corrected predictions or from worse uncorrected ones.
The same holds for the net value of corrections: against Bayes posteriors, $\Gm-\Hm=I(Y;C_S\mid X)+K_0-K_1$, where $K_0$ and $K_1$ are the expected divergences of the uncorrected and corrected predictions from them, so a worse start raises $K_0$ and the net value without better corrections (\suppref{sec:supp_theory}).
We report both parts on the same maximizing prefixes, with absolute risk before and after correction; the split is exact accounting, not a causal decomposition.

\subsection{Post-hoc response amplitude}
\label{sec:audit_amp}
If an objective lowers harm only by shrinking the response, rescaling the response of the unpenalized model should reproduce its outcomes.
Apart from the MLP readouts of \cref{sec:res_selection}, all models studied here have task readouts that are additive over concepts, so correcting $S$ moves the logits by $\sum_{j\in S}D_j(x)$, where $D_j(x)$ is the change caused by correcting concept $j$ alone.
For a linear readout over concept probabilities, correcting concept $j$ from its predicted probability $q_j(x)$ to the true value $c_j$ gives $D_j(x)=(c_j-q_j(x))\,w_j$, with $w_j$ the readout column of concept $j$; in the concept embedding models used for ultrasound, $D_j$ also depends on the image through the concept embeddings.
A post-hoc amplitude family replaces the response by $\sum_{j\in S}\beta_j D_j(x)$ with $\beta\ge0$: a single scalar ($F_1$) or one coefficient per concept or concept field ($F_J$).
The family leaves $b$ and $q$ unchanged, and $\beta=1$ recovers the native response of the model.

The coefficients are fitted on source data only.
In normalized coordinates $\theta=\beta s$, where $s$ is a source root-mean-square scale of the response (class-centered for multiclass tasks, margin-based for binary tasks; \suppref{sec:supp_families}), we minimize
\begin{equation}
J_\lambda(\theta)=\Ex_{\mathrm{src}}\,\ell(z_S^\theta,y)+\lambda\,\Ex_{\mathrm{src}}\pos{\dl^\theta}+\tfrac{\kappa}{2d}\lVert\theta-\alpha s\rVert_2^2
\label{eq:obj}
\end{equation}
over $\theta\ge0$, with $d$ the dimension of $\theta$, $\kappa=10^{-3}$, and $\alpha$ the arm's own nonnegative global scale minimizing source cross-entropy without the harm term.
The multiplier $\lambda$ trades source likelihood against harm; $\lambda=0$ gives the regularized source-risk optimum of each model's response.

\subsection{Matched source harm and uncertainty}
\label{sec:audit_frontier}
To compare a penalized model $A$ with an unpenalized model $B$, both receive the same post-hoc family, and their amplitudes are chosen to reach the same \emph{source} harm
\begin{equation}
h_f=f\cdot\min\{\Hm^A_{\mathrm{src}}(\lambda{=}0),\,\Hm^B_{\mathrm{src}}(\lambda{=}0)\}
\label{eq:budget}
\end{equation}
for $f\in\{1,0.75,0.5,0.25,0.1\}$, by a bracketed root search over $\lambda\in[0,16]$ with re-solved coefficients and recomputed source harm at every step (tolerance $\max(10^{-10},10^{-4}h_f)$, at most 60 iterations).
$\Hm$, $\Gm$ and NLL are then evaluated on held-out data; matching source harm does not force evaluation harm to match, which is why all three are reported jointly, always with their intervals.

Uncertainty comes from a paired bootstrap that resamples source and evaluation groups together and refits the scale, $\alpha$, all coefficients and the budget in every draw (1000 draws).
With a margin $\delta=0.0005$ nat, a numerical tolerance set on development data rather than a task-level equivalence bound, a point is \emph{conservatively equivalent} only if the 95\% intervals of $\Delta\Hm$, $\Delta\Gm$ and $\Delta$NLL all lie in $\pm\delta$, so equivalence is \emph{aggregate}: it concerns mean outcomes at single operating points, not the responses themselves.
For fixed-start comparisons, a joint harm--gain (Pareto) advantage requires the upper bound of $\Delta\Hm$ to be at most $0$, the lower bound of $\Delta\Gm$ to be above $0$, and consistent directions across replicas.
When starting predictions differ, we also report the pre-intervention NLL contrast, because a harm--gain ordering alone does not imply an absolute-risk advantage; if the harm interval lies inside $\pm\delta$ and the NLL interval excludes zero, we report a risk improvement at approximately matched evaluation harm.
Other contrasts are unresolved or trade-offs between harm and useful gain.

\begin{findingbox}{The audit as a protocol}
\textbf{1.~Events.} Evaluate both models on identical correction events.\par
\textbf{2.~Report.} Harm $\Hm$, useful gain $\Gm$ and absolute risk before and after correction; with different starts, split contrasts by \cref{eq:split}.\par
\textbf{3.~Starts.} Read a fixed-start contrast as a comparison of correction responses and native starts as a comparison of models.\par
\textbf{4.~Amplitude.} Compare with post-hoc amplitudes fitted on source data and matched in source harm, global and per concept or field, refitted in every bootstrap draw.\par
\textbf{5.~Decide.} Apply the rules of \cref{sec:audit_frontier}, fixed in advance: equivalence within $\pm\delta$, a joint harm--gain advantage, or a risk improvement at approximately matched harm.
\end{findingbox}

\section{Experimental Setup}
\label{sec:setup}

\paragraph{Data.}
Clinical concept correction is studied on three breast-ultrasound sources: a private image set (Private250), the concept-annotated public BUSI subset (BUSI647)~\cite{aldhabyani2020dataset} and BUS-BRA~\cite{gomezflores2024busbra}, with six expert concept fields and benign/malignant labels.
BUS-BRA images carry expert annotations of six BI-RADS-based descriptor fields, separate from the official metadata and used in~\cite{yue2026trace}: 24 indicators, with margin multi-label and the other fields single-choice.
Two readers, a breast radiologist and a breast sonographer with 8 and 9 years of breast-ultrasound reading, annotated the fields independently from the images under one set of BI-RADS rules, blinded to reports, pathology, diagnosis and model predictions; a senior reader with 16 years of experience adjudicated each disagreement, \arxivor{and all experiments and the release use these final labels}{and all experiments use these final labels, which will also be released} (\suppref{sec:supp_annotation_stats}).
Private250 images are anonymized, and the providing hospital permits their use but not their release.
CUB fine-grained recognition~\cite{wah2011caltech} provides controlled tests of the two effects on a public benchmark of 200 bird species.

\paragraph{Clinical training-source combinations.}
ResNet-18 concept embedding models are trained with CorrectiveMean or CorrectiveWorst, which average or maximize the task loss over the legal correction states of each training image, on each single source, each source pair and all three sources.
Four folds and six training seeds give 336 fits; the development pools contain 199 Private250 image groups, 452 BUSI647 images and 732 BUS-BRA cases, with original group identifiers retained.
Every model is read out on the fixed evaluation folds of all three sources, and an omitted source enters neither training nor model selection.
Uncorrected same-source AUROC is $0.965$/$0.975$ on Private250, $0.869$/$0.885$ on BUSI647 and $0.714$/$0.728$ on BUS-BRA (Mean/Worst), and concept forecasts are weak (\suppref{tab:clinical_levels,tab:supp_annotations}).
Both arms are placed on the CorrectiveMean starting logits and evaluated on all 62 nonempty proper subsets of the six fields, each keeping its own concept forecasts and correction increments; a second analysis uses strict partial corrections on common-error support.
Intervals use 1000 paired bootstrap draws conditional on the trained models, so they reflect evaluation sampling, not retraining.

\paragraph{CUB, jointly trained models.}
To test the monotonic objective where it was proposed, we follow the CUB-200-2011 setup of~\cite{verghese2026harmful}: frozen, L2-normalized CLIP ViT-B/16 image features~\cite{radford2021learning}, standardized with statistics of the training split, and a class-stratified split into 7,488 training, 1,926 validation and 2,374 test images.
Eighty 50-attribute sets were regenerated with the procedure of~\cite{verghese2026harmful}: importance-biased candidates stratified by pre-model incompleteness (concept-probe validation cross-entropy relative to a null classifier), with eight sets per stratum.
For every set we train the linear hybrid of~\cite{verghese2026harmful}: 50 sigmoid concepts and a 32-dimensional residual from the image feature, followed by a linear 200-class readout.
Three regimes share initialization and minibatch order within a set: observational training, endpoint training ($\lambda_{\mathrm{int}}{=}1$), and endpoint plus monotonic training ($\lambda_{\mathrm{int}}{=}1$, $\lambda_{\mathrm{mono}}{=}100$, stop-gradient on concepts), 240 models in total.
We report raw models and models with a source-only temperature fitted on the validation split, which also selected the checkpoints; the joint bootstrap refits the response amplitudes and conditions on the trained checkpoints and their fitted temperatures.

\paragraph{CUB, frozen response maps.}
To remove the starting point, we fix the uncorrected logits $b$, the concept probabilities and the residual, and train only the $200\times50$ concept-to-task response matrix, with or without the monotonic penalty ($0$ or $100$), for 10 concept sets and 3 replicas.
Two backgrounds differ only in the base task loss (endpoint cross-entropy, or a proper partial-state cross-entropy, so that the penalty is also compared with plain partial-state training), giving 120 response maps that share one loss normalizer.
The post-hoc families are $F_1$ and $F_{50}$; the prespecified primary comparison is the endpoint background with $F_{50}$ at $f=0.5$.
These matched comparisons use the 50 single-concept corrections of every image, the first step of every concept order; multi-concept paths are evaluated separately.

\paragraph{Implementation.}
All post-hoc problems are solved in double precision with certified optima (\suppref{sec:supp_verify}); code and compute are described in \suppref{sec:supp_compute}.

\section{Results}
\label{sec:results}

\subsection{Lower harm, less help in clinical correction}
\label{sec:res_ultrasound_gain}
\begin{figure}[t]
\centering
\includegraphics[width=\linewidth]{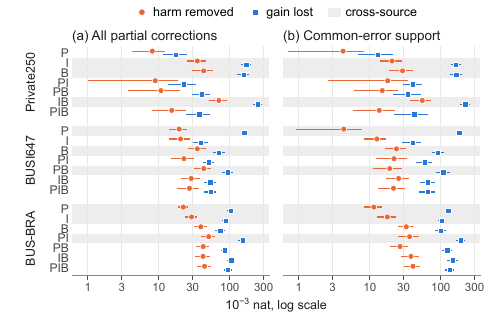}
\caption{Harm removed ($-\Delta\Hm$) and useful gain lost ($-\Delta\Gm$), CorrectiveWorst minus CorrectiveMean at the shared start of \cref{tab:ultrasound_gain} ($10^{-3}$ nat, log scale, marginal 95\% intervals).
\textbf{Gray rows} (\xsswatch) are \textbf{cross-source readouts}, as in \cref{tab:ultrasound_gain}.
Rows: training sources; blocks: evaluation source.
(a)~All partial corrections; (b)~strict partial corrections on common-error support.
In every row the gain lost exceeds the harm removed; \cref{tab:ultrasound_gain} and \suppref{tab:clinical_common} give the paired intervals of the resulting NLL increase.}
\label{fig:clinical}
\end{figure}

\begin{figure}[t]
\centering
\includegraphics[width=\linewidth]{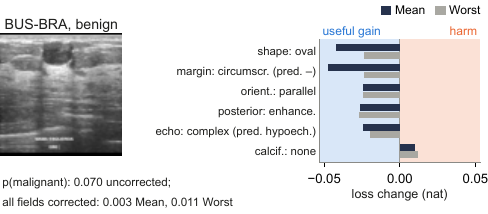}
\caption{A typical correction case. A benign BUS-BRA evaluation image~\cite{gomezflores2024busbra} and the loss change after correcting each concept field alone, CorrectiveMean and CorrectiveWorst at the shared CorrectiveMean start (BUS-BRA training, linear readout, one fold and seed); labels give the annotated value and, where it differs, the CorrectiveMean forecast.
The case has the median CorrectiveWorst-minus-CorrectiveMean corrected NLL among the evaluation cases whose single-field corrections both help and harm.}
\label{fig:case}
\end{figure}

\Cref{tab:ultrasound_gain} compares CorrectiveWorst with CorrectiveMean at CorrectiveMean's starting prediction for seven training-source combinations and three evaluation sources.
CorrectiveWorst lowers both harm and useful gain in all 21 comparisons and gives up more gain than it removes harm (\cref{fig:clinical}), so the corrected NLL rises, with all 21 conditional intervals above zero.
The 21 comparisons share sources and models, so they are related views of one study rather than independent replications.
For the three single-source, same-source comparisons, harm falls by $0.0081$, $0.0202$ and $0.0390$ nat on Private250, BUSI647 and BUS-BRA, whereas useful gain falls by $0.0173$, $0.0390$ and $0.0731$ nat.
The same ordering holds on common-error support and in the nine cross-source readouts (\suppref{tab:clinical_common,tab:clinical_levels}).
Split by the reference model's concept accuracy, CorrectiveWorst lowers harm and useful gain in every stratum, but its net cost arises only where concepts are predicted poorly (\suppref{tab:supp_strata}), which harm alone does not reveal.
In a typical case (\cref{fig:case}), correcting the shape or margin field lowers the loss about half as much for CorrectiveWorst as for CorrectiveMean, while the one harmful correction, of calcifications, harms both alike.
Both arms keep their own concept forecasts and correction increments, so the difference lies in the correction response; the CUB controls in \cref{sec:res_native,sec:res_frozen,sec:res_native_frontier} separate its amplitude from the model's starting point.

Intervention-aware training behaves alike: at a matched CEM's start and equal source harm, IntCEM~\cite{espinosa2023learning} adds less useful gain on BUS-BRA (corrected NLL $+42.7$ [$+20.8$, $+68.0$]$\times10^{-3}$ nat), and on BUS-BRA and CUB its native-start difference is almost all starting risk (\suppref{sec:supp_intcem}).

\begin{table}[t]
\centering
\footnotesize
\setlength{\tabcolsep}{6pt}
\begin{tabular}{@{}lccc@{}}
\toprule
 & \multicolumn{3}{c}{$\Delta$ corrected NLL ($10^{-3}$ nat) by evaluation source} \\
\cmidrule(l){2-4}
Training & Private250 & BUSI647 & BUS-BRA \\
\midrule
P & $+9.2$ & \xs$+140.1$ & \xs$+81.9$ \\
 & {\scriptsize$[+1.0, +18.3]$} & \xs{\scriptsize$[+121.3, +161.6]$} & \xs{\scriptsize$[+68.5, +94.8]$} \\\addlinespace[1.5pt]
I & \xs$+137.0$ & $+18.8$ & \xs$+58.6$ \\
 & \xs{\scriptsize$[+102.7, +170.7]$} & {\scriptsize$[+6.7, +31.9]$} & \xs{\scriptsize$[+46.1, +72.0]$} \\\addlinespace[1.5pt]
B & \xs$+114.7$ & \xs$+34.9$ & $+34.0$ \\
 & \xs{\scriptsize$[+73.6, +152.6]$} & \xs{\scriptsize$[+17.9, +54.3]$} & {\scriptsize$[+19.7, +48.6]$} \\\addlinespace[1.5pt]
PI & $+13.7$ & $+28.2$ & \xs$+100.8$ \\
 & {\scriptsize$[+1.1, +27.9]$} & {\scriptsize$[+14.3, +42.5]$} & \xs{\scriptsize$[+76.0, +124.6]$} \\\addlinespace[1.5pt]
PB & $+29.3$ & \xs$+50.6$ & $+42.4$ \\
 & {\scriptsize$[+13.7, +42.8]$} & \xs{\scriptsize$[+30.2, +73.3]$} & {\scriptsize$[+28.6, +57.9]$} \\\addlinespace[1.5pt]
IB & \xs$+178.5$ & $+24.5$ & $+63.4$ \\
 & \xs{\scriptsize$[+129.1, +229.6]$} & {\scriptsize$[+11.1, +37.9]$} & {\scriptsize$[+46.5, +81.4]$} \\\addlinespace[1.5pt]
PIB & $+22.1$ & $+27.0$ & $+50.3$ \\
 & {\scriptsize$[+5.2, +40.0]$} & {\scriptsize$[+12.8, +42.9]$} & {\scriptsize$[+34.5, +67.0]$} \\
\bottomrule
\end{tabular}
\caption{Corrected-NLL increase despite lower harm.
\textbf{Gray cells} (\xsswatch) are \textbf{cross-source readouts}: the evaluation source was not used in training.
Entries: CorrectiveWorst minus CorrectiveMean at the shared CorrectiveMean starting prediction, all partial corrections, with conditional 95\% intervals; four folds and six seeds, 336 fits.
Training sources: P, Private250; I, BUSI647; B, BUS-BRA.
All 21 intervals lie above zero; the underlying harm and useful-gain contrasts are shown in \cref{fig:clinical} and listed in \suppref{tab:clinical_all}.}
\label{tab:ultrasound_gain}
\end{table}

\paragraph{Exploratory: selecting by harm.}
\label{sec:res_selection}
On each clinical source, the models of both objectives (linear or MLP readout) start from the CorrectiveMean prediction and receive the same two field corrections per case, in a greedy order fitted on the calibration fold, where each criterion picks the lower-harm or the lower-NLL model.
On the evaluation fold, the harm pick has less harm but less useful gain, higher corrected NLL in every seed and fewer rescued predictions (\cref{tab:selection}).
This shared-start, shared-order contrast measures the cost of selecting by harm alone; in the primary contrast with native starts, most of the BUS-BRA linear difference is again starting risk, just as on CUB (\suppref{sec:supp_selection}).

\begin{table}[t]
\centering
\footnotesize
\setlength{\tabcolsep}{3pt}
\begin{tabular}{@{}lcccc@{}}
\toprule
 & $\Delta$NLL & $\Delta\Hm$ & $\Delta\Gm$ & $\Delta$Rescued \\
\midrule
Private250, linear & $+14.0$ {\scriptsize$[+8.9, +15.8]$} & $-3.0$ & $-17.0$ & $-1.59$ \\
Private250, MLP & $+10.1$ {\scriptsize$[+6.8, +11.3]$} & $-0.97$ & $-11.0$ & $-1.17$ \\
BUS-BRA, linear & $+42.9$ {\scriptsize$[+26.6, +64.0]$} & $-37.8$ & $-80.7$ & $-2.64$ \\
BUS-BRA, MLP & $+26.7$ {\scriptsize$[+16.9, +33.4]$} & $-22.3$ & $-49.1$ & $-3.37$ \\
BUSI647, linear & $+15.0$ {\scriptsize$[+2.3, +23.7]$} & $-10.0$ & $-25.0$ & $-0.77$ \\
BUSI647, MLP & $+18.6$ {\scriptsize$[+7.6, +27.1]$} & $-7.4$ & $-25.9$ & $-0.70$ \\
\bottomrule
\end{tabular}
\caption{Harm-selected minus NLL-selected model, two corrected fields per case, shared start and order ($10^{-3}$ nat; rescued: pp of all evaluation cases wrong before and right after correction), 95\% group-level bootstrap intervals; all $\Delta\Hm$, $\Delta\Gm$ and $\Delta$Rescued intervals exclude zero and $\Delta$NLL $>0$ in all six seeds (\suppref{tab:supp_selection_full,tab:selection_p_core}).}
\label{tab:selection}
\end{table}

\subsection{A worse start lowers start-referenced harm}
\label{sec:res_native}
On jointly trained CUB models, we compare endpoint-plus-monotonic (M) with endpoint (E) models in the configuration of~\cite{verghese2026harmful}.
On the reported path metrics the penalty reproduces the reported picture (\cref{tab:reproduce}): maximum sampled partial harm and harmful-path prevalence are lower in all 80 concept sets, accuracy barely moves, and the concept predictions are slightly better in all 80 sets.
The likelihood, which~\cite{verghese2026harmful} does not report, shows the opposite: pre-intervention, partially corrected and fully corrected NLL are all higher in 80 of 80 sets, by about $0.011$ nat for raw models and $0.005$ nat after a source-only temperature (\cref{fig:teaser}c).

\begin{table}[t]
\centering
\footnotesize
\setlength{\tabcolsep}{3pt}
\begin{tabular}{@{}lcc@{}}
\toprule
M $-$ E & Reported in~\cite{verghese2026harmful} & Ours \\
\midrule
Harmful-path prevalence (pp) & $-39.9$ & $-39.5$ \\
 & {\scriptsize$[-42.2, -37.5]$} & \\\addlinespace[1.5pt]
Max.\ sampled partial harm & $-5.0$ & $-3.92$ \\
 & {\scriptsize$[-5.3, -4.7]$} & \\\addlinespace[1.5pt]
Accuracy before correction (pp) & $-0.10$ & $-0.06$ \\
 & {\scriptsize$[-0.16, -0.04]$} & \\\addlinespace[1.5pt]
Concept mean squared error & $-0.0097$ & $-0.009$ \\
 & {\scriptsize$[-0.0107, -0.0087]$} & \\
\midrule
\rowcolor{cvprblue!7}Pre-intervention NLL & n.r. & $+4.9$ \\
\rowcolor{cvprblue!7} & & {\scriptsize$[+3.6, +6.2]$} \\[1pt]
\rowcolor{cvprblue!7}Partially corrected NLL & n.r. & $+4.7$ \\
\rowcolor{cvprblue!7} & & {\scriptsize$[+3.6, +5.9]$} \\[1pt]
\rowcolor{cvprblue!7}Fully corrected NLL & n.r. & $+5.4$ \\
\rowcolor{cvprblue!7} & & {\scriptsize$[+3.6, +7.1]$} \\
\bottomrule
\end{tabular}
\caption{Reproduction of the reported comparison, with the likelihood added: jointly trained CUB hybrids, endpoint plus monotonic training (M) minus endpoint training (E), as reported in~\cite{verghese2026harmful} for its 80 concept sets (95\% intervals over sets) and in our reproduction on 80 regenerated sets with temperature control (image-bootstrap intervals; none for the maximum-type path metrics, whose points can lie outside them).
Harm and NLL in $10^{-3}$ nat; n.r.: not reported in~\cite{verghese2026harmful}; raw models in \suppref{tab:supp_v_compare}.}
\label{tab:reproduce}
\end{table}
Two further seeds on 10 of the sets give the same two signs in all 30 seed--set pairs (\suppref{tab:supp_seeds}); the 80 sets share image features and recipe, so the counts describe concept sets rather than independent training runs.
On CUB, IntCEM shows the same signature against a matched CEM: it lowers the harmful-path prevalence by $12.6$ points but raises the partially corrected NLL by $0.012$ nat (\suppref{sec:supp_intcem}).

\begin{figure}[t]
\centering
\includegraphics[width=\linewidth]{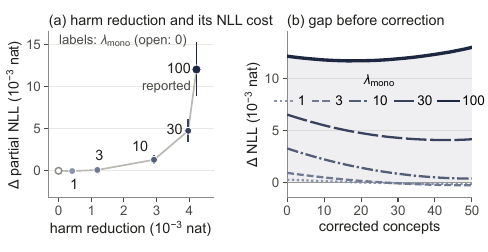}
\caption{Penalty strength on jointly trained CUB hybrids, monotonic penalty minus endpoint model, 10 concept sets $\times$ three seeds ($10^{-3}$ nat).
(a)~Reduction of the maximum sampled partial harm and change of the partially corrected NLL, with 95\% intervals for the NLL.
(b)~NLL difference after each number of corrected concepts, mean over the 512 orders.
Values: \suppref{tab:supp_strength}.}
\label{fig:strength}
\end{figure}

\Cref{eq:split} reconciles the two readings.
For raw models, the change of the maximum sampled partial harm ($-0.0042$) splits into a before-intervention part of $-0.0111$, the worse starting risk, and an after-intervention part of $+0.0069$ [$+0.0025$, $+0.0080$]: the corrected predictions on the maximizing prefixes are worse as well.
After temperature control the after-intervention part is indistinguishable from zero, and the before-intervention part ($-0.0049$) exceeds the total change ($-0.0039$).
The degradation comes with the monotonic term, not with the endpoint objective, and grows with its weight in a retrained sweep (\cref{fig:strength}).

At weights $1$ and $3$, the penalty gives 10\% and 28\% of the harm reduction of the reported weight $100$ at unchanged partially corrected NLL; from weight $10$ on, the NLL is higher before correction and stays higher along the path.
Plain cross-entropy at the same partial-correction states does the opposite: it never lowers the validation harm but lowers the partially corrected NLL, by $4.6$--$7.2\times10^{-3}$ nat (\suppref{sec:supp_results}).
Harm and help move separately.

\subsection{Fixed start: amplitude reproduces the penalty}
\label{sec:res_frozen}
\begin{figure}[t]
\centering
\includegraphics[width=\linewidth]{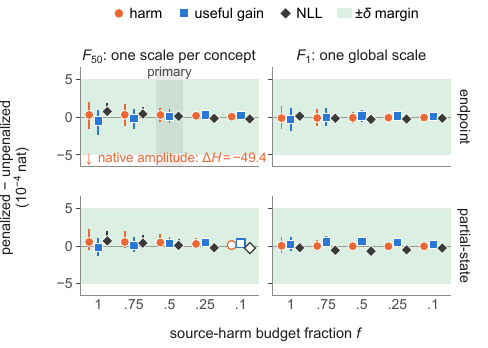}
\caption{Fixed start: matched amplitudes reproduce mean single-concept outcomes.
Frozen CUB response maps, penalized minus unpenalized harm, useful gain and NLL ($10^{-4}$ nat, 95\% joint intervals; green band: margin $\pm\delta$) in both backgrounds and both families at the five source-harm budgets.
Shaded: primary point; hollow: incomplete point (990 of 1000 draws); arrow: harm difference at native amplitude, off scale; values in \suppref{tab:frozen,tab:supp_frozen_sec}.}
\label{fig:contrasts}
\end{figure}

Freezing $b$, the concept probabilities and the residual removes the starting point, so the penalty can act only on the correction response.
In native units it acts strongly: the penalized response keeps 8.5\% of the unpenalized first-step harm and 9.3\% of its useful gain.
Fitting each arm's own amplitude at the regularized source-risk optimum ($\lambda=0$) shrinks the harm difference from $-49.4$ to $-14.8\times10^{-4}$ nat, and matching source harm brings all three differences inside $\pm\delta$ (\cref{fig:contrasts}).
At the primary operating point the differences are $+0.33$, $+0.18$ and $+0.15\times10^{-4}$ nat for $\Hm$, $\Gm$ and NLL, with all intervals inside $\pm\delta$.
The same holds at all 19 of the 20 prespecified points with complete resampling (two backgrounds, $F_1$ and $F_{50}$, five budgets), and neither arm gains a joint harm--gain advantage; the remaining point is incomplete (990 of 1000 draws) and stays inside $\pm\delta$ even with its missing draws set to the most adverse values.
Both arms keep a useful gain of about $1.9\times10^{-3}$ nat at the primary point, so matching does not reduce the responses to zero (\suppref{tab:frozen,tab:supp_abs}).
This is aggregate equivalence of mean single-concept outcomes, not response equivalence: a concept's two readout columns have median cosine $0.44$, with $77.5\%$ of the energy orthogonal, and large harms (above $0.05$ nat) are 9--19\% rarer for the penalized arm with $F_{50}$ but more frequent with $F_1$ (\suppref{sec:res_tail,sec:supp_diag}); multi-concept paths are compared with jointly trained models in \cref{sec:res_native_frontier}.
Some NLL intervals exclude zero, but all stay inside $\pm\delta$; half of $\delta$ would also contain every interval of the 19 points (post hoc; \suppref{tab:supp_margin}).
For scale (\suppref{tab:supp_abs}), $\delta$ is $0.07\%$ of the uncorrected NLL ($0.739$ nat), a $0.05\%$ change of the geometric-mean true-class probability, and about a quarter of the useful gain at the primary point.

\subsection{Amplitude baselines for jointly trained models}
\label{sec:res_native_frontier}
Amplitude also accounts for the path metrics of jointly trained models.
We give each endpoint model one post-hoc scalar, fitted on the validation split so that its first-step source harm equals that of the penalized model at native amplitude.
The scalar is small (median $0.13$; 5--95\% range $0.09$--$0.18$), in line with the strongly attenuated penalized response.
With it, the endpoint model matches the penalty's maximum sampled partial harm (difference $+0.011\times10^{-3}$ nat, $[-0.065,+0.097]\times10^{-3}$ conditional on the fitted scalar) and has lower NLL before, during and after correction, lower per-event partial harm and fewer large partial harms (\cref{fig:teaser}c; \suppref{tab:native}).
Shrinking costs E $0.002$ nat of fully corrected NLL, less than the penalty's $0.005$ nat.

\begin{figure}[t]
\centering
\includegraphics[width=\linewidth]{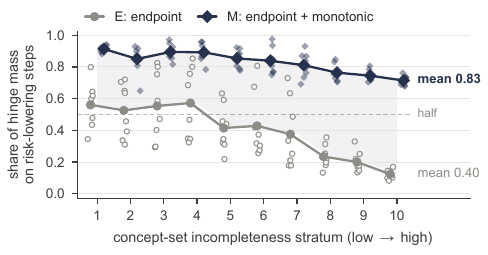}
\caption{Where the per-example penalty falls (CUB, training split).
Share of the per-example hinge mass $\Ex\pos{\dl}$ that lies on correction steps whose population risk $\Ex\,\dl$ falls, along the 512 orders of each concept set; small markers: the 80 sets, eight per incompleteness stratum (open circles: E; diamonds: M); lines: stratum means.
Observational models: $0.002$.}
\label{fig:mechanism}
\end{figure}

The penalty keeps fewer orders with any population-risk increase (45\% versus 61\%), yet its objective charges positive per-example loss increments rather than increases of population risk.
\Cref{fig:mechanism} shows where the charge falls.
Already for the endpoint model, which the penalty never shaped, 40\% of the per-example hinge mass lies on steps whose population risk falls; of what remains after training with the penalty, 83\% does on the training images, between 71\% and 91\% in every incompleteness stratum, and 70\% on the test images.
A synthetic example in which population risk falls at every step shows the same (\suppref{sec:supp_results}).
This is a candidate mechanism for the lost useful gain.

With all 80 pairs, both families and 1000 joint draws, the evaluation harm and useful-gain differences of temperature-controlled models lie inside $\pm\delta$ at every budget (\suppref{tab:supp_native}).
Because the starts differ, the NLL contrast contains the starting point, $\Delta\mathrm{NLL}=\Delta\mathrm{NLL}_{\mathrm{before}}+\Delta\Hm-\Delta\Gm$: at the primary point, the penalized-minus-endpoint contrast, $+4.90$ [$+3.57$, $+6.28$]$\times10^{-3}$ nat and positive in all 80 concept sets, is almost entirely the pre-intervention component, $+4.92$ [$+3.58$, $+6.29$]$\times10^{-3}$, with a response part $\Delta\Hm-\Delta\Gm$ of $-0.25\times10^{-4}$.
At every budget, the scaled endpoint model has the lower absolute NLL at approximately matched evaluation harm, and neither arm gains a joint harm--gain advantage in either amplitude family.

\begin{keyfindingsbox}
From the mean-case start, lower harm meant a larger loss of useful gain in all 21 pooled clinical comparisons. On CUB, matched amplitudes reproduce a harm penalty's mean single-concept outcomes at a fixed start; with joint training, its higher risk is starting risk.
\end{keyfindingsbox}

\section{Discussion}
\label{sec:discussion}

\paragraph{What a shared start measures.}
A fixed-start comparison answers which correction response is more useful from a given start, not which deployed model is better, and the answer can change with the start: with matched amplitudes, the BUS-BRA MLP comparison reverses from CorrectiveWorst's start ($-13.7$ [$-21.9$, $-2.6$]$\times10^{-3}$ nat), and so does IntCEM against a matched CEM (\suppref{sec:supp_clinical_frontier,sec:supp_intcem}).
Absolute risk at native starts, split by \cref{eq:split}, does not depend on this choice, so the audit reports both comparisons (\cref{sec:res_native_frontier}).

\paragraph{What simpler evaluations conclude.}
For all three audited objectives, harm and accuracy favor the lower-harm model, and an audit step changes each reading (\cref{tab:reversals}).

\begin{table}[t]
\centering
\footnotesize
\setlength{\tabcolsep}{3pt}
\begin{tabular}{@{}>{\raggedright\arraybackslash}p{0.42\linewidth}>{\raggedright\arraybackslash}p{0.55\linewidth}@{}}
\toprule
Harm and accuracy favor & The audit step that changes the reading \\
\midrule
Clinical, at Mean's start: Worst, lower harm in 21 of 21 & \emph{Gain}: more gain lost, so corrected NLL is higher in 21 of 21 \\\addlinespace[2pt]
Clinical: the model that harm selects on each source & \emph{Gain}: higher corrected NLL in every seed, fewer rescued predictions \\\addlinespace[2pt]
CUB: M, with lower path harm in 80 of 80 and accuracy barely changed ($-0.06$ pp) & \emph{Split}: NLL higher in 80 of 80, almost all before correction; \emph{amplitude}: scaled E has M's harm at lower NLL \\\addlinespace[2pt]
CUB, fixed start: the penalty, with 8.5\% of the harm & \emph{Amplitude}: matched scaling reproduces the mean $\Hm$, $\Gm$ and NLL \\\addlinespace[2pt]
CUB: IntCEM, with 12.6 points fewer harmful paths & \emph{Split}: partially corrected NLL $+0.012$ nat, almost all starting risk \\
\bottomrule
\end{tabular}
\caption{Where harm misleads. Worst, Mean: CorrectiveWorst, CorrectiveMean; M, E: with and without the monotonic penalty.}
\label{tab:reversals}
\end{table}

\begin{takeawaybox}
Lower harm signals better corrections only when read with useful gain and absolute risk on identical events, against amplitude baselines matched in source harm.
\end{takeawaybox}

\paragraph{Limitations.}
Our clinical models predict concepts weakly, below the majority rate for every BUS-BRA field (\suppref{sec:supp_annotation_stats}), and stronger concept predictors may change the net effect.
The evidence is pooled over datasets and concept sets.
All clinical comparisons use development data; the one independent test, a reserved BUS-BRA cohort of 146 cases, points the same way ($+8.3$ [$-1.4$, $+20.6$]$\times10^{-3}$ nat) but does not exclude zero (\suppref{sec:supp_sealed}).
Corrections set annotated values on retrospective images and ignore query choice, reader trust and cost, so NLL changes are not clinical outcomes.

\section{Conclusion}
\label{sec:conclusion}
Lower harm does not by itself mean better corrections: a model can respond less or start from a worse prediction.
From the mean-case start, clinical harm reduction cost more useful gain in all 21 pooled comparisons; on CUB, matched amplitudes reproduce a harm penalty's mean single-concept outcomes, and its higher risk with joint training is starting risk.
Reporting useful gain, absolute risk, both starts and amplitude-matched baselines separates these effects from better use of the corrections that human experts provide.

{
    \small
    \bibliographystyle{ieeenat_fullname}
    \bibliography{main}
}

\clearpage
\arxivor{\setcounter{page}{1}}{}
\setcounter{dbltopnumber}{3}\setcounter{topnumber}{3}\setcounter{totalnumber}{5}
\renewcommand{\dbltopfraction}{0.9}\renewcommand{\topfraction}{0.9}\renewcommand{\textfraction}{0.08}
\renewcommand{\dblfloatpagefraction}{0.6}\renewcommand{\floatpagefraction}{0.6}
\setlength{\textfloatsep}{12pt plus 2pt minus 2pt}\setlength{\dbltextfloatsep}{12pt plus 2pt minus 2pt}
\maketitlesupplementary
\appendix

\noindent \arxivor{This supplement follows the order of the paper; citations refer to the reference list of the main paper.}{This appendix follows the order of the paper.}

\section{Post-hoc families in detail}
\label{sec:supp_families}

\paragraph{Events and families.}
On CUB the correction events are the 50 single-concept corrections of every image, \ie, the first step of every concept order, with $D_j(x)=(c_j-q_j(x))\,w_j$.
On BUS-BRA they are the strict partial corrections on the common-error support of the shared concept predictor.
The task has one logit $t$, and $X_{ef}$ denotes the change of $t$ caused by correcting field $f$ in event $e$.
With $\theta\ge0$ in normalized coordinates, $F_1$ gives $z=b+(\theta/s)\sum_{j\in S}D_j$ and $F_J$ gives $z=b+\sum_{j\in S}(\theta_j/s)\,D_j$, with $J=50$ concepts on CUB and $6$ fields on BUS-BRA.
Per-concept and per-field families require responses additive over concepts or fields; an MLP readout is not, so its matched comparisons use only $F_1$ (\cref{sec:supp_clinical_frontier,sec:supp_selection_mlp}).

\paragraph{Response scale.}
On CUB,
\begin{equation*}
s^2=\frac{\sum_i w_i\,\frac{1}{JK}\sum_{j,k}\big(D_{jk}(x_i)-\bar D_j(x_i)\big)^2}{\sum_i w_i},
\end{equation*}
where $\bar D_j$ is the mean of $D_j$ over the $K$ classes and $w_i$ are the source weights; only the class-centered part of a logit change affects the softmax.
On BUS-BRA, $s^2=\tfrac14\sum_e w_e\big(\sum_f X_{ef}\big)^2$ with normalized event weights, \ie, half the root-mean-square margin response, which is the class-centered RMS of the equivalent two-logit model with logits $\pm t/2$.
The scale is recomputed in every bootstrap draw and, within a draw, held fixed along $\lambda$ together with the prior center, the checkpoint and the event support.

\paragraph{Objective, prior center and ridge.}
Each arm has its own prior center $c=\alpha s$, where $\alpha$ minimizes the source cross-entropy of that arm over one nonnegative global scalar.
The ridge $\tfrac{\kappa}{2d}\lVert\theta-c\rVert_2^2$ with $\kappa=10^{-3}$ pulls coordinates with little source signal toward the global solution.
With $F_1$ the ridge and its gradient vanish at the source optimum, so $F_1$ at $\lambda=0$ is exactly the global scale.
The cross-entropy is convex in the logits, the logits are affine in $\theta$, and the harm term is the positive part of a convex function; $J_\lambda$ is convex for every $\lambda\ge0$, and strictly convex because of the ridge.

\paragraph{Coordinate solutions and certificates.}
On CUB, each fitted event corrects exactly one concept: its logits depend on only that concept's coefficient.
With the source scale and prior center fixed along $\lambda$, both the event-wise cross-entropy and positive-part loss terms separate over concepts, and the diagonal ridge adds no coupling.
Thus $J_\lambda$ is a sum of 50 one-dimensional objectives for $F_{50}$; $F_1$ instead ties all concept coefficients to one scalar.
This separability applies to first-step fitting, not to multi-concept path states.
Along $\theta_j$ the loss change of an event is convex and zero at $\theta_j=0$, so its harm term has kinks only at $0$ and at most one positive crossing.
We solve each problem by a bracketed Newton iteration on the one-sided derivatives $g_-$ and $g_+$.
We accept $\theta_j>0$ if $g_-\le10^{-10}$ and $g_+\ge-10^{-10}$, and $\theta_j=0$ if $g_+\ge-10^{-10}$.
A minimizer at a kink is certified by a sign change of the one-sided derivative across a bracket of width at most $1.3\times10^{-13}$.
On BUS-BRA a strict partial correction can change several fields at once, so the six coordinates of $F_6$ interact.
Because the logistic loss is monotone in the margin, the loss of event $e$ rises exactly when the correction moves the margin the wrong way, and the kinks of the harm term lie on the hyperplanes $\sum_f X_{ef}\theta_f=0$; 20\% of the source events mix signs across fields.
We solve $F_6$ with a kink-aware active-set method and require the minimum-norm element of the subdifferential over the active kinks and bounds, computed by bounded least squares, to be at most $10^{-10}$.
On BUS-BRA, $F_1$ is smooth for $\theta>0$.

\paragraph{Root search over $\lambda$.}
\label{sec:supp_root}
For every pair, family and draw we first solve the grid exactly for both arms; source harm never increased along the grid in any computation.
If the budget $h_f$ falls on a grid point, that point is used.
Otherwise the bracketing grid interval is refined by an Illinois regula falsi iteration with safeguarded Newton proposals from the implicit derivative $dH_{\mathrm{src}}/d\lambda$.
Every trial value re-solves $\theta$ exactly and recomputes source harm directly; interpolation is used only for warm starts.
The search stops when $|H_{\mathrm{src}}(\lambda)-h_f|\le\max(10^{-10},10^{-4}h_f)$, after at most 60 iterations, and never extrapolates beyond $\lambda=16$.
A budget is unreachable if $H_{\mathrm{src}}(16)>h_f$; at $f=1$ the arm with the smaller harm at $\lambda=0$ stays there by construction.
For the jointly trained models we add one safeguard: we force a bisection step whenever the bracket has not halved over two consecutive iterations.

\paragraph{Budget reachability.}
No budget was unreachable in the settings of \cref{tab:supp_reach}; the clinical calibration analysis of \cref{sec:supp_clinical_frontier} has unreachable budgets.
At $f=1$ the penalized arm is the smaller-harm arm and stays at $\lambda=0$ in 29,929 of 30,000 frozen endpoint draws and, with $F_{50}$ on the jointly trained models, in 76,047 (temperature-controlled) and 74,548 (raw) of 80,000 set-draw pairs.
No matched multiplier exceeds $0.97$ on the frozen maps or $0.79$ on the jointly trained models, far from the end of the grid.
At $f=0.5$, $F_{50}$ sets 40--69\% of the coefficients to zero, so the matched harm is concentrated on a minority of concepts.
With $F_1$ on the jointly trained models, 11,039 (temperature-controlled, 13.8\%) and 10,742 (raw, 13.4\%) of the 80,000 set-draw pairs have a zero budget, because the source-optimal global scale of the endpoint model is zero; both arms then have $\theta=0$.

\begin{table}[t]
\centering
\small
\setlength{\tabcolsep}{2.5pt}
\resizebox{\columnwidth}{!}{%
\begin{tabular}{@{}llrrcc@{}}
\toprule
Setting & Family & Points & Unres. & $\lambda^*$, unpenalized & $\lambda^*$, penalized \\
\midrule
Frozen, endpoint & $F_{50}$ & 300k & 0 & $0.140$ ($0.100$--$0.184$) & $0.116$ ($0.090$--$0.142$) \\
Frozen, endpoint & $F_1$ & 300k & 0 & $0.077$ ($0.046$--$0.108$) & $0.082$ ($0.045$--$0.121$) \\
Frozen, partial-state & $F_{50}$ & 300k & 10 & $0.158$ ($0.093$--$0.219$) & $0.120$ ($0.090$--$0.150$) \\
Frozen, partial-state & $F_1$ & 300k & 0 & $0.096$ ($0.051$--$0.136$) & $0.132$ ($0.060$--$0.230$) \\
BUS-BRA, both rules & $F_6$, $F_1$ & 480k & 0 & -- & -- \\
Joint, temperature & $F_{50}$ & 800k & 0 & $0.096$ ($0.057$--$0.138$) & $0.106$ ($0.082$--$0.134$) \\
Joint, temperature & $F_1$ & 800k & 0 & $0.014$ ($0.000$--$0.041$) & $0.066$ ($0.033$--$0.250$)$^\ast$ \\
Joint, raw & $F_{50}$ & 800k & 0 & $0.096$ ($0.058$--$0.139$) & $0.110$ ($0.085$--$0.140$) \\
Joint, raw & $F_1$ & 800k & 0 & $0.014$ ($0.000$--$0.042$) & $0.063$ ($0.028$--$0.250$)$^\ast$ \\
\bottomrule
\end{tabular}}
\caption{Source budget matching.
Points: pairs $\times$ draws $\times$ 5 budgets $\times$ 2 arms (BUS-BRA: also 2 families and 2 checkpoint rules).
Unres.: roots that missed the tolerance within 60 iterations (\cref{sec:supp_plans}).
$\lambda^*$: median (5--95\% range) of the matched multiplier at $f=0.5$.
$^\ast$Includes the zero-budget pairs, where the endpoint arm stays at $\lambda=0$ and the penalized arm reaches zero harm at the grid point $\lambda=0.25$.}
\label{tab:supp_reach}
\end{table}

\section{Analysis protocol}
\label{sec:supp_plans}

\paragraph{Fixed in advance.}
Before computing any outcome of the post-hoc comparisons, we fixed the families, the objective \cref{eq:obj} with its scale, prior center and ridge, the grid $\lambda\in\{0,0.25,0.5,1,2,4,8,16\}$, the budget fractions, the root search with its tolerance and iteration cap, the bootstrap, the decision rules with $\delta$, and the primary points.
These are the endpoint background with $F_{50}$ at $f=0.5$ for the frozen CUB maps (\cref{sec:res_frozen}), the originally selected checkpoint with equal case weights for BUS-BRA (\cref{sec:res_bus}), and the temperature-controlled checkpoints for the jointly trained CUB models (\cref{sec:res_native_frontier}), whose plan, written after the native-amplitude results of \cref{sec:res_native}, also fixed the 512-order path audit with the one-scalar comparison of \cref{tab:native} and added a bisection safeguard to the root search (\cref{sec:supp_root}).
We report every prespecified point and chose no point, family or checkpoint by its evaluation result.
The jointly trained models have no training replicas, so agreement in at least 7 of the 10 incompleteness strata serves as a directional consistency check.
The later analyses, namely the reserved-cohort contrast, the intervention-aware comparisons, the synthetic counterexamples, the bottleneck control, the symmetric clinical calibration, the penalty-strength sweep with its controls, the per-example penalty share and the clinical per-event diagnostic, also had their design and decision rule fixed before they were run; post-hoc additions are marked as such.
The finite-query selection analysis of \cref{sec:res_selection} was one of several exploratory analyses, with the MLP readouts added after the linear results; its primary contrast uses each model's own start and uncertainty order (\cref{tab:supp_selection_controls}), and the main text shows its shared-start, shared-order contrast because that contrast isolates the correction response.
No decision rule applies to the tail comparisons of \cref{sec:res_tail} or the diagnostics of \cref{sec:supp_diag}.

\paragraph{Deviations.}
In the partial-state background with $F_{50}$, 10 of 300,000 budget roots (penalized arm, $f=0.1$) missed the tolerance within 60 iterations.
We report this point from its 990 complete draws and mark it incomplete; placing the 10 missing draws at the most adverse values ($\pm\infty$) gives $\Delta\Hm\in[-0.004,+0.66]$, $\Delta\Gm\in[+0.18,+0.81]$ and $\Delta$NLL $\in[-0.48,+0.18]$ ($10^{-4}$ nat), all inside $\pm\delta$.
Instead of refitting the unpenalized global scalar $\alpha$ in every draw, we reused the re-certified per-draw fits of the earlier analyses; a different stopping rule makes them differ by a relative $3.8\times10^{-6}$ at most, which changes evaluation NLL by at most $5.3\times10^{-10}$ nat and $\Hm$, $\Gm$ by at most $8.0\times10^{-9}$ nat.
At $\lambda=0$, $F_6$ reproduces the earlier per-field fits to the earlier solver's KKT tolerance of $10^{-7}$ (coefficients within $1.45\times10^{-5}$).
In the per-field family of \cref{sec:supp_clinical_frontier}, five numerical exceptions of the solver count as solver failures, as every unresolved root does.
Conservative equivalence and the within-margin description are not exclusive; when both hold, we report both (E and W in \cref{tab:supp_frozen_sec}).

\section{Public ultrasound training recipe}
\label{sec:supp_public_recipe}

The encoder is an ImageNet-initialized ResNet-18. Each of the 24 concept indicators has a 16-dimensional embedding, followed by a linear $384\to2$ task head.
Images are converted to RGB, resized with preserved aspect ratio and black padding to $299\times299$, and normalized with ImageNet statistics; horizontal flips are used during training.
We use SGD with learning rate $0.01$, momentum $0.9$, weight decay $4\times10^{-6}$ and batch size 64.
The learning-rate scheduler reduces the rate by $0.1$ after 10 plateau checks.
Training runs for at most 150 epochs, validates every five epochs and stops after five validation checks without improvement.
Training-time concept interventions have probability $0.25$.
The base is frozen for the branch comparison; both branches train their complete encoder and concept/task modules, with the same loss normalizer and monotonic coefficients zero or one.
The shared starting prediction uses a source-only inverse temperature fitted on the 113 calibration cases, constrained to $[0.05,20]$.
That temperature and the checkpoints remain fixed during the bootstrap.

The \arxivor{released implementation}{implementation to be released} trains the 12 base models and 24 branches from the public images and the concept annotations alone and reproduces the reported BUS-BRA audit, including its 1000 jointly refitted source/evaluation draws; it uses PyTorch 2.8.0, torchvision 0.23.0, PyTorch Lightning 1.9.5, NumPy 2.3.2, SciPy 1.15.3 and the public CEM implementation.

\section{Concept annotations and concept forecasts}
\label{sec:supp_annotation_stats}

\paragraph{Annotation protocol.}\looseness=-1
Two readers annotated the concept fields independently: a radiologist specialized in breast imaging and a sonographer specialized in breast ultrasound, with 8 and 9 years of breast-ultrasound reading, who both read breast ultrasound routinely and know the BI-RADS ultrasound lexicon.
They annotated directly from the images that the models take as input, following one set of BI-RADS rules for the six fields (shape, margin, orientation, posterior acoustic features, echo pattern and calcifications), with the original reports hidden; a field could be marked as not assessable when the image did not show enough.
Fields on which the two readers agreed were adopted, and every disagreement was resolved by a senior reader with 16 years of related experience who routinely signs breast-ultrasound reports.
\arxivor{All experiments and the annotations released with this paper use these final adjudicated labels.}{All experiments use these final adjudicated labels, which are also the labels that will be released.}
All three readers were blinded to pathology, the final diagnosis and the model predictions; the benign/malignant labels were attached only after the concept annotation was complete.

\paragraph{Ethics and data use.}
Private250 consists of retrospective, anonymized images; its use was approved by an institutional ethics committee\arxivor{ (name withheld for anonymous review)}{}, informed consent was waived because the images are retrospective and anonymized, and the providing hospital permits their use \arxivor{for this study }{}but not their release.
The public BUSI647 and BUS-BRA images are used as released, under the terms of their original publications~\cite{aldhabyani2020dataset,gomezflores2024busbra}.

\paragraph{Annotations and diagnosis.}
On the development pools used in the experiments (732 BUS-BRA cases and 452 BUSI647 images), the annotated descriptors relate to the diagnosis in the directions of the BI-RADS lexicon.
Irregular shape, a not-parallel orientation, angular, indistinct, microlobulated or spiculated margins and posterior shadowing are more frequent among malignant lesions; oval shape, circumscribed margins and a parallel orientation are more frequent among benign ones (\cref{tab:supp_annotations}).
In both pools the margin pattern carries the most information about the diagnosis and calcifications the least.

The concept forecasts of the clinical models are weak.
On BUS-BRA, their argmax accuracy is below the rate of the most frequent class for every field, objective and readout; on BUSI647 it exceeds that rate only for shape and, for the linear CorrectiveWorst models, posterior features.
A post-hoc mean one-vs-rest AUROC is $0.50$--$0.59$ per field on BUS-BRA and $0.51$--$0.67$ on BUSI647 over objectives and readouts.
Most corrections therefore replace weakly informative forecasts, and the harm and gain of \cref{sec:res_ultrasound_gain} describe models of this kind.

\begin{table*}[t]
\centering
\footnotesize
\setlength{\tabcolsep}{3.5pt}
\begin{tabular}{@{}lcccccccccc@{}}
\toprule
 & \multicolumn{5}{c}{BUS-BRA (732 cases)} & \multicolumn{5}{c}{BUSI647 (452 images)} \\
\cmidrule(lr){2-6}\cmidrule(l){7-11}
 & \multicolumn{2}{c}{Annotation vs.\ diagnosis} & \multicolumn{3}{c}{Concept forecast} & \multicolumn{2}{c}{Annotation vs.\ diagnosis} & \multicolumn{3}{c}{Concept forecast} \\
\cmidrule(lr){2-3}\cmidrule(lr){4-6}\cmidrule(lr){7-8}\cmidrule(l){9-11}
Field & AUROC & MI share & Acc. & Majority & AUROC$^\dagger$ & AUROC & MI share & Acc. & Majority & AUROC$^\dagger$ \\
\midrule
Shape & $0.768$ & $0.28$ & $0.677$ & $0.699$ & $0.534$ & $0.765$ & $0.22$ & $0.605$ & $0.588$ & $0.608$ \\
 & {\scriptsize$[0.724, 0.799]$} &  &  &  &  & {\scriptsize$[0.701, 0.794]$} &  &  &  &  \\\addlinespace[1.5pt]
Margin (pattern) & $0.803$ & $0.30$ & $0.272$ & $0.596$ & $0.532$ & $0.856$ & $0.44$ & $0.288$ & $0.595$ & $0.609$ \\
 & {\scriptsize$[0.749, 0.826]$} &  &  &  &  & {\scriptsize$[0.792, 0.881]$} &  &  &  &  \\\addlinespace[1.5pt]
Orientation & $0.712$ & $0.21$ & $0.761$ & $0.791$ & $0.567$ & $0.729$ & $0.26$ & $0.759$ & $0.810$ & $0.618$ \\
 & {\scriptsize$[0.663, 0.740]$} &  &  &  &  & {\scriptsize$[0.660, 0.761]$} &  &  &  &  \\\addlinespace[1.5pt]
Posterior features & $0.617$ & $0.07$ & $0.395$ & $0.439$ & $0.542$ & $0.738$ & $0.20$ & $0.424$ & $0.438$ & $0.608$ \\
 & {\scriptsize$[0.556, 0.663]$} &  &  &  &  & {\scriptsize$[0.676, 0.779]$} &  &  &  &  \\\addlinespace[1.5pt]
Echo pattern & $0.559$ & $0.04$ & $0.749$ & $0.831$ & $0.529$ & $0.720$ & $0.21$ & $0.527$ & $0.619$ & $0.590$ \\
 & {\scriptsize$[0.478, 0.566]$} &  &  &  &  & {\scriptsize$[0.655, 0.747]$} &  &  &  &  \\\addlinespace[1.5pt]
Calcifications & $0.510$ & $0.01$ & $0.938$ & $0.952$ & $0.594$ & $0.573$ & $0.06$ & $0.857$ & $0.892$ & $0.514$ \\
 & {\scriptsize$[0.443, 0.524]$} &  &  &  &  & {\scriptsize$[0.503, 0.602]$} &  &  &  &  \\\addlinespace[1.5pt]
\bottomrule
\end{tabular}
\caption{Concept annotations and concept forecasts on the public development pools. Annotation vs.\ diagnosis: cross-fitted single-field AUROC for malignancy (class malignancy rates from the other three evaluation folds; margin as its six-bit pattern) with 95\% percentile intervals of 1000 unit-bootstrap draws that refit the rates, which are descriptive and can exclude the point for weak fields; MI share: mutual information with the diagnosis as a share of its entropy. Concept forecast: CorrectiveMean, linear readout, evaluation role, mean of six seeds; Acc.: argmax accuracy (margin: exact six-bit pattern); Majority: rate of the most frequent class or pattern; $^\dagger$post hoc, mean one-vs-rest AUROC of the indicators of each field. Each image carries one final, adjudicated annotation.}
\label{tab:supp_annotations}
\end{table*}

\paragraph{Concept accuracy and the clinical contrast.}
We split the same-source cases of BUS-BRA and BUSI647 by the share of annotated fields that CorrectiveMean, the reference model, predicts correctly, with strata and decision rule fixed in advance (\cref{tab:supp_strata}); the pooled points equal those of \cref{tab:ultrasound_gain}.
In every stratum CorrectiveWorst lowers both harm and useful gain.
The corrected-NLL increase, however, comes from the low and middle strata; in the high stratum, 31\% and 24\% of the units with 88\% and 89\% of the fields predicted correctly, harm and useful gain fall by similar amounts and the corrected-NLL difference is unresolved on both sources.
Stratifying by CorrectiveWorst's own concept accuracy gives the same picture.
The net cost of worst-case training thus grows as concept forecasts weaken, which harm alone does not show.

\begin{table}[t]
\centering
\footnotesize
\setlength{\tabcolsep}{1.8pt}
\begin{tabular}{@{}lccc@{}}
\toprule
Stratum & Share/acc. & $\Delta$NLL & $\Delta\Gm$ \\
\midrule
\multicolumn{4}{@{}l}{\emph{BUS-BRA}} \\
low & $0.18$/$0.30$ & $+140.9$ {\scriptsize[$+91.4$, $+192.2$]} & $-175.6$ {\scriptsize[$-221.1$, $-137.8$]} \\
middle & $0.51$/$0.60$ & $+24.2$ {\scriptsize[$+7.3$, $+42.7$]} & $-67.3$ {\scriptsize[$-80.8$, $-54.9$]} \\
high & $0.31$/$0.88$ & $-11.1$ {\scriptsize[$-27.5$, $+3.5$]} & $-23.7$ {\scriptsize[$-31.4$, $-17.9$]} \\
\midrule
\multicolumn{4}{@{}l}{\emph{BUSI647}} \\
low & $0.25$/$0.28$ & $+50.7$ {\scriptsize[$+23.2$, $+80.0$]} & $-73.1$ {\scriptsize[$-100.7$, $-49.7$]} \\
middle & $0.51$/$0.58$ & $+13.5$ {\scriptsize[$+0.6$, $+26.6$]} & $-34.5$ {\scriptsize[$-45.5$, $-24.6$]} \\
high & $0.24$/$0.89$ & $-3.2$ {\scriptsize[$-16.9$, $+10.3$]} & $-13.0$ {\scriptsize[$-22.2$, $-5.2$]} \\
\bottomrule
\end{tabular}
\caption{Concept-accuracy strata of the same-source clinical contrast of \cref{tab:ultrasound_gain} ($10^{-3}$ nat; linear readout, six seeds, four folds). Strata by the share of fields that CorrectiveMean predicts correctly (low: $\le1/3$, middle: $\le2/3$, high: $>2/3$); share of seed--case units / mean accuracy; 95\% case-group bootstrap intervals conditional on the trained models. $\Delta\Hm$ lies below zero in every stratum (points $-34.7$, $-43.1$, $-34.9$ on BUS-BRA and $-22.5$, $-21.0$, $-16.2$ on BUSI647).}
\label{tab:supp_strata}
\end{table}

\section{Clinical source-combination results}
\label{sec:supp_clinical_sources}

These results complete \cref{sec:res_ultrasound_gain} and \cref{tab:ultrasound_gain}.
P denotes Private250, I denotes BUSI647, and B denotes BUS-BRA.
The study comprises 336 fits: seven source combinations, two objectives, four folds and six training seeds, paired between objectives.
The fixed-start readout adds each model's correction increment to the CorrectiveMean starting logits while retaining its own concept forecasts.
All partial-mask contrasts average the 62 nonempty proper subsets of six fields. Common-error contrasts restrict corrections to fields both models get wrong and exclude images with no eligible partial correction.
The intervals are 95\% intervals from 1000 paired bootstrap draws conditional on the trained models, not intervals over the training algorithm.
Resampling groups are the original image groups of Private250 and BUSI647 and the official Cases of BUS-BRA.

\Cref{tab:clinical_all,tab:clinical_common} give the full harm/gain decomposition under the two event definitions.
All 21 corrected-NLL intervals are above zero in each analysis, but several individual seed point estimates have the opposite sign (\cref{tab:clinical_seeds}).
For example, P$\to$P ranges from $-6.9$ to $+25.4\times10^{-3}$ nat across seeds, despite the positive conditional pooled interval.
\Cref{tab:clinical_levels} reports absolute harm, useful gain and task AUROC.
Several source-absent models discriminate weakly, so their positive NLL contrasts characterize correction behavior rather than successful clinical transfer.
The displayed values are rounded independently; the rounded loss-balance identity can differ by $0.1\times10^{-3}$ nat.

For the six conditions without Private250, which \arxivor{the released data and code reproduce}{the data and code to be released will reproduce}, the last column of \cref{tab:clinical_seeds} also resamples the six seeds with the units (2000 draws): every interval stays above zero, all 36 seed estimates are positive, and the harm and gain intervals stay below zero; with the images fixed, a $t$ interval over seeds excludes zero in five of the six conditions ($I\!\to\!I$: [$-0.1$, $+37.6$]).

\begin{table*}[t]
\centering
\small
\setlength{\tabcolsep}{4pt}
\begin{tabular}{@{}lccc@{}}
\toprule
Train $\to$ evaluate & $\Delta\Hm$ & $\Delta\Gm$ & $\Delta$ partial NLL \\
\midrule
$P\!\to\!P$ & $-8.1 [-12.2, -4.2]$ & $-17.3 [-24.7, -11.6]$ & $+9.2 [+1.0, +18.3]$ \\
$P\!\to\!I$ & $-19.3 [-24.8, -14.1]$ & $-159.4 [-179.2, -142.2]$ & $+140.1 [+121.3, +161.6]$ \\
$P\!\to\!B$ & $-22.1 [-25.9, -18.5]$ & $-104.0 [-115.7, -92.0]$ & $+81.9 [+68.5, +94.8]$ \\
$I\!\to\!P$ & $-34.9 [-46.5, -24.7]$ & $-171.9 [-201.0, -144.3]$ & $+137.0 [+102.7, +170.7]$ \\
$I\!\to\!I$ & $-20.2 [-27.4, -13.9]$ & $-39.0 [-49.0, -29.9]$ & $+18.8 [+6.7, +31.9]$ \\
$I\!\to\!B$ & $-28.9 [-35.0, -23.6]$ & $-87.5 [-98.6, -77.2]$ & $+58.6 [+46.1, +72.0]$ \\
$B\!\to\!P$ & $-42.8 [-58.8, -29.0]$ & $-157.5 [-187.0, -128.1]$ & $+114.7 [+73.6, +152.6]$ \\
$B\!\to\!I$ & $-34.9 [-45.5, -25.6]$ & $-69.8 [-85.1, -57.5]$ & $+34.9 [+17.9, +54.3]$ \\
$B\!\to\!B$ & $-39.0 [-48.0, -31.4]$ & $-73.1 [-84.3, -62.6]$ & $+34.0 [+19.7, +48.6]$ \\
$PI\!\to\!P$ & $-8.9 [-18.6, -1.0]$ & $-22.6 [-33.2, -13.5]$ & $+13.7 [+1.1, +27.9]$ \\
$PI\!\to\!I$ & $-22.6 [-31.2, -14.7]$ & $-50.8 [-60.8, -41.7]$ & $+28.2 [+14.3, +42.5]$ \\
$PI\!\to\!B$ & $-50.4 [-62.6, -39.6]$ & $-151.2 [-171.4, -131.4]$ & $+100.8 [+76.0, +124.6]$ \\
$PB\!\to\!P$ & $-10.7 [-20.0, -3.7]$ & $-40.0 [-51.8, -29.8]$ & $+29.3 [+13.7, +42.8]$ \\
$PB\!\to\!I$ & $-43.0 [-55.2, -31.3]$ & $-93.6 [-111.4, -77.9]$ & $+50.6 [+30.2, +73.3]$ \\
$PB\!\to\!B$ & $-42.2 [-51.6, -33.6]$ & $-84.6 [-95.2, -74.9]$ & $+42.4 [+28.6, +57.9]$ \\
$IB\!\to\!P$ & $-70.3 [-91.5, -50.5]$ & $-248.9 [-288.9, -211.2]$ & $+178.5 [+129.1, +229.6]$ \\
$IB\!\to\!I$ & $-28.6 [-37.5, -20.5]$ & $-53.1 [-63.1, -43.7]$ & $+24.5 [+11.1, +37.9]$ \\
$IB\!\to\!B$ & $-42.0 [-52.4, -32.5]$ & $-105.4 [-118.7, -92.6]$ & $+63.4 [+46.5, +81.4]$ \\
$PIB\!\to\!P$ & $-15.2 [-24.5, -7.9]$ & $-37.3 [-52.8, -24.0]$ & $+22.1 [+5.2, +40.0]$ \\
$PIB\!\to\!I$ & $-27.0 [-36.8, -17.8]$ & $-54.0 [-64.4, -44.0]$ & $+27.0 [+12.8, +42.9]$ \\
$PIB\!\to\!B$ & $-44.1 [-54.6, -34.8]$ & $-94.4 [-106.2, -83.6]$ & $+50.3 [+34.5, +67.0]$ \\
\bottomrule
\end{tabular}
\caption{All partial masks: CorrectiveWorst minus CorrectiveMean at the fixed CorrectiveMean start, $10^{-3}$ nat. Brackets: conditional 95\% paired-bootstrap intervals. Each source contributes to training exactly when its letter appears on the left.}
\label{tab:clinical_all}
\end{table*}

\begin{table*}[t]
\centering
\small
\setlength{\tabcolsep}{4pt}
\begin{tabular}{@{}lcccc@{}}
\toprule
Train $\to$ evaluate & Coverage & $\Delta\Hm$ & $\Delta\Gm$ & $\Delta$ partial NLL \\
\midrule
$P\!\to\!P$ & $0.41\text{--}0.50$ & $-4.2 [-8.4, -0.7]$ & $-13.1 [-21.2, -6.8]$ & $+8.9 [+1.0, +18.1]$ \\
$P\!\to\!I$ & $0.86\text{--}0.92$ & $-4.3 [-7.8, -0.9]$ & $-182.8 [-203.8, -163.4]$ & $+178.5 [+158.8, +200.3]$ \\
$P\!\to\!B$ & $0.82\text{--}0.92$ & $-11.4 [-14.7, -8.3]$ & $-128.1 [-141.6, -115.5]$ & $+116.7 [+102.9, +131.4]$ \\
$I\!\to\!P$ & $0.79\text{--}0.92$ & $-20.6 [-28.1, -14.0]$ & $-164.7 [-190.5, -138.6]$ & $+144.2 [+114.7, +173.7]$ \\
$I\!\to\!I$ & $0.52\text{--}0.69$ & $-12.6 [-17.2, -8.3]$ & $-40.5 [-53.6, -28.9]$ & $+27.9 [+15.2, +42.6]$ \\
$I\!\to\!B$ & $0.56\text{--}0.72$ & $-17.7 [-23.3, -12.8]$ & $-103.3 [-116.1, -90.9]$ & $+85.6 [+71.6, +101.4]$ \\
$B\!\to\!P$ & $0.84\text{--}0.89$ & $-29.0 [-41.2, -18.9]$ & $-166.6 [-196.9, -135.4]$ & $+137.6 [+99.1, +174.9]$ \\
$B\!\to\!I$ & $0.56\text{--}0.77$ & $-23.8 [-32.5, -16.3]$ & $-91.3 [-111.0, -74.5]$ & $+67.6 [+48.0, +90.2]$ \\
$B\!\to\!B$ & $0.46\text{--}0.69$ & $-32.7 [-41.0, -24.7]$ & $-101.0 [-118.4, -84.2]$ & $+68.3 [+49.7, +88.6]$ \\
$PI\!\to\!P$ & $0.44\text{--}0.53$ & $-17.9 [-34.7, -2.6]$ & $-40.5 [-54.7, -29.2]$ & $+22.6 [+1.0, +45.4]$ \\
$PI\!\to\!I$ & $0.44\text{--}0.53$ & $-22.0 [-33.6, -12.2]$ & $-59.3 [-75.1, -44.6]$ & $+37.3 [+18.8, +58.2]$ \\
$PI\!\to\!B$ & $0.55\text{--}0.63$ & $-36.4 [-50.2, -24.7]$ & $-191.1 [-219.0, -164.9]$ & $+154.7 [+122.0, +185.7]$ \\
$PB\!\to\!P$ & $0.46\text{--}0.62$ & $-15.0 [-25.3, -6.0]$ & $-34.6 [-52.1, -21.1]$ & $+19.7 [+1.0, +39.2]$ \\
$PB\!\to\!I$ & $0.53\text{--}0.67$ & $-19.1 [-28.1, -11.0]$ & $-109.1 [-134.4, -86.1]$ & $+89.9 [+64.5, +118.5]$ \\
$PB\!\to\!B$ & $0.44\text{--}0.54$ & $-26.7 [-34.4, -19.5]$ & $-123.2 [-142.8, -105.6]$ & $+96.6 [+75.7, +118.7]$ \\
$IB\!\to\!P$ & $0.71\text{--}0.83$ & $-54.9 [-73.2, -37.1]$ & $-222.1 [-260.1, -184.9]$ & $+167.2 [+115.3, +215.0]$ \\
$IB\!\to\!I$ & $0.46\text{--}0.53$ & $-25.5 [-35.3, -16.8]$ & $-66.3 [-82.8, -51.8]$ & $+40.8 [+22.0, +60.2]$ \\
$IB\!\to\!B$ & $0.42\text{--}0.47$ & $-38.1 [-50.2, -27.3]$ & $-146.9 [-173.7, -122.5]$ & $+108.8 [+78.1, +140.9]$ \\
$PIB\!\to\!P$ & $0.46\text{--}0.63$ & $-13.6 [-22.7, -5.8]$ & $-42.3 [-66.9, -22.2]$ & $+28.7 [+4.9, +56.0]$ \\
$PIB\!\to\!I$ & $0.44\text{--}0.54$ & $-21.6 [-31.5, -12.9]$ & $-65.9 [-83.1, -49.6]$ & $+44.3 [+23.9, +65.7]$ \\
$PIB\!\to\!B$ & $0.41\text{--}0.52$ & $-40.7 [-51.6, -30.4]$ & $-134.0 [-155.7, -113.6]$ & $+93.3 [+68.7, +119.2]$ \\
\bottomrule
\end{tabular}
\caption{Common-error support under the same six-seed models, $10^{-3}$ nat. Coverage is the range over the six seeds. Brackets: conditional 95\% paired-bootstrap intervals.}
\label{tab:clinical_common}
\end{table*}

\begin{table*}[t]
\centering
\small
\setlength{\tabcolsep}{4pt}
\begin{tabular}{@{}lrrrrrrc@{}}
\toprule
Train $\to$ evaluate & 2026 & 2027 & 2028 & 2029 & 2030 & 2031 & Seeds and units \\
\midrule
$P\!\to\!P$ & $-1.6$ & $+22.0$ & $+4.8$ & $+11.4$ & $-6.9$ & $+25.4$ &  \\
$P\!\to\!I$ & $+145.7$ & $+270.4$ & $+271.0$ & $-78.7$ & $-138.2$ & $+370.6$ &  \\
$P\!\to\!B$ & $+160.3$ & $+277.6$ & $+195.9$ & $-338.6$ & $-56.5$ & $+252.7$ &  \\
$I\!\to\!P$ & $+213.7$ & $+250.9$ & $+96.1$ & $+144.8$ & $+58.2$ & $+58.3$ &  \\
$I\!\to\!I$ & $+9.1$ & $+50.9$ & $+11.6$ & $+29.0$ & $+3.9$ & $+8.1$ & [$+0.7$, $+38.9$] \\
$I\!\to\!B$ & $+94.6$ & $+91.6$ & $+52.5$ & $+35.0$ & $+9.4$ & $+68.2$ & [$+30.3$, $+87.6$] \\
$B\!\to\!P$ & $+59.7$ & $+222.6$ & $+73.0$ & $-5.7$ & $+118.7$ & $+219.9$ &  \\
$B\!\to\!I$ & $+32.3$ & $+33.8$ & $+52.4$ & $+50.3$ & $+18.7$ & $+21.9$ & [$+11.5$, $+58.2$] \\
$B\!\to\!B$ & $+2.0$ & $+46.2$ & $+49.9$ & $+31.1$ & $+32.7$ & $+42.4$ & [$+12.6$, $+54.9$] \\
$PI\!\to\!P$ & $+26.6$ & $+14.0$ & $-2.3$ & $-13.2$ & $+24.0$ & $+32.9$ &  \\
$PI\!\to\!I$ & $+61.3$ & $+14.9$ & $+39.5$ & $+10.5$ & $+5.7$ & $+37.2$ &  \\
$PI\!\to\!B$ & $+88.9$ & $+93.9$ & $+96.7$ & $+135.9$ & $+117.6$ & $+71.9$ &  \\
$PB\!\to\!P$ & $+22.4$ & $+16.7$ & $+75.5$ & $+22.2$ & $+11.8$ & $+27.4$ &  \\
$PB\!\to\!I$ & $+96.8$ & $+96.5$ & $+26.1$ & $+52.7$ & $-30.5$ & $+62.1$ &  \\
$PB\!\to\!B$ & $+44.0$ & $+91.9$ & $+29.2$ & $+52.6$ & $-10.8$ & $+47.7$ &  \\
$IB\!\to\!P$ & $+18.0$ & $+33.8$ & $+390.5$ & $+118.3$ & $+341.3$ & $+169.2$ &  \\
$IB\!\to\!I$ & $+12.9$ & $+37.0$ & $+13.0$ & $+28.3$ & $+42.1$ & $+13.9$ & [$+6.6$, $+43.9$] \\
$IB\!\to\!B$ & $+79.0$ & $+81.9$ & $+70.5$ & $+38.8$ & $+70.5$ & $+40.0$ & [$+39.9$, $+88.2$] \\
$PIB\!\to\!P$ & $+28.8$ & $+23.3$ & $+32.9$ & $+16.9$ & $-4.2$ & $+35.3$ &  \\
$PIB\!\to\!I$ & $+37.1$ & $+7.5$ & $+36.0$ & $+37.7$ & $+17.5$ & $+26.1$ &  \\
$PIB\!\to\!B$ & $+69.2$ & $+61.1$ & $+25.6$ & $+63.2$ & $+34.9$ & $+47.7$ &  \\
\bottomrule
\end{tabular}
\caption{Seed-level corrected-NLL point contrasts on all partial masks, $10^{-3}$ nat. Each seed pools its four evaluation folds; the six seeds reuse the same evaluation images. Seeds and units: 95\% interval with the six seeds also resampled, conditions without Private250.}
\label{tab:clinical_seeds}
\end{table*}

\begin{table*}[t]
\centering
\small
\setlength{\tabcolsep}{4pt}
\begin{tabular}{@{}lccc@{}}
\toprule
Train $\to$ evaluate & $\Hm$: Mean / Worst & $\Gm$: Mean / Worst & AUROC: Mean / Worst \\
\midrule
$P\!\to\!P$ & $21.6 / 13.5$ & $42.7 / 25.4$ & $0.965 / 0.975$ \\
$P\!\to\!I$ & $74.5 / 55.2$ & $608.2 / 448.8$ & $0.526 / 0.499$ \\
$P\!\to\!B$ & $85.8 / 63.6$ & $510.1 / 406.0$ & $0.537 / 0.527$ \\
$I\!\to\!P$ & $110.2 / 75.3$ & $402.0 / 230.2$ & $0.507 / 0.557$ \\
$I\!\to\!I$ & $55.3 / 35.1$ & $112.9 / 73.9$ & $0.869 / 0.885$ \\
$I\!\to\!B$ & $95.8 / 66.8$ & $248.9 / 161.4$ & $0.629 / 0.639$ \\
$B\!\to\!P$ & $118.8 / 76.0$ & $450.3 / 292.8$ & $0.564 / 0.531$ \\
$B\!\to\!I$ & $108.0 / 73.2$ & $226.4 / 156.6$ & $0.727 / 0.739$ \\
$B\!\to\!B$ & $101.3 / 62.2$ & $218.7 / 145.6$ & $0.714 / 0.728$ \\
$PI\!\to\!P$ & $37.1 / 28.2$ & $55.7 / 33.1$ & $0.961 / 0.969$ \\
$PI\!\to\!I$ & $65.9 / 43.3$ & $131.0 / 80.2$ & $0.864 / 0.881$ \\
$PI\!\to\!B$ & $131.5 / 81.1$ & $397.0 / 245.8$ & $0.603 / 0.622$ \\
$PB\!\to\!P$ & $41.0 / 30.4$ & $108.1 / 68.1$ & $0.945 / 0.969$ \\
$PB\!\to\!I$ & $116.3 / 73.4$ & $249.8 / 156.2$ & $0.730 / 0.736$ \\
$PB\!\to\!B$ & $106.2 / 64.1$ & $234.7 / 150.1$ & $0.731 / 0.740$ \\
$IB\!\to\!P$ & $148.5 / 78.2$ & $428.0 / 179.1$ & $0.592 / 0.540$ \\
$IB\!\to\!I$ & $66.6 / 38.0$ & $115.6 / 62.5$ & $0.883 / 0.894$ \\
$IB\!\to\!B$ & $102.2 / 60.3$ & $228.8 / 123.4$ & $0.747 / 0.779$ \\
$PIB\!\to\!P$ & $34.3 / 19.2$ & $87.2 / 49.9$ & $0.961 / 0.972$ \\
$PIB\!\to\!I$ & $63.4 / 36.4$ & $119.8 / 65.8$ & $0.881 / 0.894$ \\
$PIB\!\to\!B$ & $102.2 / 58.1$ & $211.6 / 117.2$ & $0.762 / 0.784$ \\
\bottomrule
\end{tabular}
\caption{Absolute correction outcomes at the fixed CorrectiveMean start, and task AUROC from each model's own uncorrected predictions. Harm and useful gain are in $10^{-3}$ nat; AUROC is unitless. Values average the six-seed cohort.}
\label{tab:clinical_levels}
\end{table*}

\section{Symmetric amplitude calibration of the clinical comparison}
\label{sec:supp_clinical_frontier}

\paragraph{Setup.}
\Cref{tab:ultrasound_gain} keeps each arm's native response amplitude.
To test whether CorrectiveWorst only responds less, both arms receive the same source-fitted amplitude family and are compared at matched source harm.
It uses the two public same-source conditions, B$\to$B and I$\to$I, with linear and MLP readouts, selected checkpoints, six seeds and four folds.
Public sources keep every step reproducible \arxivor{from released data}{once the annotations are released}, and two of them are enough to test this explanation.
Each arm's response to a correction set, $r(S)=Z(S)-Z(\varnothing)$, receives one global scale $\beta\ge0$ ($F_1$), fitted on the calibration role with \cref{eq:obj} and the root search of \cref{sec:audit_frontier} to the budgets \cref{eq:budget}; $\lambda=0$ gives each arm its own source-optimal scale and is reported as the unmatched control.
Events are the 62 nonempty proper field subsets per case.
The primary start is the shared CorrectiveMean start of \cref{tab:ultrasound_gain}; the shared CorrectiveWorst start and the native starts are secondary.
Intervals come from 1000 linked original-group bootstrap draws, in which each group's count applies to every role, fold, seed, readout and arm and every source-only quantity is refitted.

\begin{table*}[t]
\centering
\footnotesize
\setlength{\tabcolsep}{4pt}
\begin{tabular}{@{}lcccccc@{}}
\toprule
 & $\lambda=0$ & $f=1$ & $f=0.75$ & $f=0.5$ & $f=0.25$ & $f=0.1$ \\
\midrule
\multicolumn{7}{@{}l}{\emph{Shared CorrectiveMean start (primary)}} \\
BUS-BRA, linear & $+33.9$ & $+22.3$ & $+17.7$ & $+12.3$ & $+6.2$ & $+2.4$ \\
 & {\scriptsize$[+16.3, +48.4]$} & {\scriptsize$[+3.8, +40.1]$} & {\scriptsize$[+2.8, +35.5]$} & {\scriptsize$[+1.3, +28.4]$} & {\scriptsize$[-0.05, +17.5]$} & {\scriptsize incomplete (999)} \\\addlinespace[1.5pt]
BUS-BRA, MLP & $+4.6$ & $+3.8$ & $+2.2$ & $+0.7$ & $-0.4$ & $-0.4$ \\
 & {\scriptsize$[-2.7, +14.8]$} & {\scriptsize$[-2.7, +14.5]$} & {\scriptsize$[-2.9, +10.2]$} & {\scriptsize$[-3.3, +6.4]$} & {\scriptsize$[-3.0, +3.1]$} & {\scriptsize$[-1.8, +1.2]$} \\\addlinespace[1.5pt]
BUSI647, linear & $+25.4$ & $+13.9$ & $+11.9$ & $+8.6$ & $+4.5$ & $+1.8$ \\
 & {\scriptsize$[+7.8, +45.0]$} & {\scriptsize$[+1.5, +31.1]$} & {\scriptsize$[+1.1, +28.2]$} & {\scriptsize$[+0.4, +22.7]$} & {\scriptsize incomplete (996)} & {\scriptsize incomplete (948)} \\\addlinespace[1.5pt]
BUSI647, MLP & $+15.9$ & $+9.4$ & $+7.2$ & $+4.6$ & $+1.9$ & $+0.6$ \\
 & {\scriptsize$[+2.4, +27.4]$} & {\scriptsize$[+0.2, +23.1]$} & {\scriptsize$[+0.3, +19.5]$} & {\scriptsize$[-0.2, +14.9]$} & {\scriptsize$[-1.5, +8.5]$} & {\scriptsize incomplete (977)} \\\addlinespace[1.5pt]
\midrule
\multicolumn{7}{@{}l}{\emph{Shared CorrectiveWorst start}} \\
BUS-BRA, linear & $+4.8$ & $+4.6$ & $+3.6$ & $+2.5$ & $+1.4$ & $+0.7$ \\
 & {\scriptsize$[-8.6, +16.3]$} & {\scriptsize$[-7.2, +15.4]$} & {\scriptsize$[-5.9, +12.8]$} & {\scriptsize$[-5.3, +10.6]$} & {\scriptsize$[-4.4, +7.4]$} & {\scriptsize incomplete (983)} \\\addlinespace[1.5pt]
BUS-BRA, MLP & $-17.7$ & $-18.4$ & $-16.8$ & $-13.7$ & $-8.3$ & $-3.7$ \\
 & {\scriptsize$[-29.3, -9.2]$} & {\scriptsize$[-27.5, -4.3]$} & {\scriptsize$[-25.5, -3.7]$} & {\scriptsize$[-21.9, -2.6]$} & {\scriptsize$[-14.3, -1.2]$} & {\scriptsize incomplete (999)} \\\addlinespace[1.5pt]
BUSI647, linear & $+5.1$ & $+3.8$ & $+4.4$ & $+4.6$ & $+3.3$ & $+1.5$ \\
 & {\scriptsize$[-8.3, +17.1]$} & {\scriptsize incomplete (999)} & {\scriptsize incomplete (999)} & {\scriptsize incomplete (999)} & {\scriptsize incomplete (996)} & {\scriptsize incomplete (926)} \\\addlinespace[1.5pt]
BUSI647, MLP & $+7.9$ & $+5.1$ & $+5.2$ & $+4.8$ & $+3.3$ & $+1.6$ \\
 & {\scriptsize$[-8.4, +18.7]$} & {\scriptsize$[-6.0, +16.6]$} & {\scriptsize$[-4.3, +13.9]$} & {\scriptsize$[-2.8, +11.1]$} & {\scriptsize$[-1.0, +7.2]$} & {\scriptsize incomplete (992)} \\\addlinespace[1.5pt]
\midrule
\multicolumn{7}{@{}l}{\emph{Native starts}} \\
BUS-BRA, linear & $-0.6$ & $-6.5$ & $-12.7$ & $-19.6$ & $-26.5$ & $-30.6$ \\
 & {\scriptsize$[-34.1, +27.3]$} & {\scriptsize$[-44.1, +26.9]$} & {\scriptsize$[-48.8, +21.3]$} & {\scriptsize$[-56.0, +14.8]$} & {\scriptsize$[-65.2, +10.4]$} & {\scriptsize incomplete (980)} \\\addlinespace[1.5pt]
BUS-BRA, MLP & $+24.7$ & $+25.2$ & $+27.3$ & $+29.9$ & $+33.5$ & $+36.3$ \\
 & {\scriptsize$[+5.6, +45.1]$} & {\scriptsize$[+6.5, +48.1]$} & {\scriptsize$[+7.4, +50.3]$} & {\scriptsize$[+8.5, +54.3]$} & {\scriptsize$[+10.0, +58.2]$} & {\scriptsize incomplete (999)} \\\addlinespace[1.5pt]
BUSI647, linear & $+0.9$ & $-14.0$ & $-18.4$ & $-22.6$ & $-25.7$ & $-26.5$ \\
 & {\scriptsize$[-46.1, +39.1]$} & {\scriptsize$[-57.3, +26.7]$} & {\scriptsize$[-61.0, +23.4]$} & {\scriptsize$[-65.4, +19.3]$} & {\scriptsize incomplete (983)} & {\scriptsize incomplete (917)} \\\addlinespace[1.5pt]
BUSI647, MLP & $+56.2$ & $+58.2$ & $+56.3$ & $+53.5$ & $+49.6$ & $+46.7$ \\
 & {\scriptsize$[+21.8, +89.2]$} & {\scriptsize$[+23.1, +89.9]$} & {\scriptsize$[+21.3, +87.7]$} & {\scriptsize$[+19.5, +85.1]$} & {\scriptsize$[+16.0, +83.1]$} & {\scriptsize incomplete (988)} \\\addlinespace[1.5pt]
\bottomrule
\end{tabular}
\caption{Symmetric amplitude calibration on the public same-source conditions: CorrectiveWorst minus CorrectiveMean in corrected NLL on the evaluation role ($10^{-3}$ nat), both arms with one source-fitted global scale, at their own source-optimal scale ($\lambda=0$) and at the matched source-harm budgets $h_f=f\min(H_0^{\mathrm{Mean}},H_0^{\mathrm{Worst}})$. Seed means over six seeds of four pooled evaluation folds; 95\% percentile intervals of 1000 linked group-bootstrap draws with every source-only quantity refitted; \emph{incomplete}: unresolved budget roots in some draws (number of complete draws in parentheses), so no interval is reported.}
\label{tab:supp_clinical_frontier}
\end{table*}

\paragraph{Result.}
We classify each primary cell (shared CorrectiveMean start, $\Delta$NLL) as favoring CorrectiveMean if its interval lies above zero, CorrectiveWorst if it lies below zero, and unresolved otherwise; we call CorrectiveWorst dominated by calibrated CorrectiveMean if at least three of the four conditions favor CorrectiveMean at all five budgets and none favors CorrectiveWorst at any budget.
The outcome is mixed (\cref{tab:supp_clinical_frontier}).
CorrectiveMean has the lower corrected NLL at $f\in\{1,0.75,0.5\}$ with both linear readouts and at $f\in\{1,0.75\}$ with the BUSI647 MLP readout; the BUS-BRA MLP readout is unresolved at every budget, the smaller budgets are unresolved or incomplete, and no cell favors CorrectiveWorst.
The differences shrink with the budget as both arms approach the uncorrected start.
Matching source harm leaves the evaluation-harm differences of every primary cell inside their intervals around zero, and the NLL differences come from useful gain, \eg $\Delta\Gm=-13.7$ [$-31.5$, $-2.8$] and $-7.8$ [$-24.6$, $+1.2$]$\times10^{-3}$ nat at $f=0.5$ for the BUS-BRA and BUSI647 linear readouts.
No interval of the share of events with harm above $0.05$ nat lies below zero at the primary start; it lies above zero for the BUS-BRA MLP readout at $f\ge0.25$ (\eg $+0.55$ [$+0.15$, $+0.98$] pp at $f=0.5$) and for the BUS-BRA linear readout at $f=0.5$.
At $f=0.5$ the points are those of \cref{tab:supp_selection_mlp}; the intervals differ slightly because the draws are new.

\paragraph{Starts.}
The comparison depends on the start.
At the shared CorrectiveWorst start, the BUS-BRA MLP readout favors CorrectiveWorst at every budget with a complete interval (\eg $-13.7$ [$-21.9$, $-2.6$]$\times10^{-3}$ nat at $f=0.5$), and the other conditions are unresolved or incomplete.
With native starts, the starting risk dominates: CorrectiveWorst starts worse with the MLP readouts (before-correction NLL $+38.5$ [$+12.9$, $+64.8$] on BUS-BRA and $+44.5$ [$+9.9$, $+79.3$] on BUSI647, $10^{-3}$ nat) and better, though unresolved, with the linear readouts ($-33.4$ [$-73.1$, $+5.1$] and $-26.2$ [$-69.2$, $+14.6$]).

\paragraph{Per-field family.}
With one coefficient per field ($F_6$, linear readouts), 2.9--5.3\% of the arm-draw problems on BUS-BRA and 18.0--20.8\% on BUSI647 stay unresolved across the five budgets, mostly because the root search finds no certified solution; no budget therefore has a complete point or interval, and we report none.
At each arm's own source-optimal per-field coefficients ($\lambda=0$), CorrectiveWorst has the higher corrected NLL on both sources, $+43.3$ and $+57.8\times10^{-3}$ nat, positive in all six seeds, with lower useful gain ($-46.1$ and $-53.5$); 996 and 998 of the 1000 draws are complete.

\paragraph{Zero and unreachable budgets.}
A budget is zero when one arm's response lowers the source NLL at no positive scale; both arms then stay uncorrected and the draw contributes no difference.
At the primary start this holds in 2,502 (BUSI647 linear) and 2,294 (BUSI647 MLP) of 24,000 model-draws, whose calibration roles have 70 groups per fold, and in 436 and 0 on BUS-BRA, with 113 groups per fold; it pulls the BUSI647 contrasts toward zero.
The incomplete cells have budgets below the source harm the root search reaches at $\lambda=16$ (\cref{sec:supp_root}), in 1--94 model-draws per cell at the primary start; one secondary cell also has a root that missed its tolerance.

\paragraph{Verification.}
A separate NumPy/SciPy implementation recomputes the primary cells for all models in four draws: all 4,608 resolution states agree, every root meets its budget tolerance, the evaluation sums agree to a relative $1.6\times10^{-15}$ and the point contrasts to $1.9\times10^{-16}$ nat, and with independently solved roots the $\Delta$NLL points differ by at most $3.7\times10^{-7}$ nat, within the root tolerance; the points of \cref{tab:supp_selection_mlp} are reproduced to at most $2.4\times10^{-11}$ nat.

\section{Intervention-aware training under the same calibration}
\label{sec:supp_intcem}

\paragraph{Setup.}
To test whether training for interventions changes the comparison, we train the official intervention-aware concept embedding model (IntCEM)~\cite{espinosa2023learning} and a matched concept embedding model (CEM) on the BUS-BRA development pool, with the folds, roles and recipe of \cref{sec:supp_public_recipe} and six seeds per fold (48 fits).
Both models share the encoder, 16-dimensional concept embeddings, the linear readout, training-time random interventions ($p=0.25$) and separate probability generators per concept; IntCEM adds its learned intervention policy over the six fields and its rollout loss at the package defaults.
Each checkpoint is selected by the model's own validation loss, at epochs 4--39 for CEM (median 19) and 14--99 for IntCEM (median 51.5).
The analysis is that of \cref{sec:supp_clinical_frontier}: one source-fitted global scale per model at the matched source-harm budgets and a linked group bootstrap of 1000 draws.
The primary contrast is IntCEM minus CEM in corrected NLL at the shared CEM start and $f=0.5$.
On the CUB setup of \cref{sec:res_native}, with the 10 concept sets and three seeds of \cref{tab:supp_strength} and the training recipe of the jointly trained hybrids, we compare the same two models on the frozen image features (60 fits); IntCEM's policy acts on the 50 concepts, and checkpoints are selected at epochs 9--13 for CEM (median 11) and 12--30 for IntCEM (median 16.5).
Its analysis is the matched-source-harm comparison of \cref{sec:res_native_frontier} with one global scale per model: the 50 single-concept corrections of every image, fitted on the validation split and evaluated on the test split, with the 1000 image-bootstrap draws of that comparison; its primary contrast is the same.

\begin{table*}[t]
\centering
\footnotesize
\setlength{\tabcolsep}{4pt}
\begin{tabular}{@{}lcccccc@{}}
\toprule
 & $\lambda=0$ & $f=1$ & $f=0.75$ & $f=0.5$ & $f=0.25$ & $f=0.1$ \\
\midrule
\multicolumn{7}{@{}l}{\emph{Shared CEM start (primary)}} \\
$\Delta$NLL & $+84.0$ & $+59.7$ & $+53.2$ & $+42.7$ & $+25.5$ & $+11.2$ \\
 & {\scriptsize$[+49.4, +122.7]$} & {\scriptsize$[+27.6, +93.6]$} & {\scriptsize$[+26.2, +84.4]$} & {\scriptsize$[+20.8, +68.0]$} & {\scriptsize incomplete (998)} & {\scriptsize incomplete (989)} \\\addlinespace[1.5pt]
$\Delta\Gm$ & $-92.4$ & $-54.4$ & $-49.4$ & $-40.5$ & $-25.0$ & $-11.3$ \\
 & {\scriptsize$[-124.7, -49.9]$} & {\scriptsize$[-78.3, -25.9]$} & {\scriptsize$[-72.3, -23.1]$} & {\scriptsize$[-62.1, -18.7]$} & {\scriptsize incomplete (998)} & {\scriptsize incomplete (989)} \\\addlinespace[1.5pt]
$\Delta\Hm$ & $-8.5$ & $+5.3$ & $+3.8$ & $+2.2$ & $+0.4$ & $+0.0$ \\
 & {\scriptsize$[-34.5, +42.8]$} & {\scriptsize$[-15.5, +35.4]$} & {\scriptsize$[-10.7, +27.5]$} & {\scriptsize$[-6.7, +18.8]$} & {\scriptsize incomplete (998)} & {\scriptsize incomplete (989)} \\\addlinespace[1.5pt]
\midrule
\multicolumn{7}{@{}l}{\emph{Shared IntCEM start}} \\
$\Delta$NLL & $-89.4$ & $-70.9$ & $-61.6$ & $-49.2$ & $-30.4$ & $-13.8$ \\
 & {\scriptsize$[-118.5, -63.8]$} & {\scriptsize$[-97.9, -44.1]$} & {\scriptsize$[-84.8, -38.7]$} & {\scriptsize$[-70.2, -31.0]$} & {\scriptsize$[-46.1, -18.5]$} & {\scriptsize$[-22.5, -8.1]$} \\\addlinespace[1.5pt]
$\Delta\Gm$ & $+110.9$ & $+66.8$ & $+59.0$ & $+47.6$ & $+29.7$ & $+13.6$ \\
 & {\scriptsize$[+88.4, +134.0]$} & {\scriptsize$[+42.7, +90.2]$} & {\scriptsize$[+37.2, +81.5]$} & {\scriptsize$[+29.0, +68.4]$} & {\scriptsize$[+17.3, +46.4]$} & {\scriptsize$[+7.6, +22.5]$} \\\addlinespace[1.5pt]
$\Delta\Hm$ & $+21.6$ & $-4.1$ & $-2.7$ & $-1.6$ & $-0.7$ & $-0.3$ \\
 & {\scriptsize$[-2.8, +41.1]$} & {\scriptsize$[-24.0, +12.4]$} & {\scriptsize$[-17.2, +9.5]$} & {\scriptsize$[-10.7, +6.2]$} & {\scriptsize$[-4.8, +2.7]$} & {\scriptsize$[-1.7, +1.1]$} \\\addlinespace[1.5pt]
\midrule
\multicolumn{7}{@{}l}{\emph{Native starts}} \\
$\Delta$NLL & $-309.7$ & $-306.6$ & $-310.6$ & $-314.2$ & $-313.7$ & $-307.7$ \\
 & {\scriptsize$[-411.4, -224.1]$} & {\scriptsize$[-407.1, -216.4]$} & {\scriptsize$[-414.3, -218.6]$} & {\scriptsize$[-417.8, -219.4]$} & {\scriptsize$[-419.7, -215.6]$} & {\scriptsize incomplete (977)} \\\addlinespace[1.5pt]
$\Delta\Gm$ & $-8.5$ & $+10.4$ & $+13.3$ & $+15.6$ & $+14.0$ & $+7.8$ \\
 & {\scriptsize$[-55.0, +44.5]$} & {\scriptsize$[-26.8, +41.0]$} & {\scriptsize$[-20.3, +39.9]$} & {\scriptsize$[-11.7, +36.3]$} & {\scriptsize$[-3.5, +27.2]$} & {\scriptsize incomplete (977)} \\\addlinespace[1.5pt]
$\Delta\Hm$ & $-18.5$ & $+3.5$ & $+2.4$ & $+1.0$ & $-0.1$ & $-0.2$ \\
 & {\scriptsize$[-49.1, +8.0]$} & {\scriptsize$[-13.4, +17.1]$} & {\scriptsize$[-10.5, +12.7]$} & {\scriptsize$[-8.2, +8.1]$} & {\scriptsize$[-5.0, +3.6]$} & {\scriptsize incomplete (977)} \\\addlinespace[1.5pt]
\bottomrule
\end{tabular}
\caption{Intervention-aware training on the BUS-BRA development pool: IntCEM minus a matched CEM ($10^{-3}$ nat), linear readout, both arms with one source-fitted global scale at their own source-optimal scale ($\lambda=0$) and at the matched source-harm budgets. Seed means over six seeds of four pooled evaluation folds; 95\% percentile intervals of 1000 linked group-bootstrap draws with every source-only quantity refitted; \emph{incomplete}: unresolved budget roots in some draws (number of complete draws in parentheses).}
\label{tab:supp_intcem}
\end{table*}

\paragraph{Result on BUS-BRA.}
At the primary start the CEM has the lower corrected NLL at every budget with a complete interval (\cref{tab:supp_intcem}; $+42.7$ [$+20.8$, $+68.0$]$\times10^{-3}$ nat at $f=0.5$, positive in all six seeds), because the IntCEM responses add less useful gain on the CEM start ($\Delta\Gm=-40.5$ [$-62.1$, $-18.7$]).
The contrast therefore favors the CEM.
At the shared IntCEM start the ordering reverses at every budget ($-49.2$ [$-70.2$, $-31.0$] at $f=0.5$).
With native starts IntCEM has a far lower corrected NLL ($-314.2$ [$-417.8$, $-219.4$] at $f=0.5$), almost all of it present before correction ($-299.7$ [$-411.2$, $-196.5$]): the uncorrected CEM predictions have an evaluation NLL of $0.98$, against $0.68$ for IntCEM.
The ordering thus depends on the shared start, as in \cref{sec:supp_clinical_frontier}, and the primary contrast gives no evidence that intervention-aware training expands the useful-correction frontier at a fixed start; with native starts, IntCEM has the lower corrected NLL, most of it present before correction.

\begin{table*}[t]
\centering
\footnotesize
\setlength{\tabcolsep}{4pt}
\begin{tabular}{@{}lcccccc@{}}
\toprule
 & $\lambda=0$ & $f=1$ & $f=0.75$ & $f=0.5$ & $f=0.25$ & $f=0.1$ \\
\midrule
\multicolumn{7}{@{}l}{\emph{Shared CEM start (primary)}} \\
$\Delta$NLL & $+0.26$ & $+0.27$ & $+0.24$ & $+0.18$ & $+0.11$ & $+0.05$ \\
 & {\scriptsize$[+0.05, +0.47]$} & {\scriptsize$[+0.07, +0.42]$} & {\scriptsize$[+0.09, +0.35]$} & {\scriptsize$[+0.08, +0.26]$} & {\scriptsize$[+0.05, +0.15]$} & {\scriptsize$[+0.02, +0.06]$} \\\addlinespace[1.5pt]
$\Delta\Gm$ & $+0.54$ & $-0.30$ & $-0.27$ & $-0.20$ & $-0.12$ & $-0.05$ \\
 & {\scriptsize$[-1.83, +3.07]$} & {\scriptsize$[-0.51, -0.06]$} & {\scriptsize$[-0.43, -0.08]$} & {\scriptsize$[-0.32, -0.07]$} & {\scriptsize$[-0.18, -0.05]$} & {\scriptsize$[-0.08, -0.02]$} \\\addlinespace[1.5pt]
$\Delta\Hm$ & $+0.80$ & $-0.04$ & $-0.03$ & $-0.02$ & $-0.01$ & $+0.00$ \\
 & {\scriptsize$[-1.58, +3.25]$} & {\scriptsize$[-0.26, +0.18]$} & {\scriptsize$[-0.20, +0.13]$} & {\scriptsize$[-0.13, +0.09]$} & {\scriptsize$[-0.07, +0.04]$} & {\scriptsize$[-0.03, +0.02]$} \\\addlinespace[1.5pt]
\midrule
\multicolumn{7}{@{}l}{\emph{Shared IntCEM start}} \\
$\Delta$NLL & $-1.54$ & $-0.89$ & $-0.63$ & $-0.39$ & $-0.18$ & $-0.07$ \\
 & {\scriptsize$[-1.87, -1.21]$} & {\scriptsize$[-1.16, -0.64]$} & {\scriptsize$[-0.82, -0.45]$} & {\scriptsize$[-0.52, -0.28]$} & {\scriptsize$[-0.24, -0.13]$} & {\scriptsize$[-0.09, -0.05]$} \\\addlinespace[1.5pt]
$\Delta\Gm$ & $+16.45$ & $+0.79$ & $+0.56$ & $+0.34$ & $+0.16$ & $+0.06$ \\
 & {\scriptsize$[+14.31, +18.81]$} & {\scriptsize$[+0.53, +1.04]$} & {\scriptsize$[+0.35, +0.74]$} & {\scriptsize$[+0.20, +0.47]$} & {\scriptsize$[+0.08, +0.22]$} & {\scriptsize$[+0.03, +0.08]$} \\\addlinespace[1.5pt]
$\Delta\Hm$ & $+14.91$ & $-0.09$ & $-0.07$ & $-0.05$ & $-0.02$ & $-0.01$ \\
 & {\scriptsize$[+12.73, +17.28]$} & {\scriptsize$[-0.40, +0.13]$} & {\scriptsize$[-0.30, +0.10]$} & {\scriptsize$[-0.20, +0.07]$} & {\scriptsize$[-0.10, +0.03]$} & {\scriptsize$[-0.04, +0.01]$} \\\addlinespace[1.5pt]
\midrule
\multicolumn{7}{@{}l}{\emph{Native starts}} \\
$\Delta$NLL & $+150.94$ & $+151.64$ & $+151.90$ & $+152.14$ & $+152.36$ & $+152.49$ \\
 & {\scriptsize$[+42.20, +261.41]$} & {\scriptsize$[+42.51, +262.08]$} & {\scriptsize$[+42.78, +262.39]$} & {\scriptsize$[+43.03, +262.69]$} & {\scriptsize$[+43.27, +262.96]$} & {\scriptsize$[+43.40, +263.11]$} \\\addlinespace[1.5pt]
$\Delta\Gm$ & $+16.92$ & $+0.85$ & $+0.61$ & $+0.39$ & $+0.18$ & $+0.07$ \\
 & {\scriptsize$[+14.62, +19.45]$} & {\scriptsize$[+0.44, +1.19]$} & {\scriptsize$[+0.30, +0.86]$} & {\scriptsize$[+0.18, +0.55]$} & {\scriptsize$[+0.07, +0.26]$} & {\scriptsize$[+0.03, +0.10]$} \\\addlinespace[1.5pt]
$\Delta\Hm$ & $+15.30$ & $-0.07$ & $-0.06$ & $-0.04$ & $-0.02$ & $-0.01$ \\
 & {\scriptsize$[+12.94, +17.87]$} & {\scriptsize$[-0.44, +0.20]$} & {\scriptsize$[-0.33, +0.15]$} & {\scriptsize$[-0.22, +0.10]$} & {\scriptsize$[-0.11, +0.05]$} & {\scriptsize$[-0.05, +0.02]$} \\\addlinespace[1.5pt]
\bottomrule
\end{tabular}
\caption{Intervention-aware training on CUB: IntCEM minus a matched CEM ($10^{-4}$ nat) on the frozen image features of \cref{sec:res_native}, both arms with one source-fitted global scale at their own source-optimal scale ($\lambda=0$) and at the matched source-harm budgets, over the 50 single-concept corrections of every test image. Means over 30 units (10 concept sets, three seeds); 95\% percentile intervals of the 1000 image-bootstrap draws of the CUB amplitude analysis, conditional on the trained models, with every source-only quantity refitted; every budget root resolves in every draw.}
\label{tab:supp_intcem_cub}
\end{table*}

\paragraph{Result on CUB.}
The primary contrast is again positive at every budget (\cref{tab:supp_intcem_cub}; $+0.18$ [$+0.08$, $+0.26$]$\times10^{-4}$ nat at $f=0.5$, positive in 28 of 30 units), again because the IntCEM responses add less useful gain on the CEM start ($\Delta\Gm=-0.20$ [$-0.32$, $-0.07$]$\times10^{-4}$).
The difference is small, about 3\% of the useful gain at this budget and far below the margin $\delta$.
At the shared IntCEM start the sign reverses at every budget; at both starts every matched contrast stays inside $\pm\delta$, so on CUB the start changes a sign, not a practical ordering.
With native starts IntCEM has the higher corrected NLL ($+152.1$ [$+43.0$, $+262.7$]$\times10^{-4}$ nat at $f=0.5$), all of it present before correction ($+152.6$ [$+43.5$, $+263.2$]): the uncorrected test NLL is $0.662$ for IntCEM and $0.646$ for the CEM.
On the path metrics at native amplitude, the harmful-path prevalence of IntCEM is lower by $12.6$ percentage points [$4.6$, $14.9$] and its maximum sampled partial harm shows no detectable change ($+0.0003$ [$-0.0003$, $+0.0021$]), while its partially corrected NLL is higher ($+0.012$ [$+0.002$, $+0.023$]), as for the monotonic penalty of \cref{sec:res_native}.
On BUS-BRA the primary contrast favors the CEM and reverses with the shared start; on CUB every matched contrast stays inside $\pm\delta$. On both datasets the native difference is almost entirely starting risk, whose sign follows the uncorrected predictions.

\paragraph{Verification.}
Readout parity is exact for all 48 BUS-BRA models, and the separate implementation of \cref{sec:supp_clinical_frontier} reproduces their point contrasts to $7.6\times10^{-17}$ nat.
On CUB, all 60 response maps pass the parity and additivity checks (largest gap $3.8\times10^{-13}$), and a separate NumPy implementation reproduces the matched scales to a relative $8.0\times10^{-5}$, within the budget tolerance, and the evaluation NLL, harm and gain to $10^{-7}$ nat.

\section{Finite-query model selection}
\label{sec:supp_selection}

These results complete \cref{sec:res_selection} and \cref{tab:selection}.

\paragraph{Setup.}
On each of the three clinical sources, the candidates are the CorrectiveMean and CorrectiveWorst models of one readout, fold and seed, trained on that source alone; four folds and six seeds give 24 fold--seed units per source and readout.
The linear-readout models are those of \cref{tab:ultrasound_gain}.
The MLP readout replaces the linear task head (770 parameters) by one hidden layer of 128 ReLU units (49,538 parameters); encoder, concept embeddings, objective, training recipe, initialization of the shared layers, sample order and checkpoint-selection rule follow the paired linear model.
Both models are placed on the zero-correction logit of CorrectiveMean, and each adds its own response to a correction set, as in \cref{tab:ultrasound_gain}.
A query sets one concept field to its annotated value; the multi-label margin field is one query, and queries also count on fields that are already correct.
The shared query order is fitted per fold and seed on the calibration role by greedy forward selection, minimizing the mean corrected NLL of the two models, and both models receive the first $k=2$ fields of it for every case.
On the calibration role, the harm criterion picks the model with the lower harm $\Hm$ and the NLL criterion the model with the lower corrected NLL; ties within $10^{-12}$ go to CorrectiveMean.
The picked models are evaluated on the evaluation role, pooled over the four evaluation folds per seed and averaged over the six seeds with equal weight.
Rescued (broken) cases are wrong (right) before correction and right (wrong) after it, as fractions of all evaluation cases.
Intervals use 1000 bootstrap draws of original groups within evaluation folds; each group keeps one count in its calibration and evaluation roles across folds, seeds, readouts and both models, and every draw refits the query order and repeats both selections.
They condition on the trained models.

\begin{table*}[t]
\centering
\footnotesize
\setlength{\tabcolsep}{4pt}
\begin{tabular}{@{}lllcccccc@{}}
\toprule
Source & Readout & Checkpoint & $\Delta$NLL & $\Delta\Hm$ & $\Delta\Gm$ & $\Delta$Rescued (pp) & $\Delta$Broken (pp) & Worst picked \\
\midrule
BUS-BRA & linear & selected & $+42.9$ & $-37.8$ & $-80.7$ & $-2.64$ & $-0.71$ & 22 / 4 \\
 & & & {\scriptsize$[+26.6, +64.0]$} & {\scriptsize$[-50.1, -27.7]$} & {\scriptsize$[-102.0, -62.4]$} & {\scriptsize$[-3.48, -1.87]$} & {\scriptsize$[-1.23, -0.43]$} & \\\addlinespace[1.5pt]
BUS-BRA & MLP & selected & $+26.7$ & $-22.3$ & $-49.1$ & $-3.37$ & $-1.34$ & 20 / 4 \\
 & & & {\scriptsize$[+16.9, +33.4]$} & {\scriptsize$[-24.3, -12.8]$} & {\scriptsize$[-51.8, -35.5]$} & {\scriptsize$[-3.62, -1.82]$} & {\scriptsize$[-1.57, -0.46]$} & \\\addlinespace[1.5pt]
BUSI647 & linear & selected & $+15.0$ & $-10.0$ & $-25.0$ & $-0.77$ & $-0.29$ & 16 / 2 \\
 & & & {\scriptsize$[+2.3, +23.7]$} & {\scriptsize$[-19.6, -5.0]$} & {\scriptsize$[-34.8, -14.5]$} & {\scriptsize$[-1.33, -0.26]$} & {\scriptsize$[-0.55, +0.07]$} & \\\addlinespace[1.5pt]
BUSI647 & MLP & selected & $+18.6$ & $-7.4$ & $-25.9$ & $-0.70$ & $-0.15$ & 20 / 2 \\
 & & & {\scriptsize$[+7.6, +27.1]$} & {\scriptsize$[-12.5, -4.2]$} & {\scriptsize$[-35.0, -14.5]$} & {\scriptsize$[-1.33, -0.29]$} & {\scriptsize$[-0.48, -0.04]$} & \\\addlinespace[1.5pt]
BUSI647 & linear & last & $+20.6$ & $-10.4$ & $-30.9$ & $-0.41$ & $-0.26$ & 17 / 0 \\
 & & & {\scriptsize$[+5.1, +32.8]$} & {\scriptsize$[-16.0, -3.3]$} & {\scriptsize$[-41.9, -13.6]$} & {\scriptsize$[-0.81, -0.15]$} & {\scriptsize$[-0.48, +0.04]$} & \\\addlinespace[1.5pt]
BUSI647 & MLP & last & $+21.5$ & $-6.5$ & $-28.0$ & $-0.41$ & $-0.07$ & 15 / 1 \\
 & & & {\scriptsize$[+6.2, +31.8]$} & {\scriptsize$[-11.5, -0.93]$} & {\scriptsize$[-37.2, -11.4]$} & {\scriptsize$[-0.77, -0.11]$} & {\scriptsize$[-0.33, +0.07]$} & \\
\bottomrule
\end{tabular}
\caption{Harm-selected minus NLL-selected model at the shared start with the shared query order, $k=2$ ($10^{-3}$ nat; rescued and broken: percentage points of all evaluation cases), 95\% group-level bootstrap intervals.
Worst picked: units, of 24, in which the harm and the NLL criterion pick CorrectiveWorst.
The last checkpoint is a sensitivity analysis.}
\label{tab:supp_selection_full}
\end{table*}

\paragraph{Full contrasts.}
\Cref{tab:supp_selection_full} adds the broken cases, the criterion picks and the last checkpoints.
The harm criterion picks CorrectiveWorst in 15--22 of the 24 units, the NLL criterion in 0--4.
The harm-selected model also breaks fewer predictions, but it rescues fewer still.
Not every seed moves every quantity: at the selected checkpoint on BUSI647, the linear $\Delta\Hm$ is negative in five of six seeds and the rescued difference is zero in one seed for each readout.
At the last checkpoint the BUSI647 contrasts keep their direction.

\begin{table*}[t]
\centering
\footnotesize
\setlength{\tabcolsep}{6pt}
\begin{tabular}{@{}lcccc@{}}
\toprule
 & \multicolumn{2}{c}{BUS-BRA} & \multicolumn{2}{c}{BUSI647} \\
\cmidrule(lr){2-3}\cmidrule(l){4-5}
Start, query order & linear & MLP & linear & MLP \\
\midrule
Own start, own uncertainty (planned primary) & $+60.0$ & $+23.8$ & $+8.6$ & $+31.4$ \\
 & {\scriptsize$[+13.4, +92.8]$} & {\scriptsize$[+3.9, +42.2]$} & {\scriptsize$[-22.0, +55.6]$} & {\scriptsize$[-0.42, +69.0]$} \\\addlinespace[1.5pt]
\quad of which before correction & $+54.9$ & $+19.0$ & $-4.3$ & $+24.2$ \\
 & {\scriptsize$[+7.9, +85.1]$} & {\scriptsize$[-1.4, +39.3]$} & {\scriptsize$[-31.8, +45.3]$} & {\scriptsize$[-6.2, +63.7]$} \\\addlinespace[1.5pt]
Shared start, own uncertainty & $+32.4$ & $+10.1$ & $+10.2$ & $+12.9$ \\
 & {\scriptsize$[+11.4, +40.0]$} & {\scriptsize$[+4.8, +13.7]$} & {\scriptsize$[+1.1, +20.8]$} & {\scriptsize$[+3.4, +17.4]$} \\\addlinespace[1.5pt]
Shared start, shared order (\cref{tab:selection}) & $+42.9$ & $+26.7$ & $+15.0$ & $+18.6$ \\
 & {\scriptsize$[+26.6, +64.0]$} & {\scriptsize$[+16.9, +33.4]$} & {\scriptsize$[+2.3, +23.7]$} & {\scriptsize$[+7.6, +27.1]$} \\\addlinespace[1.5pt]
Always CorrectiveMean $-$ NLL-selected, shared order & -- & -- & $+0.89$ & $+0.75$ \\
 &  &  & {\scriptsize$[-4.0, +4.9]$} & {\scriptsize$[-1.3, +6.0]$} \\
\bottomrule
\end{tabular}
\caption{Selection contrasts at $k=2$ under different starts and query orders, harm-selected minus NLL-selected unless stated ($\Delta$NLL, $10^{-3}$ nat), 95\% group-level bootstrap intervals.
Own uncertainty: each model queries its most uncertain fields first.
The before-correction row is the part of the first row present before any correction.
--: not evaluated.}
\label{tab:supp_selection_controls}
\end{table*}

\paragraph{Starts and query orders.}
The primary contrast of this analysis uses each model's own start and its own uncertainty order (\cref{tab:supp_selection_controls}).
On BUS-BRA it is positive for both readouts, but most of it is present before any correction: $+54.9$ of $+60.0\times10^{-3}$ nat for the linear readout and $+19.0$ of $+23.8$ for the MLP readout, whose before-correction interval includes zero.
On BUSI647 it is unresolved for both readouts.
With the start shared, the own uncertainty orders still give positive contrasts, but the two models then query different fields; the shared order removes this difference.
On BUSI647, the NLL-selected model is not shown to differ from always choosing CorrectiveMean, so the contrasts measure a cost of the harm criterion, not a better rule.

\begin{table*}[t]
\centering
\footnotesize
\setlength{\tabcolsep}{6pt}
\begin{tabular}{@{}llcccc@{}}
\toprule
 & & \multicolumn{2}{c}{Unscaled $\Delta$NLL} & \multicolumn{2}{c}{Matched source harm, $f=0.5$} \\
\cmidrule(lr){3-4}\cmidrule(l){5-6}
Source & Readout & selected & last & $\Delta$NLL & $\Delta\Hm$ \\
\midrule
BUS-BRA & linear & \cref{tab:ultrasound_gain} & $+102.7$ & $+12.3$ & $-1.4$ \\
 & &  & {\scriptsize$[+81.0, +129.2]$} & {\scriptsize$[+1.4, +26.1]$} & {\scriptsize$[-7.1, +3.4]$} \\\addlinespace[1.5pt]
 & MLP & $+11.7$ & $+82.3$ & $+0.69$ & $+0.66$ \\
 & & {\scriptsize$[+7.0, +16.4]$} & {\scriptsize$[+65.4, +102.0]$} & {\scriptsize$[-3.4, +6.2]$} & {\scriptsize$[-2.8, +5.2]$} \\\addlinespace[1.5pt]
BUSI647 & linear & \cref{tab:ultrasound_gain} & $+32.8$ & $+8.6$ & $+0.79$ \\
 & &  & {\scriptsize$[+19.5, +46.1]$} & {\scriptsize$[+0.05, +22.1]$} & {\scriptsize$[-4.0, +4.8]$} \\\addlinespace[1.5pt]
 & MLP & $+16.5$ & $+36.5$ & $+4.6$ & $+1.0$ \\
 & & {\scriptsize$[+9.5, +24.3]$} & {\scriptsize$[+25.5, +49.0]$} & {\scriptsize$[-0.52, +14.9]$} & {\scriptsize$[-2.9, +6.6]$} \\
\bottomrule
\end{tabular}
\caption{CorrectiveWorst minus CorrectiveMean at the shared start on all 62 correction events ($10^{-3}$ nat), 95\% group-level bootstrap intervals.
Unscaled: native responses at the selected and the last checkpoint; the linear selected values are those of \cref{tab:ultrasound_gain}.
Matched: one source-fitted global response scale per model, both at the source harm $h_{0.5}$ of \cref{eq:budget}, selected checkpoint.}
\label{tab:supp_selection_mlp}
\end{table*}

\paragraph{MLP readout.}
\label{sec:supp_selection_mlp}
Without amplitude matching, CorrectiveWorst has the higher corrected NLL with the MLP readout on both sources (\cref{tab:supp_selection_mlp}); on BUS-BRA its harm is lower by $5.1\times10^{-3}$ nat and its useful gain by $16.8\times10^{-3}$ nat, with lower harm in four of six seeds.
The size depends on the checkpoint: early stopping selected the first validation check (epoch 4) for 44 of 48 BUS-BRA MLP models and the first or second check for 40 of 48 BUSI647 MLP models, and the last checkpoints give larger differences.
The response of an MLP readout is not a sum over concepts, so the matched comparison uses only the global scale, $z=b+\beta\,r(S)$ with $r(S)$ the full response to the correction set $S$ and $\beta\ge0$, fitted with the objective \cref{eq:obj} and the root search of \cref{sec:audit_frontier} on the calibration role.
At $f=0.5$, the linear readout keeps a positive NLL difference, with a lower endpoint near zero on BUSI647, whereas both MLP differences are unresolved.
Source harm is matched, evaluation harm is not forced to match, and its differences include zero.

\subsection{Finite-query selection on Private250}
\label{sec:supp_selection_p}
The Private250 comparison applies the two selection criteria to the same clinical correction task, with four folds and six seeds.
At the shared CorrectiveMean start and shared calibration-fitted query order, the harm-selected model has higher corrected NLL and lower useful gain for both readouts (\cref{tab:selection_p_core}).
The NLL, harm and gain directions agree in all six seeds at $k=2$; the starting-NLL contrast is zero.
Harm selection picks CorrectiveWorst in 22 of 24 linear units and all 24 MLP units, whereas corrected-NLL selection picks it once for each readout; the criteria agree in 3 and 1 units, respectively.
The six seed-level NLL contrasts are $+7.320,+25.381,+12.283,+14.066,+3.633,+21.530$ for the linear readout and $+6.612,+11.790,+12.192,+11.321,+9.371,+9.013$ for the MLP readout ($10^{-3}$ nat).
Among cases wrong before correction, the rescued fraction falls by $17.82$ [$7.49$, $26.39$] pp for the linear readout and $15.35$ [$5.25$, $25.23$] pp for the MLP readout. Intervals in this subsection use 1000 linked original-group bootstrap draws conditional on the trained models.

\begin{table*}[t]
\centering
\footnotesize
\setlength{\tabcolsep}{4pt}
\begin{tabular}{@{}lcccccc@{}}
\toprule
Readout & $\Delta$NLL & $\Delta\Hm$ & $\Delta\Gm$ & $\Delta$Accuracy & $\Delta$Rescued & $\Delta$Broken \\
\midrule
linear & $+14.04$ & $-3.00$ & $-17.04$ & $-1.42$ & $-1.59$ & $-0.17$ \\
 & {\scriptsize$[+8.88, +15.79]$} & {\scriptsize$[-3.49, -1.60]$} & {\scriptsize$[-18.31, -11.17]$} & {\scriptsize$[-2.18, -0.42]$} & {\scriptsize$[-2.35, -0.67]$} & {\scriptsize$[-0.50, +0.00]$} \\\addlinespace[1.5pt]
MLP & $+10.05$ & $-0.97$ & $-11.02$ & $-1.17$ & $-1.17$ & $+0.00$ \\
 & {\scriptsize$[+6.79, +11.25]$} & {\scriptsize$[-1.38, -0.32]$} & {\scriptsize$[-11.87, -7.65]$} & {\scriptsize$[-1.93, -0.42]$} & {\scriptsize$[-1.93, -0.42]$} & {\scriptsize$[-0.17, +0.00]$} \\\addlinespace[1.5pt]
\bottomrule
\end{tabular}
\caption{Private250, harm-selected minus NLL-selected at the selected checkpoint, shared start and query order, $k=2$. NLL, harm and gain are in $10^{-3}$ nat; accuracy, rescued and broken are in percentage points of all evaluation groups. Brackets: conditional 95\% linked-group intervals.}
\label{tab:selection_p_core}
\end{table*}

\begin{table*}[t]
\centering
\footnotesize
\setlength{\tabcolsep}{4pt}
\begin{tabular}{@{}llcccc@{}}
\toprule
Readout & $k$ & $\Delta$NLL & $\Delta\Hm$ & $\Delta\Gm$ & $\Delta$Rescued \\
\midrule
linear & 1 & $+8.24$ & $-1.26$ & $-9.50$ & $-1.01$ \\
 &  & {\scriptsize$[+5.14, +9.11]$} & {\scriptsize$[-1.50, -0.67]$} & {\scriptsize$[-10.22, -6.18]$} & {\scriptsize$[-1.68, -0.34]$} \\\addlinespace[1.5pt]
linear & 2 & $+14.04$ & $-3.00$ & $-17.04$ & $-1.59$ \\
 &  & {\scriptsize$[+8.88, +15.79]$} & {\scriptsize$[-3.49, -1.60]$} & {\scriptsize$[-18.31, -11.17]$} & {\scriptsize$[-2.35, -0.67]$} \\\addlinespace[1.5pt]
linear & 3 & $+17.13$ & $-5.22$ & $-22.36$ & $-1.51$ \\
 &  & {\scriptsize$[+11.21, +19.62]$} & {\scriptsize$[-6.80, -3.52]$} & {\scriptsize$[-24.43, -16.41]$} & {\scriptsize$[-2.26, -0.67]$} \\\addlinespace[1.5pt]
MLP & 1 & $+6.33$ & $-0.26$ & $-6.59$ & $-0.75$ \\
 &  & {\scriptsize$[+3.94, +6.38]$} & {\scriptsize$[-0.61, -0.09]$} & {\scriptsize$[-6.60, -4.26]$} & {\scriptsize$[-1.34, -0.17]$} \\\addlinespace[1.5pt]
MLP & 2 & $+10.05$ & $-0.97$ & $-11.02$ & $-1.17$ \\
 &  & {\scriptsize$[+6.79, +11.25]$} & {\scriptsize$[-1.38, -0.32]$} & {\scriptsize$[-11.87, -7.65]$} & {\scriptsize$[-1.93, -0.42]$} \\\addlinespace[1.5pt]
MLP & 3 & $+12.95$ & $-2.18$ & $-15.13$ & $-1.93$ \\
 &  & {\scriptsize$[+8.97, +13.55]$} & {\scriptsize$[-3.12, -1.59]$} & {\scriptsize$[-15.76, -11.13]$} & {\scriptsize$[-2.60, -0.84]$} \\\addlinespace[1.5pt]
\bottomrule
\end{tabular}
\caption{Private250 budget sensitivity at the shared start and shared query order, selected checkpoints; harm-selected minus NLL-selected ($10^{-3}$ nat; rescued: pp). NLL rises in all six seeds except the linear $k=3$ contrast, which has five positive seeds.}
\label{tab:selection_p_budget}
\end{table*}

\paragraph{Starts, queries and checkpoints.}
With the start shared, the own-uncertainty and random-query comparisons also have positive NLL intervals for both readouts (\cref{tab:selection_p_controls}).
At native starts, the NLL point contrasts are negative and their intervals include zero; their before-correction components are also negative.
At the shared start and order, the last checkpoints retain positive NLL contrasts in all six seeds (\cref{tab:selection_p_last}).
The before-NLL selector ties at a shared start and returns CorrectiveMean.

\begin{table*}[t]
\centering
\footnotesize
\setlength{\tabcolsep}{4pt}
\begin{tabular}{@{}llcccc@{}}
\toprule
Readout & Start and query order & $\Delta$NLL & $\Delta\Hm$ & $\Delta\Gm$ & $\Delta$NLL before \\
\midrule
linear & Shared start, shared order & $+14.04$ & $-3.00$ & $-17.04$ & $+0.00$ \\
 &  & {\scriptsize$[+8.88, +15.79]$} & {\scriptsize$[-3.49, -1.60]$} & {\scriptsize$[-18.31, -11.17]$} & {\scriptsize$[+0.00, +0.00]$} \\\addlinespace[1.5pt]
linear & Shared start, own uncertainty & $+7.03$ & $-3.82$ & $-10.85$ & $+0.00$ \\
 &  & {\scriptsize$[+4.56, +10.22]$} & {\scriptsize$[-4.53, -1.49]$} & {\scriptsize$[-12.70, -7.74]$} & {\scriptsize$[+0.00, +0.00]$} \\\addlinespace[1.5pt]
linear & Shared start, random & $+7.21$ & $-3.42$ & $-10.63$ & $+0.00$ \\
 &  & {\scriptsize$[+4.65, +8.91]$} & {\scriptsize$[-4.24, -2.50]$} & {\scriptsize$[-12.01, -8.07]$} & {\scriptsize$[+0.00, +0.00]$} \\\addlinespace[1.5pt]
linear & Own start, own uncertainty & $-9.29$ & $-1.73$ & $-5.29$ & $-12.86$ \\
 &  & {\scriptsize$[-30.82, +17.41]$} & {\scriptsize$[-3.10, +0.68]$} & {\scriptsize$[-9.24, -0.69]$} & {\scriptsize$[-36.51, +12.99]$} \\\addlinespace[1.5pt]
linear & Own start, random & $-10.02$ & $-1.11$ & $-4.97$ & $-13.87$ \\
 &  & {\scriptsize$[-32.87, +18.89]$} & {\scriptsize$[-2.65, +0.11]$} & {\scriptsize$[-7.88, -0.27]$} & {\scriptsize$[-36.49, +16.02]$} \\\addlinespace[1.5pt]
MLP & Shared start, shared order & $+10.05$ & $-0.97$ & $-11.02$ & $+0.00$ \\
 &  & {\scriptsize$[+6.79, +11.25]$} & {\scriptsize$[-1.38, -0.32]$} & {\scriptsize$[-11.87, -7.65]$} & {\scriptsize$[+0.00, +0.00]$} \\\addlinespace[1.5pt]
MLP & Shared start, own uncertainty & $+4.35$ & $-3.25$ & $-7.60$ & $+0.00$ \\
 &  & {\scriptsize$[+1.93, +5.81]$} & {\scriptsize$[-3.87, -1.79]$} & {\scriptsize$[-8.80, -4.53]$} & {\scriptsize$[+0.00, +0.00]$} \\\addlinespace[1.5pt]
MLP & Shared start, random & $+5.20$ & $-2.84$ & $-8.04$ & $+0.00$ \\
 &  & {\scriptsize$[+3.00, +5.68]$} & {\scriptsize$[-3.02, -1.82]$} & {\scriptsize$[-8.15, -5.23]$} & {\scriptsize$[+0.00, +0.00]$} \\\addlinespace[1.5pt]
MLP & Own start, own uncertainty & $-23.74$ & $-0.70$ & $-2.65$ & $-25.68$ \\
 &  & {\scriptsize$[-35.73, +11.41]$} & {\scriptsize$[-1.83, +1.13]$} & {\scriptsize$[-4.44, +0.88]$} & {\scriptsize$[-38.08, +11.24]$} \\\addlinespace[1.5pt]
MLP & Own start, random & $-16.30$ & $-0.26$ & $-1.71$ & $-17.75$ \\
 &  & {\scriptsize$[-34.96, +12.19]$} & {\scriptsize$[-1.10, +0.93]$} & {\scriptsize$[-3.46, +0.60]$} & {\scriptsize$[-36.76, +11.49]$} \\\addlinespace[1.5pt]
\bottomrule
\end{tabular}
\caption{Private250 selected-checkpoint controls, $k=2$, harm-selected minus NLL-selected ($10^{-3}$ nat). Native-start contrasts include both starting quality and the correction response. Shared-order rows use identical queries; own-uncertainty rows may query different fields.}
\label{tab:selection_p_controls}
\end{table*}

\begin{table*}[t]
\centering
\footnotesize
\setlength{\tabcolsep}{4pt}
\begin{tabular}{@{}llcc@{}}
\toprule
Readout & Start and query order & Harm pick $-$ before-NLL pick & Before-NLL pick $-$ NLL pick \\
\midrule
linear & Shared start, shared order & $+14.21$ & $-0.17$ \\
 &  & {\scriptsize$[+9.50, +16.00]$} & {\scriptsize$[-1.73, +0.18]$} \\\addlinespace[1.5pt]
linear & Shared start, own uncertainty & $+7.70$ & $-0.67$ \\
 &  & {\scriptsize$[+5.41, +9.93]$} & {\scriptsize$[-2.24, +1.39]$} \\\addlinespace[1.5pt]
linear & Shared start, random & $+7.56$ & $-0.35$ \\
 &  & {\scriptsize$[+5.88, +9.18]$} & {\scriptsize$[-2.05, +0.53]$} \\\addlinespace[1.5pt]
linear & Own start, own uncertainty & $-7.31$ & $-1.98$ \\
 &  & {\scriptsize$[-30.59, +18.14]$} & {\scriptsize$[-8.61, +10.30]$} \\\addlinespace[1.5pt]
linear & Own start, random & $-10.02$ & $+0.00$ \\
 &  & {\scriptsize$[-32.05, +18.89]$} & {\scriptsize$[-8.26, +8.37]$} \\\addlinespace[1.5pt]
MLP & Shared start, shared order & $+10.20$ & $-0.15$ \\
 &  & {\scriptsize$[+7.22, +11.34]$} & {\scriptsize$[-0.92, +0.11]$} \\\addlinespace[1.5pt]
MLP & Shared start, own uncertainty & $+5.77$ & $-1.42$ \\
 &  & {\scriptsize$[+4.11, +6.86]$} & {\scriptsize$[-3.12, -0.37]$} \\\addlinespace[1.5pt]
MLP & Shared start, random & $+5.34$ & $-0.14$ \\
 &  & {\scriptsize$[+4.30, +6.02]$} & {\scriptsize$[-1.71, -0.03]$} \\\addlinespace[1.5pt]
MLP & Own start, own uncertainty & $-23.74$ & $+0.00$ \\
 &  & {\scriptsize$[-34.85, +13.16]$} & {\scriptsize$[-13.45, +6.70]$} \\\addlinespace[1.5pt]
MLP & Own start, random & $-16.30$ & $+0.00$ \\
 &  & {\scriptsize$[-33.30, +12.65]$} & {\scriptsize$[-13.65, +5.26]$} \\\addlinespace[1.5pt]
\bottomrule
\end{tabular}
\caption{Private250 selected-checkpoint baseline contrasts in corrected NLL, $k=2$ ($10^{-3}$ nat). The before-NLL pick selects by the uncorrected NLL on calibration data; with the shared start it always chooses CorrectiveMean.}
\label{tab:selection_p_baselines}
\end{table*}

\begin{table*}[t]
\centering
\footnotesize
\setlength{\tabcolsep}{4pt}
\begin{tabular}{@{}llccccc@{}}
\toprule
Readout & Start and query order & $\Delta$NLL & $\Delta\Hm$ & $\Delta\Gm$ & $\Delta$Rescued & Worst picked \\
\midrule
linear & Shared start, shared order & $+19.27$ & $-3.56$ & $-22.83$ & $-2.76$ & 23/0 \\
 &  & {\scriptsize$[+11.87, +20.41]$} & {\scriptsize$[-3.89, -1.85]$} & {\scriptsize$[-23.32, -14.56]$} & {\scriptsize$[-3.27, -0.92]$} &  \\\addlinespace[1.5pt]
linear & Shared start, own uncertainty & $+9.80$ & $-4.60$ & $-14.39$ & $-0.59$ & 24/9 \\
 &  & {\scriptsize$[+4.45, +11.96]$} & {\scriptsize$[-5.75, -2.36]$} & {\scriptsize$[-15.91, -8.11]$} & {\scriptsize$[-1.09, -0.08]$} &  \\\addlinespace[1.5pt]
linear & Shared start, random & $+10.34$ & $-3.79$ & $-14.13$ & $-1.20$ & 24/6 \\
 &  & {\scriptsize$[+6.86, +11.52]$} & {\scriptsize$[-4.66, -2.53]$} & {\scriptsize$[-15.15, -10.21]$} & {\scriptsize$[-1.60, -0.54]$} &  \\\addlinespace[1.5pt]
linear & Own start, own uncertainty & $-1.59$ & $-1.37$ & $-9.50$ & $-1.51$ & 21/10 \\
 &  & {\scriptsize$[-23.91, +24.63]$} & {\scriptsize$[-2.41, +2.36]$} & {\scriptsize$[-10.47, -0.31]$} & {\scriptsize$[-2.35, +0.25]$} &  \\\addlinespace[1.5pt]
linear & Own start, random & $-5.05$ & $-0.65$ & $-7.08$ & $-1.52$ & 19/11 \\
 &  & {\scriptsize$[-26.41, +24.10]$} & {\scriptsize$[-1.98, +1.19]$} & {\scriptsize$[-8.73, -0.89]$} & {\scriptsize$[-2.04, -0.01]$} &  \\\addlinespace[1.5pt]
MLP & Shared start, shared order & $+14.59$ & $-0.62$ & $-15.21$ & $-1.76$ & 24/0 \\
 &  & {\scriptsize$[+10.19, +14.93]$} & {\scriptsize$[-1.53, -0.31]$} & {\scriptsize$[-15.55, -11.21]$} & {\scriptsize$[-2.35, -0.75]$} &  \\\addlinespace[1.5pt]
MLP & Shared start, own uncertainty & $+5.32$ & $-4.09$ & $-9.41$ & $-0.59$ & 24/3 \\
 &  & {\scriptsize$[+2.58, +6.55]$} & {\scriptsize$[-4.34, -2.09]$} & {\scriptsize$[-9.89, -5.55]$} & {\scriptsize$[-1.17, +0.00]$} &  \\\addlinespace[1.5pt]
MLP & Shared start, random & $+6.78$ & $-3.27$ & $-10.06$ & $-1.02$ & 24/0 \\
 &  & {\scriptsize$[+4.59, +7.42]$} & {\scriptsize$[-3.58, -2.40]$} & {\scriptsize$[-10.38, -7.48]$} & {\scriptsize$[-1.31, -0.52]$} &  \\\addlinespace[1.5pt]
MLP & Own start, own uncertainty & $-26.58$ & $+0.43$ & $-1.69$ & $-0.08$ & 16/15 \\
 &  & {\scriptsize$[-46.34, +11.47]$} & {\scriptsize$[-1.19, +1.60]$} & {\scriptsize$[-4.30, +1.97]$} & {\scriptsize$[-0.84, +0.92]$} &  \\\addlinespace[1.5pt]
MLP & Own start, random & $-27.18$ & $+0.05$ & $-1.47$ & $-0.35$ & 16/15 \\
 &  & {\scriptsize$[-50.86, +9.21]$} & {\scriptsize$[-1.44, +0.85]$} & {\scriptsize$[-4.03, +1.35]$} & {\scriptsize$[-0.81, +0.39]$} &  \\\addlinespace[1.5pt]
\bottomrule
\end{tabular}
\caption{Private250 last-checkpoint sensitivity, $k=2$, harm-selected minus NLL-selected ($10^{-3}$ nat; rescued: pp). Worst picked gives harm/NLL criterion counts out of 24 units. The last checkpoint is a sensitivity analysis.}
\label{tab:selection_p_last}
\end{table*}

\begin{table*}[t]
\centering
\footnotesize
\setlength{\tabcolsep}{4pt}
\begin{tabular}{@{}lcccc@{}}
\toprule
Readout & $\Delta$NLL & $\Delta\Hm$ & $\Delta\Gm$ & $\Delta$Rescued \\
\midrule
linear & $+15.93$ & $-3.34$ & $-19.27$ & $-1.84$ \\
 & {\scriptsize$[+12.29, +17.65]$} & {\scriptsize$[-4.35, -2.61]$} & {\scriptsize$[-20.55, -16.22]$} & {\scriptsize$[-2.60, -0.92]$} \\\addlinespace[1.5pt]
MLP & $+10.20$ & $-1.10$ & $-11.29$ & $-1.17$ \\
 & {\scriptsize$[+8.27, +11.83]$} & {\scriptsize$[-1.84, -0.56]$} & {\scriptsize$[-12.47, -9.92]$} & {\scriptsize$[-2.01, -0.50]$} \\
\bottomrule
\end{tabular}
\caption{Private250 direct candidate contrast, CorrectiveWorst minus CorrectiveMean, selected checkpoint at the shared start and query order, $k=2$ ($10^{-3}$ nat; rescued: pp). This contrasts the two candidates directly rather than their criterion-selected outcomes.}
\label{tab:selection_p_candidates}
\end{table*}

\begin{table*}[t]
\centering
\footnotesize
\setlength{\tabcolsep}{4pt}
\begin{tabular}{@{}lcccc@{}}
\toprule
Outcome & Linear: harm pick & Linear: NLL pick & MLP: harm pick & MLP: NLL pick \\
\midrule
NLL before & $+342.48$ & $+342.48$ & $+334.84$ & $+334.84$ \\
 & {\scriptsize$[+316.79, +369.78]$} & {\scriptsize$[+316.79, +369.78]$} & {\scriptsize$[+307.30, +363.30]$} & {\scriptsize$[+307.30, +363.30]$} \\\addlinespace[1.5pt]
NLL after & $+330.42$ & $+316.39$ & $+327.66$ & $+317.61$ \\
 & {\scriptsize$[+304.68, +356.34]$} & {\scriptsize$[+292.94, +344.21]$} & {\scriptsize$[+300.29, +354.60]$} & {\scriptsize$[+292.31, +345.67]$} \\\addlinespace[1.5pt]
$\Hm$ & $+10.86$ & $+13.86$ & $+8.72$ & $+9.69$ \\
 & {\scriptsize$[+9.01, +13.42]$} & {\scriptsize$[+10.86, +16.44]$} & {\scriptsize$[+7.13, +10.31]$} & {\scriptsize$[+7.77, +11.19]$} \\\addlinespace[1.5pt]
$\Gm$ & $+22.92$ & $+39.95$ & $+15.90$ & $+26.91$ \\
 & {\scriptsize$[+20.55, +27.32]$} & {\scriptsize$[+34.90, +42.26]$} & {\scriptsize$[+14.74, +18.85]$} & {\scriptsize$[+23.27, +29.76]$} \\\addlinespace[1.5pt]
Accuracy (\%) & $+92.29$ & $+93.72$ & $+92.38$ & $+93.55$ \\
 & {\scriptsize$[+90.62, +94.06]$} & {\scriptsize$[+92.04, +95.14]$} & {\scriptsize$[+90.37, +94.14]$} & {\scriptsize$[+91.71, +94.97]$} \\\addlinespace[1.5pt]
Rescued (\%) & $+1.59$ & $+3.18$ & $+0.75$ & $+1.93$ \\
 & {\scriptsize$[+1.01, +2.60]$} & {\scriptsize$[+2.09, +4.27]$} & {\scriptsize$[+0.34, +1.51]$} & {\scriptsize$[+1.01, +2.93]$} \\\addlinespace[1.5pt]
Broken (\%) & $+0.50$ & $+0.67$ & $+0.25$ & $+0.25$ \\
 & {\scriptsize$[+0.08, +1.01]$} & {\scriptsize$[+0.25, +1.26]$} & {\scriptsize$[+0.00, +0.67]$} & {\scriptsize$[+0.00, +0.75]$} \\\addlinespace[1.5pt]
\bottomrule
\end{tabular}
\caption{Absolute outcomes of the two Private250 selection criteria, selected checkpoint, shared start and order, $k=2$. NLL, harm and gain: $10^{-3}$ nat; accuracy, rescued and broken: percentages of all evaluation groups.}
\label{tab:selection_p_levels}
\end{table*}

\section{Additional results}
\label{sec:supp_results}

\begin{table*}[t]
\centering
\footnotesize
\setlength{\tabcolsep}{6pt}
\begin{tabular}{@{}lcccc@{}}
\toprule
 & Raw & \multicolumn{3}{c}{Temperature-controlled} \\
\cmidrule(lr){2-2}\cmidrule(l){3-5}
 & M $-$ E & M $-$ E & E$_\beta$ $-$ E & M $-$ E$_\beta$ \\
\midrule
Max.\ sampled partial harm & $-4.17$ {\scriptsize[$-6.91$, $-4.83$]} & $-3.92$ {\scriptsize[$-6.38$, $-4.50$]} & $-3.93$ {\scriptsize[$-6.36$, $-4.55$]} & $+0.01$ {\scriptsize[$-0.06$, $+0.10$]} \\
Harmful-path prevalence (\%) & $-37.7$ {\scriptsize[$-36.3$, $-26.0$]} & $-39.5$ {\scriptsize[$-38.2$, $-28.0$]} & $-24.5$ {\scriptsize[$-21.6$, $-17.4$]} & $-15.0$ {\scriptsize[$-18.9$, $-7.7$]} \\
Per-event partial harm & -- & $-16.5$ {\scriptsize[$-17.4$, $-15.7$]} & $-17.4$ {\scriptsize[$-18.3$, $-16.6$]} & $+0.93$ {\scriptsize[$+0.81$, $+1.05$]} \\
Partial harm $>0.05$ (\%) & -- & $-10.6$ {\scriptsize[$-11.1$, $-10.1$]} & $-11.4$ {\scriptsize[$-12.0$, $-10.9$]} & $+0.82$ {\scriptsize[$+0.72$, $+0.91$]} \\
\midrule
Pre-intervention NLL & $+11.1$ {\scriptsize[$+8.2$, $+14.2$]} & $+4.9$ {\scriptsize[$+3.6$, $+6.2$]} & $0$ & $+4.9$ {\scriptsize[$+3.6$, $+6.2$]} \\
Partially corrected NLL & $+10.9$ {\scriptsize[$+8.2$, $+13.9$]} & $+4.7$ {\scriptsize[$+3.6$, $+5.9$]} & $+0.6$ {\scriptsize[$-0.4$, $+1.7$]} & $+4.1$ {\scriptsize[$+2.8$, $+5.4$]} \\
Fully corrected NLL & $+11.7$ {\scriptsize[$+8.6$, $+15.0$]} & $+5.4$ {\scriptsize[$+3.6$, $+7.1$]} & $+2.0$ {\scriptsize[$+0.1$, $+4.1$]} & $+3.3$ {\scriptsize[$+2.1$, $+4.6$]} \\
\midrule
Pre-intervention acc.\ (pp) & $-0.06$ & $-0.06$ & -- & -- \\
Fully corrected acc.\ (pp) & $-0.17$ & $-0.17$ & -- & -- \\
\bottomrule
\end{tabular}
\caption{Jointly trained CUB hybrids, mean paired differences over 80 concept sets ($10^{-3}$ nat unless stated).
M: endpoint$+$monotonic; E: endpoint; E$_\beta$: E with one source-fitted scalar matching M's native first-step source harm.
Brackets: 95\% percentile intervals of image bootstraps conditional on the trained models (M $-$ E: 1000 draws; E$_\beta$: scalar fixed at its source fit).
For the maximum-type rows, resampling noise raises a maximum over many prefixes, by different amounts for the two models, so the interval of their difference can exclude the point estimate; these intervals are descriptive, and inference on the path metrics rests on the per-set signs (80 of 80 sets) and the basic intervals given in the text.
Accuracy rows are point estimates.}
\label{tab:native}
\end{table*}

\paragraph{Jointly trained CUB models: reported and reproduced metrics.}
\Cref{tab:supp_v_compare} sets the CUB metrics reported in~\cite{verghese2026harmful} beside our reproduction of the same comparison, M minus E.
On the reported metrics the two agree in sign and are of similar size: fewer harmful paths, a lower maximum partial harm, an accuracy change of at most 0.1 pp and slightly better concept predictions.
The lower rows add what~\cite{verghese2026harmful} does not report: the NLL before, during and after correction is higher, most of the lower partial harm is the worse start, and one source-fitted scale on E reaches the maximum sampled partial harm of M at a lower corrected NLL.

\begin{table*}[t]
\centering
\footnotesize
\setlength{\tabcolsep}{6pt}
\begin{tabular}{@{}lccc@{}}
\toprule
M $-$ E & Reported in~\cite{verghese2026harmful} & Ours, raw & Ours, temperature-controlled \\
\midrule
Harmful-path prevalence (pp) & $-39.9$ {\scriptsize[$-42.2$, $-37.5$]} & $-37.7$ {\scriptsize[$-36.3$, $-26.0$]} & $-39.5$ {\scriptsize[$-38.2$, $-28.0$]} \\
Max.\ sampled partial harm ($10^{-3}$ nat) & $-5.0$ {\scriptsize[$-5.3$, $-4.7$]} & $-4.17$ {\scriptsize[$-6.91$, $-4.83$]} & $-3.92$ {\scriptsize[$-6.38$, $-4.50$]} \\
Task accuracy before correction (pp) & $-0.10$ {\scriptsize[$-0.16$, $-0.04$]} & $-0.06$ & $-0.06$ \\
Concept mean squared error & $-0.0097$ {\scriptsize[$-0.0107$, $-0.0087$]} & $-0.009$ & $-0.009$ \\
\midrule
Pre-intervention NLL ($10^{-3}$ nat) & n.r. & $+11.1$ {\scriptsize[$+8.2$, $+14.2$]} & $+4.9$ {\scriptsize[$+3.6$, $+6.2$]} \\
Partially corrected NLL ($10^{-3}$ nat) & n.r. & $+10.9$ {\scriptsize[$+8.2$, $+13.9$]} & $+4.7$ {\scriptsize[$+3.6$, $+5.9$]} \\
Fully corrected NLL ($10^{-3}$ nat) & n.r. & $+11.7$ {\scriptsize[$+8.6$, $+15.0$]} & $+5.4$ {\scriptsize[$+3.6$, $+7.1$]} \\
Max.\ partial harm, before-intervention part ($10^{-3}$ nat) & n.r. & $-11.1$ & $-4.9$ \\
Max.\ partial harm, after-intervention part ($10^{-3}$ nat) & n.r. & $+6.9$ {\scriptsize[$+2.5$, $+8.0$]} & indistinguishable from 0 \\
\midrule
M $-$ E$_\beta$, max.\ sampled partial harm ($10^{-3}$ nat) & n.r. & -- & $+0.01$ {\scriptsize[$-0.06$, $+0.10$]} \\
M $-$ E$_\beta$, partially corrected NLL ($10^{-3}$ nat) & n.r. & -- & $+4.1$ {\scriptsize[$+2.8$, $+5.4$]} \\
\bottomrule
\end{tabular}
\caption{Jointly trained CUB hybrids, endpoint$+$monotonic (M) minus endpoint (E).
First column: values reported in~\cite{verghese2026harmful} for its 80 concept sets, with its 95\% intervals over concept sets.
Other columns: our reproduction on 80 regenerated sets with the same image features, residual, $\lambda$ values, stop-gradient, 512 orders and harm threshold, from \cref{tab:native} and \cref{sec:res_native}; brackets are image-bootstrap intervals conditional on the trained models, and maximum-type points can lie outside them.
E$_\beta$: E with one source-fitted scalar matching the native first-step source harm of M.
n.r.: not reported in~\cite{verghese2026harmful}.
Concept predictions do not depend on the task temperature.}
\label{tab:supp_v_compare}
\end{table*}

\paragraph{Maximum-type path metrics.}
The percentile intervals of \cref{tab:native} for the maximum sampled partial harm and the harmful-path prevalence exclude their point estimates because resampling inflates a maximum over many prefixes, and by different amounts for the two models: the bootstrap raises the mean maximum sampled partial harm of the temperature-controlled endpoint models from $4.22$ to $5.89$ and that of the penalized models only from $0.30$ to $0.49$ ($10^{-3}$ nat).
Basic bootstrap intervals, which reflect the percentile interval about the point estimate, are [$-3.33$, $-1.45$] (temperature-controlled) and [$-3.52$, $-1.44$] (raw)$\times10^{-3}$ nat for the harm difference and [$-51.1$, $-40.9$] and [$-49.4$, $-39.1$] pp for the prevalence difference; like the per-set signs (lower in 80 of 80 sets), they exclude zero.

\begin{figure*}[t]
\centering
\includegraphics[width=\textwidth]{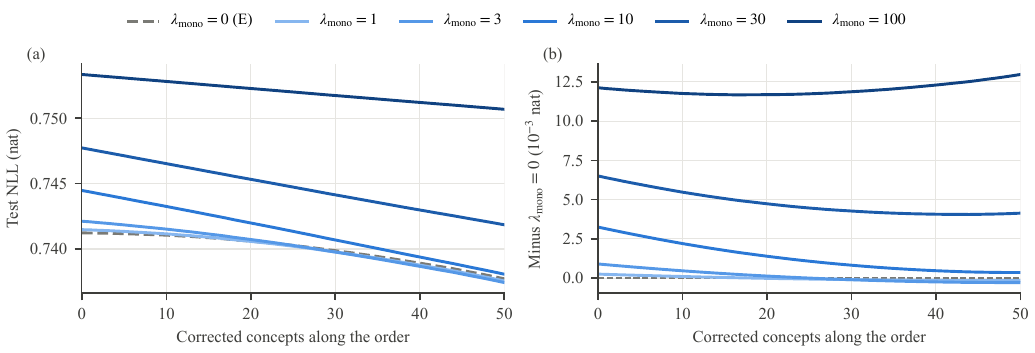}
\caption{Absolute test NLL along the 512 fixed correction orders of the penalty-strength sweep (\cref{tab:supp_strength}), mean over 10 concept sets $\times$ three seeds.
(a) NLL after each number of corrected concepts; (b) difference to the endpoint models ($\lambda_{\mathrm{mono}}=0$).
The weak penalties stay within $10^{-3}$ nat of the endpoint models along the whole path; from $\lambda_{\mathrm{mono}}=10$ the NLL is higher before correction and stays higher as concepts are corrected.}
\label{fig:supp_strength_depth}
\end{figure*}

\paragraph{Penalty strength and two training controls.}
We retrain the jointly trained hybrids of \cref{sec:res_native} on 10 concept sets (the lowest-numbered set of each stratum) with three seeds, at $\lambda_{\mathrm{mono}}\in\{0,1,3,10,30,100\}$ and with the recipe of the main study.
Two controls replace the hinge in the same objective: an L2 penalty on the concept-to-task weights (nine strengths), and plain task cross-entropy at the same sampled partial-correction states with the same stop-gradient (five strengths).
Each control is matched per set and seed to the validation maximum sampled partial harm of each $\lambda_{\mathrm{mono}}$; test data enter only the contrasts (\cref{tab:supp_strength}).
Every $\lambda_{\mathrm{mono}}\ge1$ lowers the maximum sampled partial harm in all 30 pairs, and the NLL cost grows with the weight.
At $\lambda_{\mathrm{mono}}\in\{1,3\}$ the harm reduction is 10\% and 28\% of that at $\lambda_{\mathrm{mono}}=100$, the partially corrected NLL is unchanged within its interval, the pre-intervention NLL is slightly higher and the fully corrected NLL slightly lower; from $\lambda_{\mathrm{mono}}=10$ on, the partially corrected NLL rises.
\Cref{fig:supp_strength_depth} shows the absolute NLL at every path prefix.
The NLL cost of \cref{sec:res_native} therefore comes with the strong penalty.
At matched validation harm the monotonic penalty has the lower partially corrected NLL against L2 shrinkage for $\lambda_{\mathrm{mono}}\le30$ and the higher one at $\lambda_{\mathrm{mono}}=100$; on test data the L2 controls have slightly lower maximum harm at $\lambda_{\mathrm{mono}}\in\{3,10,30\}$, so these comparisons are not exactly harm-neutral.
Partial-state cross-entropy does not lower the validation harm at any strength ($6.6$--$6.7$ against $6.5\times10^{-3}$ nat for E), so it cannot be matched; it lowers the partially corrected NLL instead, by $4.6$--$7.2\times10^{-3}$ nat relative to E.

\begin{table*}[t]
\centering
\footnotesize
\setlength{\tabcolsep}{5pt}
\begin{tabular}{@{}lcccccc@{}}
\toprule
 & \multicolumn{5}{c}{Monotonic penalty minus endpoint ($\lambda_{\mathrm{mono}}=0$)} & Minus L2 control \\
\cmidrule(lr){2-6}\cmidrule(l){7-7}
 & & & \multicolumn{3}{c}{NLL} & NLL \\
\cmidrule(lr){4-6}\cmidrule(l){7-7}
$\lambda_{\mathrm{mono}}$ & Max.\ partial harm & Prevalence (pp) & Pre-intervention & Partially corrected & Fully corrected & Partially corrected \\
\midrule
1 & $-0.42$ & $-2.2$ & $+0.25$ & $-0.03$ & $-0.16$ & $-0.17$ \\
 &  &  & {\scriptsize$[+0.10, +0.40]$} & {\scriptsize$[-0.17, +0.12]$} & {\scriptsize$[-0.30, -0.02]$} & {\scriptsize$[-0.22, -0.10]$} \\\addlinespace[1.5pt]
3 & $-1.19$ & $-7.2$ & $+0.90$ & $+0.10$ & $-0.29$ & $-0.55$ \\
 &  &  & {\scriptsize$[+0.65, +1.15]$} & {\scriptsize$[-0.13, +0.32]$} & {\scriptsize$[-0.52, -0.04]$} & {\scriptsize$[-0.70, -0.39]$} \\\addlinespace[1.5pt]
10 & $-2.92$ & $-26.7$ & $+3.25$ & $+1.31$ & $+0.35$ & $-3.00$ \\
 &  &  & {\scriptsize$[+2.62, +3.93]$} & {\scriptsize$[+0.76, +1.92]$} & {\scriptsize$[-0.36, +1.03]$} & {\scriptsize$[-3.55, -2.47]$} \\\addlinespace[1.5pt]
30 & $-3.97$ & $-43.0$ & $+6.51$ & $+4.74$ & $+4.14$ & $-1.00$ \\
 &  &  & {\scriptsize$[+5.07, +8.03]$} & {\scriptsize$[+3.45, +6.17]$} & {\scriptsize$[+2.57, +5.76]$} & {\scriptsize$[-1.74, -0.24]$} \\\addlinespace[1.5pt]
100 & $-4.22$ & $-38.7$ & $+12.13$ & $+12.00$ & $+12.97$ & $+5.85$ \\
 &  &  & {\scriptsize$[+8.91, +15.52]$} & {\scriptsize$[+8.91, +15.26]$} & {\scriptsize$[+9.54, +16.47]$} & {\scriptsize$[+3.26, +8.48]$} \\\addlinespace[1.5pt]
\bottomrule
\end{tabular}
\caption{Penalty strength on jointly trained CUB hybrids, 10 concept sets (one per stratum) $\times$ three seeds; mean paired differences ($10^{-3}$ nat unless stated) with 95\% image-bootstrap intervals conditional on the trained models. Last column: monotonic penalty minus an L2 penalty on the concept-to-task weights whose strength is chosen per set and seed to match the validation maximum sampled partial harm. Harm and prevalence intervals are omitted because maximum-type points can lie outside their percentile intervals; no unit has a positive harm or prevalence difference.}
\label{tab:supp_strength}
\end{table*}

\paragraph{Where the per-example penalty falls.}
The monotonic objective of~\cite{verghese2026harmful} is motivated by population risk along correction paths, but the trained penalty is a per-example hinge on each correction step.
On the 240 jointly trained CUB models we measure, over every step of the 512 orders, the share $\pi$ of the per-example hinge mass $\mathbb E[(\Delta\ell)_+]$ that lies on steps whose population risk $\mathbb E[\Delta\ell]$ decreases (\cref{tab:supp_mismatch}).
We call the share large if $\pi\ge0.5$ for M on the fit split, on average and in at least 7 of the 10 strata.
It is large: $\pi=0.83$ for M, between $0.71$ and $0.91$ in every stratum, against $0.40$ for E and $0.002$ for observational models, and positive population changes account for only 0.5\% of the hinge mass of M (E: 5.6\%).
On the test split, $\pi$ is $0.70$ for M and $0.51$ for E.
What the penalty still charges in M is therefore mostly individual losses inside steps that lower the population risk; as a share after training, this may partly reflect the removal of population-harmful steps.
The endpoint share, for a model the penalty never shaped, and the synthetic counterexample below show that the mismatch arises without this selection.

\begin{table}[t]
\centering
\footnotesize
\setlength{\tabcolsep}{3pt}
\begin{tabular}{@{}lcccc@{}}
\toprule
 & \multicolumn{2}{c}{Fit split} & \multicolumn{2}{c}{Test split} \\
\cmidrule(lr){2-3}\cmidrule(l){4-5}
 & $\pi$ & Range & $\pi$ & Range \\
\midrule
Observational & 0.00 & 0.00--0.02 & 0.17 & 0.06--0.32 \\
Endpoint (E) & 0.40 & 0.08--0.85 & 0.51 & 0.37--0.72 \\
Endpoint$+$mono.\ (M) & 0.83 & 0.65--0.98 & 0.70 & 0.49--0.89 \\
\midrule
M $-$ E & $+0.43$ & 80/80 $>0$ & $+0.18$ & 79/80 $>0$ \\
\bottomrule
\end{tabular}
\caption{Share $\pi$ of the per-example hinge mass $\mathbb E[(\Delta\ell)_+]$ that lies on correction steps whose population risk decreases, averaged over the 80 concept sets and 512 orders of the jointly trained CUB models; range over sets; last row: number of sets with M $-$ E $>0$.}
\label{tab:supp_mismatch}
\end{table}

\paragraph{Two synthetic counterexamples.}
We isolate the two mechanisms in two small synthetic examples, separate from the empirical studies: three independent binary concepts, a logistic label, noisy concept evidence with Bayes concept forecasts and a linear concept head, on 200,000 simulated cases.
(a) The NLL-optimal head lowers the population risk along every correction step (largest change $-0.011$ nat), yet individual losses rise inside these steps (per-example hinge mass $0.064$ nat).
Adding the per-example hinge penalty of the monotonic objective raises the NLL before, during and after correction ($\lambda=1$: $+0.005$, $+0.011$ and $+0.019$ nat), and from $\lambda=10$ the fitted head ignores the concepts altogether (all weights zero, NLL $0.693$ in every state), whereas a population-level penalty on $(\mathbb E[\Delta\ell])_+$ leaves the NLL-optimal head unchanged.
The per-example hinge thus costs likelihood even where there is no population harm to remove.
(b) For the NLL-optimal head, one global amplitude reaches a given correction harm only at a higher NLL than one amplitude per concept, by $0.07$, $0.29$ and $0.55\times10^{-3}$ nat at 75, 50 and 25\% of the native harm, because the per-concept family shrinks the weaker concepts more (0.60, 0.52 and 0.43 at 50\%).
A global-scale baseline can therefore be beaten by changing the direction of the response, which is why we also report per-concept and per-field families where the readout allows.

\paragraph{Concept bottleneck models without a residual.}
The native correction responses of our models are too strong for the source likelihood: the source-optimal global scale $\beta^{\mathrm{NLL}}$ has median $0.27$ on BUS-BRA (\cref{sec:res_bus}).
A residual that carries concept information is not needed for this.
On the CUB setup of \cref{sec:res_native}, with the 10 concept sets and three seeds of \cref{tab:supp_strength} and the training recipe of the jointly trained hybrids, we train concept bottleneck models without a residual: independently (the head on the annotated concepts), sequentially (the head on the predicted concept probabilities) and jointly.
On the validation split, the scale that minimizes the NLL of the 50 single-concept corrections has median $0.41$ (5--95\%: $0.38$--$0.45$) for the independent models, $0.16$ ($0.08$--$0.21$) for the sequential models and $0$ for all joint models, whose corrections raise the source NLL at every positive scale; the endpoint hybrids, with their residual, give $0.30$ ($0.004$--$0.72$).
All 30 independent models have $\beta^{\mathrm{NLL}}<1$, so a bottleneck without a residual does not imply $\beta^{\mathrm{NLL}}\approx1$; the prefixes of 16 fixed orders give the same picture.
The concept layers and the sequential and joint models were still improving at the 300-epoch limit of the recipe.

\paragraph{Frozen CUB, secondary combinations.}
\begin{table}[t]
\centering
\footnotesize
\setlength{\tabcolsep}{4pt}
\begin{tabular}{@{}lccc@{}}
\toprule
 & \multicolumn{3}{c}{Penalized $-$ unpenalized ($10^{-4}$ nat)} \\
\cmidrule(l){2-4}
Operating point & $\Delta\Hm$ & $\Delta\Gm$ & $\Delta$NLL \\
\midrule
$\lambda=0$ (unmatched) & $-14.8$ & $-15.6$ & $+0.81$ \\
 & {\scriptsize$[-22.9, -12.5]$} & {\scriptsize$[-23.1, -13.2]$} & {\scriptsize$[-0.63, +1.63]$} \\\addlinespace[1.5pt]
$f=1$ & $+0.34$ & $-0.46$ & $+0.80$ \\
 & {\scriptsize$[-1.53, +2.10]$} & {\scriptsize$[-2.39, +1.04]$} & {\scriptsize$[+0.14, +1.96]$} \\\addlinespace[1.5pt]
$f=0.75$ & $+0.35$ & $-0.11$ & $+0.46$ \\
 & {\scriptsize$[-1.08, +1.73]$} & {\scriptsize$[-1.56, +1.12]$} & {\scriptsize$[-0.16, +1.33]$} \\\addlinespace[1.5pt]
$f=0.5$ (primary) & $+0.33$ & $+0.18$ & $+0.15$ \\
 & {\scriptsize$[-0.65, +1.28]$} & {\scriptsize$[-0.80, +1.07]$} & {\scriptsize$[-0.38, +0.77]$} \\\addlinespace[1.5pt]
$f=0.25$ & $+0.22$ & $+0.35$ & $-0.13$ \\
 & {\scriptsize$[-0.29, +0.76]$} & {\scriptsize$[-0.19, +0.87]$} & {\scriptsize$[-0.46, +0.28]$} \\\addlinespace[1.5pt]
$f=0.1$ & $+0.10$ & $+0.30$ & $-0.20$ \\
 & {\scriptsize$[-0.10, +0.39]$} & {\scriptsize$[+0.05, +0.57]$} & {\scriptsize$[-0.38, +0.05]$} \\
\bottomrule
\end{tabular}
\caption{Frozen CUB response maps, penalized minus unpenalized, endpoint background, $F_{50}$; 30 paired maps, $10^{-4}$ nat.
Brackets: 95\% intervals of the joint source/evaluation bootstrap (1000 draws, all post-hoc parameters refitted).
The margin is $\delta=5\times10^{-4}$ nat; in native units $\Delta\Hm=-49.4\times10^{-4}$.}
\label{tab:frozen}
\end{table}

\Cref{tab:supp_frozen_sec} completes \cref{tab:frozen} with the other three combinations of background and family.
All 19 complete points are conservatively equivalent; the incomplete point is discussed in \cref{sec:supp_plans}.
Neither arm has a Pareto advantage at any point.
\Cref{tab:supp_replica} gives the replica-level estimates at $f=0.5$.
With $F_1$ in the endpoint background, all three replicas have $\Delta\Hm\le0$ and $\Delta\Gm>0$ at $f\le0.5$, and two of three in the partial-state background, but the pooled intervals give no Pareto advantage there either.

\begin{table}[t]
\centering
\small
\setlength{\tabcolsep}{2.5pt}
\resizebox{\columnwidth}{!}{%
\begin{tabular}{@{}lcccl@{}}
\toprule
Point & $\Delta\Hm$ & $\Delta\Gm$ & $\Delta$NLL & Label \\
\midrule
\multicolumn{5}{@{}l}{\emph{Endpoint background, $F_1$}} \\
$\lambda=0$ & $-13.8$ [$-18.4$, $-9.1$] & $-14.3$ [$-18.7$, $-9.7$] & $+0.50$ [$+0.05$, $+1.02$] & -- \\
$f=1$ & $-0.09$ [$-1.56$, $+1.48$] & $-0.25$ [$-1.76$, $+1.27$] & $+0.17$ [$-0.24$, $+0.69$] & E \\
$f=0.75$ & $-0.07$ [$-1.18$, $+1.11$] & $+0.05$ [$-1.10$, $+1.23$] & $-0.12$ [$-0.44$, $+0.25$] & E \\
$f=0.5$ & $-0.05$ [$-0.79$, $+0.74$] & $+0.21$ [$-0.58$, $+1.02$] & $-0.26$ [$-0.49$, $-0.03$] & E, W$_\mathrm{P}$ \\
$f=0.25$ & $-0.03$ [$-0.40$, $+0.37$] & $+0.20$ [$-0.21$, $+0.62$] & $-0.22$ [$-0.35$, $-0.11$] & E, W$_\mathrm{P}$ \\
$f=0.1$ & $-0.01$ [$-0.16$, $+0.15$] & $+0.10$ [$-0.06$, $+0.27$] & $-0.11$ [$-0.17$, $-0.07$] & E, W$_\mathrm{P}$ \\
\midrule
\multicolumn{5}{@{}l}{\emph{Partial-state background, $F_{50}$}} \\
$\lambda=0$ & $-18.2$ [$-26.3$, $-16.0$] & $-19.0$ [$-26.5$, $-16.7$] & $+0.87$ [$-0.61$, $+1.69$] & -- \\
$f=1$ & $+0.57$ [$-0.46$, $+2.28$] & $-0.14$ [$-1.34$, $+1.12$] & $+0.71$ [$+0.11$, $+2.07$] & E, W$_\mathrm{U}$ \\
$f=0.75$ & $+0.57$ [$-0.25$, $+2.02$] & $+0.15$ [$-0.77$, $+1.22$] & $+0.42$ [$-0.16$, $+1.50$] & E \\
$f=0.5$ & $+0.50$ [$-0.09$, $+1.63$] & $+0.39$ [$-0.28$, $+1.26$] & $+0.11$ [$-0.36$, $+1.00$] & E \\
$f=0.25$ & $+0.31$ [$+0.005$, $+1.06$] & $+0.52$ [$+0.11$, $+1.09$] & $-0.21$ [$-0.52$, $+0.45$] & E \\
$f=0.1$ & $+0.16$ [$+0.02$, $+0.61$] & $+0.41$ [$+0.22$, $+0.77$] & $-0.25$ [$-0.47$, $+0.13$] & incomplete (990/1000) \\
\midrule
\multicolumn{5}{@{}l}{\emph{Partial-state background, $F_1$}} \\
$\lambda=0$ & $-25.5$ [$-30.7$, $-20.6$] & $-26.4$ [$-31.5$, $-21.2$] & $+0.92$ [$+0.37$, $+1.52$] & -- \\
$f=1$ & $+0.06$ [$-0.88$, $+1.19$] & $+0.26$ [$-0.67$, $+1.29$] & $-0.20$ [$-0.70$, $+0.49$] & E \\
$f=0.75$ & $+0.03$ [$-0.68$, $+0.87$] & $+0.57$ [$-0.15$, $+1.39$] & $-0.54$ [$-0.91$, $-0.09$] & E, W$_\mathrm{P}$ \\
$f=0.5$ & $+0.003$ [$-0.46$, $+0.55$] & $+0.65$ [$+0.15$, $+1.25$] & $-0.65$ [$-0.91$, $-0.36$] & E, W$_\mathrm{P}$ \\
$f=0.25$ & $-0.006$ [$-0.23$, $+0.26$] & $+0.48$ [$+0.21$, $+0.81$] & $-0.48$ [$-0.65$, $-0.34$] & E, W$_\mathrm{P}$ \\
$f=0.1$ & $-0.005$ [$-0.09$, $+0.10$] & $+0.23$ [$+0.12$, $+0.37$] & $-0.24$ [$-0.31$, $-0.17$] & E, W$_\mathrm{P}$ \\
\bottomrule
\end{tabular}}
\caption{Frozen CUB response maps, secondary combinations, penalized minus unpenalized, 30 paired maps, $10^{-4}$ nat, 95\% joint source/evaluation intervals (1000 draws).
E: conservative equivalence ($\Delta\Hm$, $\Delta\Gm$ and $\Delta$NLL intervals inside $\pm\delta$, $\delta=5\times10^{-4}$).
W: the $\Delta$NLL interval excludes zero while $\Delta\Hm$ stays inside $\pm\delta$, favoring the penalized (P) or the unpenalized (U) arm.
The last point of the partial-state $F_{50}$ block has 990 complete draws.}
\label{tab:supp_frozen_sec}
\end{table}

\begin{table}[t]
\centering
\small
\setlength{\tabcolsep}{3pt}
\resizebox{\columnwidth}{!}{%
\begin{tabular}{@{}lccc@{}}
\toprule
Combination & $\Delta\Hm$ & $\Delta\Gm$ & $\Delta$NLL \\
\midrule
Endpoint, $F_{50}$ (primary) & $+3.6$, $+3.3$, $+3.0$ & $+1.4$, $+4.1$, $-0.02$ & $+2.2$, $-0.8$, $+3.0$ \\
Endpoint, $F_1$ & $-0.3$, $-0.6$, $-0.7$ & $+2.9$, $+2.0$, $+1.3$ & $-3.1$, $-2.6$, $-2.0$ \\
Partial-state, $F_{50}$ & $+5.0$, $+5.4$, $+4.5$ & $+3.7$, $+6.1$, $+1.9$ & $+1.4$, $-0.7$, $+2.6$ \\
Partial-state, $F_1$ & $+0.6$, $-0.4$, $-0.1$ & $+7.8$, $+5.4$, $+6.3$ & $-7.2$, $-5.9$, $-6.3$ \\
\bottomrule
\end{tabular}}
\caption{Frozen CUB, replica-level point estimates at $f=0.5$ (replicas 1--3), penalized minus unpenalized, $10^{-5}$ nat.
In the primary combination every replica interval crosses zero and lies within $\pm1.4\times10^{-4}$.}
\label{tab:supp_replica}
\end{table}

\paragraph{Absolute harm and gain at matched budgets.}
Matching the source harm does not drive both arms to a zero response.
\Cref{tab:supp_abs} lists the absolute evaluation means of each arm in the primary combination.
At the primary point the unpenalized and the penalized arm keep a useful gain of $1.90$ and $1.92\times10^{-3}$ nat against a harm of $1.70$ and $1.73\times10^{-3}$ nat; their gains differ by 1.0\%, and $\delta$ is 26\% of the unpenalized gain.
Both arms start from the same uncorrected NLL, so in every row the NLL equals the uncorrected NLL plus $\Hm$ minus $\Gm$.

\begin{table}[t]
\centering
\small
\setlength{\tabcolsep}{4pt}
\begin{tabular}{@{}llcccc@{}}
\toprule
Point & Arm & $\Hm$ & $\Gm$ & $\Gm-\Hm$ & NLL \\
\midrule
$f=1$ & Unpenalized & $3.368$ & $3.634$ & $0.266$ & $0.73922$ \\
 & Penalized & $3.402$ & $3.588$ & $0.186$ & $0.73930$ \\
\midrule
$f=0.5$ & Unpenalized & $1.696$ & $1.904$ & $0.208$ & $0.73928$ \\
 & Penalized & $1.729$ & $1.922$ & $0.194$ & $0.73929$ \\
\midrule
$f=0.1$ & Unpenalized & $0.349$ & $0.418$ & $0.070$ & $0.73942$ \\
 & Penalized & $0.359$ & $0.449$ & $0.090$ & $0.73940$ \\
\bottomrule
\end{tabular}
\caption{Frozen CUB response maps, endpoint background, $F_{50}$: absolute evaluation means of each arm at matched budgets, point estimates over 30 maps; $\Hm$, $\Gm$ and $\Gm-\Hm$ in $10^{-3}$ nat, NLL in nat.
The uncorrected NLL is $0.73949$ nat for both arms.}
\label{tab:supp_abs}
\end{table}

\paragraph{Cross-family tail ordering.}
At $f=0.5$ the point estimates order the four combinations in the same way in both backgrounds, from unpenalized$+F_1$ through penalized$+F_1$ and penalized$+F_{50}$ to unpenalized$+F_{50}$ ($1.9$, $3.2$, $6.5$ and $7.2\times10^{-3}$ in the endpoint background).
The tail contrast changes direction with the amplitude family.
The families have different $\lambda=0$ source harms, so the same $f$ does not impose an identical absolute budget across $F_1$ and $F_{50}$; this cross-family ordering is descriptive.

\paragraph{Margin sensitivity (post hoc).}
For each prespecified point, \cref{tab:supp_margin} gives the smallest symmetric margin that contains all three existing 95\% intervals of $\Delta\Hm$, $\Delta\Gm$ and $\Delta$NLL.
Over the 19 completely resampled points the largest value is $2.39\times10^{-4}$ nat, so a margin of $\pm2.5\times10^{-4}$, half of $\delta$, would also contain all of them; at the primary point the value is $1.28\times10^{-4}$.
The three intervals are not independent, since $\Delta\mathrm{NLL}=\Delta\Hm-\Delta\Gm$ on fixed-start events.

\begin{table}[t]
\centering
\small
\setlength{\tabcolsep}{4pt}
\begin{tabular}{@{}lccccc@{}}
\toprule
Combination & $f=1$ & $0.75$ & $0.5$ & $0.25$ & $0.1$ \\
\midrule
Endpoint, $F_{50}$ & $2.39$ & $1.73$ & $1.28$ & $0.87$ & $0.57$ \\
Endpoint, $F_1$ & $1.76$ & $1.23$ & $1.02$ & $0.62$ & $0.27$ \\
Partial-state, $F_{50}$ & $2.28$ & $2.02$ & $1.63$ & $1.09$ & $0.77^{*}$ \\
Partial-state, $F_1$ & $1.29$ & $1.39$ & $1.25$ & $0.81$ & $0.37$ \\
\bottomrule
\end{tabular}
\caption{Frozen CUB maps: smallest symmetric margin ($10^{-4}$ nat) that contains the 95\% intervals of $\Delta\Hm$, $\Delta\Gm$ and $\Delta$NLL, a post-hoc description.
$^{*}$Incomplete point, 990 of 1000 draws.}
\label{tab:supp_margin}
\end{table}

\paragraph{Tail harm depends on the amplitude family.}
\label{sec:res_tail}
The fraction of events with harm above $0.05$ nat, a secondary rate without a decision rule, reverses with the family at matched source harm: against $F_{50}$ the penalized arm has 9--19\% fewer large-harm events in both CUB backgrounds (intervals exclude zero for $f\ge0.25$), against $F_1$ it has more, and jointly trained models show the same reversal ($-0.24$ and $+0.89\times10^{-3}$ at $f=0.5$, both excluding zero; \suppref{tab:supp_tail}).

\paragraph{Large-harm rates by family.}
\Cref{tab:supp_tail} lists the difference in the fraction of events with harm above $0.05$ nat for every setting and budget.
At $f=0.5$, the arm-level rates order the four frozen combinations in the same way in both backgrounds: unpenalized$+F_1$ $<$ penalized$+F_1$ $<$ penalized$+F_{50}$ $<$ unpenalized$+F_{50}$, with $1.92$, $3.19$, $6.49$ and $7.18\times10^{-3}$ in the endpoint background and $0.71$, $2.42$, $5.33$ and $6.61\times10^{-3}$ in the partial-state background.
Budgets are set by each family's own $\lambda=0$ harm, so the evaluation mean harm of $F_1$ is 7--37\% lower than that of $F_{50}$ at the same $f$.
This does not explain the ordering: at $f=0.75$ the unpenalized arm with $F_1$ has higher mean harm ($2.36\times10^{-3}$) than the penalized arm with $F_{50}$ at $f=0.5$ ($1.73\times10^{-3}$), yet a lower large-harm rate ($5.67$ versus $6.49\times10^{-3}$).

\paragraph{Per-event outcomes.}
At $f=0.5$, the matched responses reproduce mean outcomes but not single events.
Over the 30 endpoint-background maps, the per-event loss changes of the two arms correlate at $r=0.51$ with $F_{50}$ and $0.53$ with $F_1$, and their mean absolute difference is $1.31$ and $1.14$ times the mean absolute change; 12\% ($F_{50}$) and 37\% ($F_1$) of events are harmful under one arm and helpful under the other, and with $F_{50}$ a further 31\% are changed by one arm only.
The same arm under $F_{50}$ and $F_1$ correlates at $r=0.60$ (unpenalized) and $0.71$ (penalized).
The worst 1\% of events change the loss by $+0.086$ and $+0.089$ nat on average with $F_{50}$ and $+0.040$ and $+0.051$ nat with $F_1$ (unpenalized, penalized); the partial-state background shows the same.

\begin{table*}[t]
\centering
\small
\setlength{\tabcolsep}{3pt}
\resizebox{\textwidth}{!}{%
\begin{tabular}{@{}lcccccc@{}}
\toprule
 & \multicolumn{2}{c}{Frozen, endpoint} & \multicolumn{2}{c}{Frozen, partial-state} & \multicolumn{2}{c}{Jointly trained, temperature} \\
\cmidrule(lr){2-3}\cmidrule(lr){4-5}\cmidrule(l){6-7}
Point & $F_{50}$ & $F_1$ & $F_{50}$ & $F_1$ & $F_{50}$ & $F_1$ \\
\midrule
$\lambda=0$ & $-9.97$ [$-15.6$, $-9.35$] & $-9.06$ [$-12.2$, $-5.83$] & $-12.6$ [$-18.0$, $-11.9$] & $-13.8$ [$-17.8$, $-10.1$] & $-2.69$ [$-8.98$, $-4.29$] & $+4.30$ [$+2.58$, $+5.71$] \\
$f=1$ & $-1.71$ [$-3.53$, $-1.25$] & $+0.15$ [$-0.92$, $+1.13$] & $-2.42$ [$-3.68$, $-1.92$] & $+1.78$ [$+1.17$, $+2.19$] & $-0.47$ [$-2.52$, $-0.52$] & $+2.75$ [$+1.65$, $+3.63$] \\
$f=0.75$ & $-1.19$ [$-2.72$, $-0.88$] & $+1.03$ [$+0.31$, $+1.62$] & $-1.92$ [$-3.01$, $-1.57$] & $+2.01$ [$+1.63$, $+2.35$] & $-0.36$ [$-1.75$, $-0.26$] & $+1.92$ [$+1.06$, $+2.71$] \\
$f=0.5$ & $-0.68$ [$-1.82$, $-0.46$] & $+1.27$ [$+0.90$, $+1.60$] & $-1.28$ [$-2.19$, $-1.07$] & $+1.71$ [$+1.35$, $+2.01$] & $-0.24$ [$-1.07$, $-0.08$] & $+0.89$ [$+0.45$, $+1.41$] \\
$f=0.25$ & $-0.31$ [$-0.87$, $-0.10$] & $+0.59$ [$+0.45$, $+0.76$] & $-0.58$ [$-1.08$, $-0.36$] & $+0.70$ [$+0.53$, $+0.91$] & $-0.14$ [$-0.45$, $+0.04$] & $+0.14$ [$+0.06$, $+0.27$] \\
$f=0.1$ & $-0.16$ [$-0.33$, $+0.09$] & $+0.06$ [$+0.04$, $+0.10$] & $-0.22$ [$-0.39$, $+0.01$] & $+0.11$ [$+0.07$, $+0.16$] & $-0.01$ [$-0.15$, $+0.09$] & $+0.002$ [$+0.001$, $+0.007$] \\
\bottomrule
\end{tabular}}
\caption{Difference in the large-harm rate (fraction of events with harm above $0.05$ nat), penalized minus unpenalized, $10^{-3}$, 95\% joint intervals.
Every draw refits the amplitudes and the rate responds discontinuously to them, so some point estimates lie outside their percentile intervals.}
\label{tab:supp_tail}
\end{table*}

\paragraph{Jointly trained models, three seeds.}
We retrain the three regimes of 10 concept sets, the lowest-numbered set of each incompleteness stratum, with two further training seeds.
With the original models this gives 60 new and 30 reused fits; the same 512 orders and a 1000-draw image bootstrap conditional on the trained models are used.
For every seed, the maximum sampled partial harm is lower and the partially corrected NLL higher in all 10 sets (\cref{tab:supp_seeds}); for raw models, the partial-NLL difference averaged over the 30 seed--set pairs is $+12.0$ [$+8.9$, $+15.3$]$\times10^{-3}$ nat (over all 80 sets: $+10.9$, \cref{tab:native}).
This confirms the sign of the effect across seeds for these 10 sets.

\begin{table}[t]
\centering
\small
\setlength{\tabcolsep}{3pt}
\begin{tabular}{@{}lccc@{}}
\toprule
Seed & Harm lower & NLL higher & $\Delta$ partial NLL \\
\midrule
Original & 10/10 & 10/10 & $+11.75$ [$+8.77$, $+14.79$] \\
Second & 10/10 & 10/10 & $+12.74$ [$+9.25$, $+16.19$] \\
Third & 10/10 & 10/10 & $+11.51$ [$+8.27$, $+15.19$] \\
\midrule
All & 30/30 & 30/30 & $+12.00$ [$+8.91$, $+15.26$] \\
All$^{a}$ & & & $+12.00$ [$+8.49$, $+15.57$] \\
All$^{b}$ & & & $+12.00$ [$+8.09$, $+16.51$] \\
\bottomrule
\end{tabular}
\caption{Three-seed confirmation on 10 concept sets, raw jointly trained models, endpoint$+$mono minus endpoint; original seed (models reused) and two further seeds (models retrained).
Harm lower: sets with lower maximum sampled partial harm; NLL higher: sets with higher partially corrected NLL.
$\Delta$ partial NLL in $10^{-3}$ nat: mean over the 10 sets (All: over the 30 seed--set pairs), with the 95\% percentile interval of a 1000-draw image bootstrap conditional on the trained models; 2000 draws that also resample $^{a}$the three seeds within each set and $^{b}$the 10 sets.}
\label{tab:supp_seeds}
\end{table}

\paragraph{Jointly trained models, all budgets.}
With $F_1$, 14\% of the set-draw pairs have a zero source budget because the endpoint model's source-optimal global scale is zero; these pairs are kept, with no correction response in either arm.
The two arms sit on opposite sides of their source optimum: the optimal global scale has median $0.36$ for the endpoint models and $3.6$ for the penalized models.

\Cref{tab:supp_native} lists every prespecified point of the matched-source-harm comparison for the temperature-controlled models, and \cref{tab:supp_native_raw} for the raw models.
For the temperature-controlled models, the pre-intervention component is the same at every point, $\Delta\mathrm{NLL}_{\mathrm{before}}=+49.2$ [$+35.8$, $+62.9$] ($10^{-4}$ nat), positive in all 80 sets.
The NLL difference is positive in all 80 sets at every point; at the matched budgets, $\Delta\Hm$ is positive in 61--65 sets with $F_{50}$ and in 42--43 sets with $F_1$, and $\Delta\Gm$ in 71--75 and 72 sets.
At $\lambda=0$, the per-concept family gives the penalized arm slightly less harm and gain, whereas the global scale gives it more of both: the source-optimal global scale enlarges the penalized response (median $3.6$) and shrinks the endpoint response (median $0.36$).

Raw models show the same risk improvement at every budget of both families, with no Pareto advantage for either arm.
Their pre-intervention component is more than twice as large, $+110.6$ [$+82.3$, $+140.9$] ($10^{-4}$ nat), again positive in all 80 sets, while the matched response differences stay close to those of the temperature-controlled models: at $F_{50}$, $f=0.5$, the response part $\Delta\Hm-\Delta\Gm$ is $-0.19\times10^{-4}$ (temperature-controlled: $-0.25\times10^{-4}$).
At the matched budgets, $\Delta\Hm$ is positive in 60--66 sets with $F_{50}$ and in 34--36 sets with $F_1$, and $\Delta\Gm$ in 65--73 and 69--71 sets; with $F_1$, 7 of the 10 strata have a positive $\Delta\Hm$, whereas the other 3 have $\Delta\Hm\le0$ and $\Delta\Gm>0$.
These stratum point estimates describe heterogeneity; the pooled intervals do not meet the joint harm--gain criterion.
The NLL difference remains positive in every stratum and set, reflecting the pre-intervention component.
The $\lambda=0$ pattern is the same, with source-optimal global scales of median $3.4$ (penalized) and $0.36$ (endpoint).
The large-harm rate reverses with the family as before: at $f=0.5$ the difference is $-0.15$ [$-1.20$, $-0.02$] with $F_{50}$ and $+1.10$ [$+0.56$, $+1.74$] with $F_1$ ($10^{-3}$).

\begin{table}[t]
\centering
\small
\setlength{\tabcolsep}{2.5pt}
\resizebox{\columnwidth}{!}{%
\begin{tabular}{@{}lcccc@{}}
\toprule
Point & $\Delta\Hm$ & $\Delta\Gm$ & $\Delta$NLL & Strata \\
\midrule
\multicolumn{5}{@{}l}{\emph{$F_{50}$}} \\
$\lambda=0$ & $-3.69$ [$-11.8$, $-4.53$] & $-3.49$ [$-11.0$, $-4.52$] & $+49.0$ [$+35.5$, $+62.5$] & 0/0/10 \\
$f=1$ & $+0.43$ [$-0.92$, $+2.16$] & $+0.57$ [$-0.97$, $+1.96$] & $+49.1$ [$+36.0$, $+63.0$] & 10/10/10 \\
$f=0.75$ & $+0.37$ [$-0.62$, $+1.68$] & $+0.59$ [$-0.50$, $+1.72$] & $+49.0$ [$+35.9$, $+62.8$] & 10/10/10 \\
$f=0.5$ & $+0.28$ [$-0.34$, $+1.18$] & $+0.54$ [$-0.14$, $+1.35$] & $+49.0$ [$+35.7$, $+62.8$] & 10/10/10 \\
$f=0.25$ & $+0.17$ [$-0.11$, $+0.66$] & $+0.37$ [$+0.06$, $+0.85$] & $+49.0$ [$+35.7$, $+62.8$] & 10/10/10 \\
$f=0.1$ & $+0.08$ [$-0.02$, $+0.29$] & $+0.20$ [$+0.09$, $+0.43$] & $+49.1$ [$+35.7$, $+62.8$] & 10/10/10 \\
\midrule
\multicolumn{5}{@{}l}{\emph{$F_1$}} \\
$\lambda=0$ & $+5.43$ [$+1.06$, $+9.29$] & $+6.20$ [$+1.74$, $+10.0$] & $+48.5$ [$+35.2$, $+62.2$] & 9/10/10 \\
$f=1$ & $+0.06$ [$-0.72$, $+0.87$] & $+0.66$ [$-0.16$, $+1.47$] & $+48.6$ [$+35.4$, $+62.3$] & 8/10/10 \\
$f=0.75$ & $+0.05$ [$-0.54$, $+0.65$] & $+0.58$ [$-0.03$, $+1.22$] & $+48.7$ [$+35.4$, $+62.3$] & 8/10/10 \\
$f=0.5$ & $+0.03$ [$-0.36$, $+0.43$] & $+0.45$ [$+0.04$, $+0.90$] & $+48.8$ [$+35.5$, $+62.5$] & 8/10/10 \\
$f=0.25$ & $+0.01$ [$-0.18$, $+0.22$] & $+0.26$ [$+0.04$, $+0.48$] & $+49.0$ [$+35.6$, $+62.6$] & 8/10/10 \\
$f=0.1$ & $+0.01$ [$-0.07$, $+0.09$] & $+0.11$ [$+0.02$, $+0.20$] & $+49.1$ [$+35.7$, $+62.8$] & 8/10/10 \\
\bottomrule
\end{tabular}}
\caption{Jointly trained CUB hybrids, temperature-controlled, endpoint$+$monotonic minus endpoint, each with the post-hoc family, 80 concept sets, $10^{-4}$ nat, 95\% joint source/evaluation intervals (1000 draws).
Strata: number of the 10 incompleteness strata with a positive point estimate of $\Delta\Hm$/$\Delta\Gm$/$\Delta$NLL.
At every budget the endpoint model shows a risk improvement at approximately matched evaluation harm.
With $F_{50}$ at $\lambda=0$, the point estimates of $\Delta\Hm$ and $\Delta\Gm$ lie outside their percentile intervals because every draw refits the amplitudes.}
\label{tab:supp_native}
\end{table}

\begin{table}[t]
\centering
\small
\setlength{\tabcolsep}{2.5pt}
\resizebox{\columnwidth}{!}{%
\begin{tabular}{@{}lcccc@{}}
\toprule
Point & $\Delta\Hm$ & $\Delta\Gm$ & $\Delta$NLL & Strata \\
\midrule
\multicolumn{5}{@{}l}{\emph{$F_{50}$}} \\
$\lambda=0$ & $-3.43$ [$-11.7$, $-3.89$] & $-3.33$ [$-11.2$, $-4.11$] & $+110.5$ [$+82.3$, $+140.6$] & 0/0/10 \\
$f=1$ & $+0.47$ [$-1.05$, $+2.52$] & $+0.50$ [$-1.27$, $+2.12$] & $+110.6$ [$+82.6$, $+141.0$] & 10/10/10 \\
$f=0.75$ & $+0.41$ [$-0.71$, $+2.01$] & $+0.55$ [$-0.70$, $+1.83$] & $+110.5$ [$+82.4$, $+140.8$] & 10/10/10 \\
$f=0.5$ & $+0.32$ [$-0.40$, $+1.43$] & $+0.51$ [$-0.27$, $+1.44$] & $+110.4$ [$+82.3$, $+140.7$] & 10/10/10 \\
$f=0.25$ & $+0.20$ [$-0.13$, $+0.78$] & $+0.36$ [$+0.003$, $+0.90$] & $+110.5$ [$+82.2$, $+140.7$] & 10/10/10 \\
$f=0.1$ & $+0.10$ [$-0.02$, $+0.35$] & $+0.20$ [$+0.07$, $+0.45$] & $+110.5$ [$+82.3$, $+140.8$] & 10/10/10 \\
\midrule
\multicolumn{5}{@{}l}{\emph{$F_1$}} \\
$\lambda=0$ & $+5.18$ [$+0.44$, $+9.16$] & $+5.91$ [$+1.16$, $+10.1$] & $+109.9$ [$+81.7$, $+140.2$] & 8/10/10 \\
$f=1$ & $+0.02$ [$-0.85$, $+0.89$] & $+0.59$ [$-0.28$, $+1.49$] & $+110.1$ [$+81.9$, $+140.3$] & 7/10/10 \\
$f=0.75$ & $+0.02$ [$-0.64$, $+0.67$] & $+0.53$ [$-0.13$, $+1.23$] & $+110.1$ [$+82.0$, $+140.4$] & 7/10/10 \\
$f=0.5$ & $+0.01$ [$-0.43$, $+0.45$] & $+0.41$ [$-0.03$, $+0.90$] & $+110.2$ [$+82.0$, $+140.6$] & 7/10/10 \\
$f=0.25$ & $+0.004$ [$-0.21$, $+0.22$] & $+0.24$ [$+0.01$, $+0.48$] & $+110.4$ [$+82.2$, $+140.7$] & 7/10/10 \\
$f=0.1$ & $+0.001$ [$-0.09$, $+0.09$] & $+0.10$ [$+0.01$, $+0.20$] & $+110.5$ [$+82.3$, $+140.8$] & 7/10/10 \\
\bottomrule
\end{tabular}}
\caption{As \cref{tab:supp_native}, raw checkpoints.
Every budget shows the same improvement as for the temperature-controlled models.}
\label{tab:supp_native_raw}
\end{table}

\section{Monotonic penalty on BUS-BRA}
\label{sec:res_bus}

\paragraph{Setup.}
Four folds partition the 732 BUS-BRA cases, one image per case; each fold uses 289 training, 147 validation, 113 calibration and 183 evaluation cases, with no official Case identifier shared across roles or with the reserved test split.
Concept embedding models with an ImageNet-initialized ResNet-18 encoder (12.5M trainable parameters) give one base model per fold and seed (4 folds, 3 seeds); from each, two branches are trained with and without a monotonic penalty ($\lambda=1$), 24 branches with best and last checkpoints, and are compared at the base model's source-calibrated starting prediction (\cref{sec:supp_public_recipe}).
Corrections are strict partial corrections on the fields that both models get wrong.
The post-hoc families are $F_1$ and $F_6$ (one coefficient per field), fitted on the calibration split; because one event may correct several fields, the six coefficients of $F_6$ are coupled (\cref{sec:supp_families}).
The primary report uses the originally selected checkpoint and weights cases equally; source-selected checkpoints and event weighting are both secondary analyses.

\paragraph{Result.}
The source-optimal global scale varies widely across the 24 branches and is usually far below one (median $0.27$, 22 below $0.5$, a 22-fold range), so any comparison in native units depends on amplitude.
\Cref{tab:bus} shows how the measured effect changes with the comparator.
On identical events with a shared concept predictor, the penalized models are more harmful in native units ($\Delta\Hm=+0.0545$); after one source-fitted global scale they show fewer large harms but lose useful gain and have higher NLL; when the source split also chooses between per-field and global scaling, no difference is detectable.
In the matched-budget comparison, which uses the originally selected checkpoints, $F_6$ is unresolved at every budget, with interval half-widths of $0.004$--$0.014$ nat (\cref{tab:supp_bus}).
With $F_1$, the unpenalized model plus a global scale has the lower NLL for $f\ge0.5$ (\eg, $+0.0033$ [$+0.0004$, $+0.0083$] at $f=0.5$), while the evaluation $\Hm$ and $\Gm$ intervals cross zero.
No matched comparison gives the penalized models lower harm together with higher useful gain; only their large-harm rates fall, under the global scale at the cost of useful gain and likelihood ($-0.83$ pp) and with $F_1$ at $f=1$ and $0.75$ ($-0.93$ [$-1.46$, $-0.08$] and $-0.61$ [$-1.31$, $-0.01$] pp).

\begin{table}[t]
\centering
\footnotesize
\setlength{\tabcolsep}{3pt}
\begin{tabular}{@{}lccc@{}}
\toprule
 & \multicolumn{3}{c}{Penalized $-$ unpenalized ($10^{-2}$ nat)} \\
\cmidrule(l){2-4}
Amplitude control & $\Delta$NLL & $\Delta\Hm$ & $\Delta\Gm$ \\
\midrule
Native units & $+5.9$ & $+5.4$ & $-0.5$ \\
 & {\scriptsize$[+1.1, +11.3]$} & {\scriptsize$[+0.3, +11.0]$} & {\scriptsize$[-1.2, +0.2]$} \\\addlinespace[1.5pt]
Global scale ($F_1$) & $+1.0$ & $+0.1$ & $-1.0$ \\
 & {\scriptsize$[+0.3, +1.5]$} & {\scriptsize$[-0.9, +0.6]$} & {\scriptsize$[-1.6, -0.4]$} \\\addlinespace[1.5pt]
Per field or global ($F_6$/$F_1$) & $+0.3$ & $+0.5$ & $+0.2$ \\
 & {\scriptsize$[-0.1, +1.4]$} & {\scriptsize$[-0.8, +1.3]$} & {\scriptsize$[-1.3, +0.5]$} \\
\bottomrule
\end{tabular}
\caption{BUS-BRA, penalized minus unpenalized, $10^{-2}$ nat, shared concept predictor, strict partial corrections on common-error support, case-weighted, source-selected checkpoints (last row: the source split also selects between $F_6$ and $F_1$); 95\% joint intervals including source fitting and checkpoint selection.
The large-harm rate changes from $+0.10$ to $-0.83$ to $+0.72$ pp across the three rows (only the middle interval excludes zero).}
\label{tab:bus}
\end{table}

\paragraph{Breast ultrasound, all budgets.}
\Cref{tab:supp_bus} lists the primary BUS-BRA reading (originally selected checkpoint, cases weighted equally) at every prespecified point.
By our decision rules, every point is unresolved or a trade-off.
With source-selected checkpoints, the $F_1$ NLL intervals exclude zero at $f=1$, $0.75$ and $0.5$ ($+9.6$ [$+3.1$, $+15.6$], $+6.6$ [$+1.7$, $+11.7$] and $+3.5$ [$+0.3$, $+7.2$], $10^{-3}$ nat); with event weighting, only at $f=1$ ($+10.6$ [$+1.2$, $+26.5$]).
All $F_6$ intervals contain zero under both secondary readings.
At $f=0.5$ the three seeds give $F_1$ NLL differences of $+0.6$, $+2.6$ and $+6.6$ ($10^{-3}$ nat), and $F_6$ differences of $-0.4$, $+0.1$ and $+3.3$.
Matching source harm does not transfer to the evaluation split: at $f=1$ with $F_1$ the source harm is matched by construction, yet the evaluation $\Delta\Hm$ is $+5.6\times10^{-3}$, because each fold-seed has only 58--86 eligible calibration cases.
\label{sec:supp_bus_fixed}Intervals that fix the post-hoc parameters can be misleadingly narrow: for the per-field control, a fixed-source NLL interval excluded zero ($[+0.0003,+0.0069]$) but crossed zero once source fitting and checkpoint selection were resampled.
Of the 2,196 evaluation cases of the 12 fold-seeds, 1,290 fall in the common-error support of strict partial corrections.
On average, each fold-seed has 461 source events, and source selection chooses the last checkpoint in 32\% (unpenalized) and 36\% (penalized) of the draws.

\begin{table}[t]
\centering
\small
\setlength{\tabcolsep}{2.5pt}
\resizebox{\columnwidth}{!}{%
\begin{tabular}{@{}lcccc@{}}
\toprule
Point & $\Delta\Hm$ & $\Delta\Gm$ & $\Delta$NLL & $\Delta$ large harm (pp) \\
\midrule
\multicolumn{5}{@{}l}{\emph{$F_6$}} \\
$\lambda=0$ & $-0.4$ [$-10.8$, $+10.4$] & $-4.3$ [$-10.3$, $+1.6$] & $+3.9$ [$-4.6$, $+13.1$] & $-0.30$ [$-1.29$, $+0.76$] \\
$f=1$ & $+2.1$ [$-8.6$, $+14.5$] & $-1.2$ [$-6.6$, $+3.2$] & $+3.3$ [$-5.0$, $+14.0$] & $-0.20$ [$-1.06$, $+0.94$] \\
$f=0.75$ & $+1.4$ [$-8.0$, $+12.2$] & $-0.6$ [$-6.1$, $+3.8$] & $+2.1$ [$-4.4$, $+11.3$] & $-0.08$ [$-0.99$, $+1.05$] \\
$f=0.5$ & $+0.8$ [$-7.1$, $+8.8$] & $-0.2$ [$-5.9$, $+3.8$] & $+1.0$ [$-4.0$, $+8.7$] & $-0.01$ [$-0.91$, $+1.33$] \\
$f=0.25$ & $+0.2$ [$-5.7$, $+7.2$] & $-0.5$ [$-5.7$, $+3.6$] & $+0.8$ [$-3.9$, $+6.8$] & $+0.24$ [$-0.81$, $+1.46$] \\
$f=0.1$ & $+0.0$ [$-5.1$, $+5.9$] & $-0.4$ [$-4.7$, $+3.0$] & $+0.4$ [$-3.2$, $+5.3$] & $-0.52$ [$-0.97$, $+1.24$] \\
\midrule
\multicolumn{5}{@{}l}{\emph{$F_1$}} \\
$\lambda=0$ & $+0.6$ [$-6.4$, $+9.9$] & $-9.6$ [$-15.1$, $-3.7$] & $+10.1$ [$+4.6$, $+19.2$] & $-1.13$ [$-1.77$, $-0.36$] \\
$f=1$ & $+5.6$ [$-0.1$, $+16.2$] & $-3.6$ [$-8.8$, $+0.7$] & $+9.2$ [$+3.8$, $+20.1$] & $-0.93$ [$-1.46$, $-0.08$] \\
$f=0.75$ & $+4.0$ [$-0.2$, $+12.0$] & $-2.3$ [$-7.3$, $+1.6$] & $+6.3$ [$+2.1$, $+14.6$] & $-0.61$ [$-1.31$, $-0.01$] \\
$f=0.5$ & $+2.4$ [$-0.2$, $+7.7$] & $-0.9$ [$-5.4$, $+2.5$] & $+3.3$ [$+0.4$, $+8.3$] & $-0.27$ [$-1.21$, $+0.09$] \\
$f=0.25$ & $+1.0$ [$-0.2$, $+3.3$] & $+0.3$ [$-2.8$, $+2.7$] & $+0.8$ [$-1.0$, $+3.6$] & $-0.41$ [$-0.94$, $+0.34$] \\
$f=0.1$ & $+0.4$ [$-0.1$, $+1.1$] & $+0.4$ [$-1.0$, $+1.9$] & $-0.1$ [$-1.0$, $+1.1$] & $+0.41$ [$-0.46$, $+0.87$] \\
\bottomrule
\end{tabular}}
\caption{BUS-BRA matched-budget comparison, penalized minus unpenalized, originally selected checkpoint, cases weighted equally, 12 fold-seeds pooled within seed and averaged over seeds; $10^{-3}$ nat except the large-harm rate (percentage points); 95\% joint source/evaluation intervals (1000 draws).}
\label{tab:supp_bus}
\end{table}

\section{Descriptive structure diagnostics}
\label{sec:supp_diag}

These diagnostics use only trained weights and source responses, involve no new fitting and come without intervals; the CUB and BUS-BRA diagnostics are post hoc.

\paragraph{CUB, frozen response maps.}
Both arms share $b$, $q$ and the concept targets, so their single-concept responses differ only in the readout columns $w_j$.
$F_{50}$ can rescale each column but cannot change its direction over classes, and only the class-centered part of a column affects the softmax.
In the endpoint background, the class-centered cosine between the penalized and the unpenalized column of the same concept has median $0.44$ over the 1,500 columns (5/25/75/95th percentiles $0.17$/$0.33$/$0.57$/$0.71$); no column exceeds $0.9$, and on average 77.5\% of the energy is orthogonal.
The penalized column is shorter (median norm ratio $0.21$, 5--95\% range $0.09$--$0.43$).
The partial-state background gives the same picture (median cosine $0.49$, 74.8\% orthogonal energy).
For reference, two unpenalized maps that differ only in the base task loss have median cosine $0.999$, 0.4\% orthogonal energy and norm ratio $0.99$.
Relative to the shared initialization $W_0$, the unpenalized map moves far in orthogonal directions (median orthogonal change $3.7\lVert W_0\rVert$, 5--95\% range $1.6$--$9.0$) and grows by 32\% along $W_0$, whereas the penalized map changes little orthogonally ($0.28\lVert W_0\rVert$, $0.15$--$0.42$) and shrinks by 17\% along $W_0$.

\paragraph{BUS-BRA.}
With the concept predictor of the unpenalized model shared, we compare the per-field responses of the two models on 5,366 source case-field pairs in the common-error support.
On pairs where both responses are nonzero, the absolute response ratio (penalized over unpenalized) has median $0.95$.
Of the variance of its log absolute ratio, 7.1\% is a constant per fold-seed, which $F_1$ can absorb, and another 5.3\% is a constant per fold-seed and field, which $F_6$ can absorb in addition; the remaining 87.6\% varies from image to image.
Within fields the standard deviation of the log absolute ratio is $0.43$--$0.90$, and the two responses have opposite signs in 3.6\% of the pairs (up to 9\% in one field).

\paragraph{Clinical CorrectiveWorst versus CorrectiveMean.}
On the calibration role of the public same-source conditions of \cref{sec:supp_clinical_frontier}, we regress the CorrectiveWorst response on the CorrectiveMean response over the 62 subsets of each model pair.
The slope averages $0.59$ for both linear readouts and $0.75$ (BUS-BRA) and $0.69$ (BUSI647) for the MLP readouts, but ranges from $0.17$ to $1.82$ across the 24 model pairs of a condition; $R^2$ averages $0.76$--$0.84$ (model range $0.29$--$0.97$), and the two responses have opposite signs in 8.4--11.5\% of the events.
For single-field responses, 2.7--14.0\% of the variance of the log absolute ratio is a constant per fold-seed, which $F_1$ can absorb, another 7.7--12.3\% is a constant per fold-seed and field, which $F_6$ can absorb in addition, and 75.6--88.5\% varies from case to case; in three of the six BUS-BRA fields the responses have opposite signs in 21--33\% of the cases.
CorrectiveWorst is therefore not a uniformly scaled CorrectiveMean, and one global scale leaves most of the difference in place.

\paragraph{Reading.}
These diagnostics do not say whether the differences in direction help or hurt: on CUB and BUS-BRA they did not translate into a frontier advantage that the matched comparisons could detect, nor in the clinical comparison at the primary start (\cref{sec:supp_clinical_frontier}).

\section{Correction value under the Bayes reference}
\label{sec:supp_theory}

These identities follow from the propriety of the logarithmic score~\cite{gneiting2007strictly} and fix what a harm measurement can and cannot imply.
Fix a correction set $S$ and its event distribution, a starting prediction $p_0(y\mid x)$ and a corrected prediction $q(y\mid x,c_S)$.
With $\ell$ the log loss and $\Delta\ell=\ell(q)-\ell(p_0)$ per event, useful gain and harm satisfy $\Gm-\Hm=\mathbb E[\log q(y\mid x,c_S)-\log p_0(y\mid x)]$.

\paragraph{The same start.}
For the Bayes corrector $q^{*}=p(y\mid x,c_S)$ at the same start,
$(\Gm^{*}-\Hm^{*})-(\Gm-\Hm)=\mathbb E\,\mathrm{KL}\big(p(\cdot\mid x,c_S)\,\big\|\,q(\cdot\mid x,c_S)\big)\ge0$.
Any corrector therefore falls short of the reference by its expected divergence in the corrected state, and a harm below $\Hm^{*}$ comes with a useful gain that is lower than $\Gm^{*}$ by at least the difference in harm.
The statement compares a corrector with the reference at a fixed start; it does not imply that one trained model must lose gain relative to another.

\paragraph{The Bayes start.}
If the start is $p_0=p(y\mid x)$, then $\Gm^{*}-\Hm^{*}=I(Y;C_S\mid X)\ge0$, so the Bayes corrector never raises the population risk.
Its per-event harm is positive exactly when $p(y\mid x,c_S)<p(y\mid x)$ with positive probability; uncertainty about the concepts alone does not imply it.
For example, if $X$ is constant, $Y$ is a fair bit and $C=(Y,N)$ with $N$ independent noise, then $p(y\mid c)=1\ge p(y)=1/2$ for every event and $\Hm^{*}=0$, although $C$ is not determined by $X$.

\paragraph{Real models.}
For a trained model, $\Gm-\Hm=I(Y;C_S\mid X)+K_0-K_1$, with $K_0=\mathbb E\,\mathrm{KL}(p(\cdot\mid x)\,\|\,p_0(\cdot\mid x))$ the error of the start and $K_1=\mathbb E\,\mathrm{KL}(p(\cdot\mid x,c_S)\,\|\,q(\cdot\mid x,c_S))$ the error after correction.
A worse start raises $K_0$ and with it the measured net value of corrections without making the corrected predictions better, which is the effect that the before/after split of \cref{eq:split} isolates.

\section{Numerical verification}
\label{sec:supp_verify}

\paragraph{Solvers and certificates.}
The fused GPU kernels for source sums and evaluation metrics agree with a direct reference implementation to a relative $1.3\times10^{-15}$ on the frozen maps and $2.5\times10^{-13}$ on the jointly trained models, and response maps rebuilt from the stored predictions reproduce the stored source NLL of all 160 jointly trained models to $3.3\times10^{-16}$.
Every grid solve is certified.
On the frozen maps, 1.92 million solves have a largest nonsmooth KKT violation of $1.0\times10^{-10}$, and 1,629 minimizers at kinks are bracketed to $1.3\times10^{-13}$; on BUS-BRA, all 768,000 solves have minimum-norm subgradients of at most $1.0\times10^{-10}$; on the jointly trained models, all solves converge to $1.0\times10^{-10}$, with kink certificates no wider than $1.15\times10^{-13}$.
The identities $\Delta\mathrm{NLL}=\Delta\Hm-\Delta\Gm$ at a fixed start and $\Delta\mathrm{NLL}=\Delta\mathrm{NLL}_{\mathrm{before}}+\Delta\Hm-\Delta\Gm$ for jointly trained models hold draw by draw to $4.5\times10^{-16}$.

\paragraph{Replays and independent checks.}
At $\lambda=0$, every family reproduces the fits of the earlier analyses: $F_{50}$ coefficients to $5.7\times10^{-8}$, $F_1$ to rounding error apart from the stopping rule of \cref{sec:supp_plans}, and evaluation NLL, $\Hm$ and $\Gm$ to $8.0\times10^{-9}$; the 512-order path audit reproduces the stored path metrics of all 80 sets to $5.1\times10^{-15}$.
A separate CPU implementation without the GPU kernels re-solves sampled coordinate problems with bounded Brent on the frozen and jointly trained maps, and checks 288 BUS-BRA $F_6$ solutions against SLSQP on an exact smooth reformulation.
Our objective is never higher by more than $5.6\times10^{-16}$, coefficients agree to the precision of the reference solvers, budgets are matched within tolerance, and evaluation metrics agree to $1.4\times10^{-15}$.

\section{Earlier joint ultrasound comparison and reserved cohort}
\label{sec:supp_sealed}

\paragraph{Datasets.}
A further comparison trains jointly on Private250, 250 images from an institutional collection, and BUSI647, the 647 benign and malignant BUSI images (437 and 210), with a selection split and seed cohort different from \cref{sec:supp_clinical_sources}.
The main BUS-BRA and CUB experiments use their own models (\cref{sec:supp_public_recipe,sec:supp_compute}).

\begin{table*}[t]
\centering
\small
\setlength{\tabcolsep}{4pt}
\begin{tabular}{@{}llcccc@{}}
\toprule
Start & Image set & $\Delta$NLL before & $\Delta\Hm$ & $\Delta\Gm$ & $\Delta$ partial NLL \\
\midrule
Native & Pooled (130) & $-1.54$ [$-11.75$, $+10.60$] & $-2.12$ [$-3.13$, $-1.19$] & $-7.93$ [$-10.35$, $-5.92$] & $+4.27$ [$-5.32$, $+15.68$] \\
 & Private250 (40) & $-8.73$ [$-20.08$, $+0.77$] & $-2.35$ [$-2.97$, $-1.78$] & $-12.85$ [$-16.97$, $-9.86$] & $+1.77$ [$-6.36$, $+9.23$] \\
 & BUSI647 (90) & $+1.65$ [$-11.45$, $+16.83$] & $-2.02$ [$-3.50$, $-0.84$] & $-5.75$ [$-9.07$, $-3.39$] & $+5.38$ [$-6.93$, $+20.02$] \\
\midrule
Fixed & Pooled (130) & $0$ & $-1.28$ [$-2.19$, $-0.50$] & $-4.77$ [$-6.68$, $-3.21$] & $+3.49$ [$+1.60$, $+5.62$] \\
 & Private250 (40) & $0$ & $-1.09$ [$-1.65$, $-0.62$] & $-4.46$ [$-8.19$, $-1.99$] & $+3.37$ [$+0.66$, $+7.20$] \\
 & BUSI647 (90) & $0$ & $-1.36$ [$-2.74$, $-0.21$] & $-4.91$ [$-7.26$, $-3.18$] & $+3.55$ [$+1.27$, $+6.26$] \\
\bottomrule
\end{tabular}
\caption{Supplementary breast ultrasound comparison: CorrectiveWorst minus CorrectiveMean, 9 models per objective, all 62 nonempty proper subsets of the six fields as correction masks, raw probabilities, $10^{-2}$ nat.
The pooled analysis combines 40 Private250 images and 90 images from the concept-annotated BUSI647 subset.
Fixed: both arms on the CorrectiveMean starting logits, each with its own concept forecasts and correction response.
Brackets: 95\% intervals of a 1000-draw image bootstrap conditional on the trained models, for each image set separately.
}
\label{tab:supp_private}
\end{table*}

\paragraph{Supplementary ultrasound comparison.}
The sample counts refer to a fixed internal selection split, not to the sizes of the image collections.
Private250 has 250 images in 249 conservative image-derived groups: 199 groups formed the training pool and 50 were reserved for the earlier internal validation; the training pool was split into 159 fitting and 40 selection groups.
The existing BUSI647 training pool has 452 images, split into 362 fitting and 90 selection groups; the original remaining splits contain 98 validation and 97 test images and are not the 90-image set reported here.

One comparison from the internal study bears directly on the harm--gain question of the introduction.
It uses the CorrectiveMean and CorrectiveWorst objectives of \cref{sec:setup}.
Each objective was trained with 3 seeds and 3 folds, and the 9 models per objective were read out on the same 130 development images (40 from Private250 and 90 from BUSI647) with raw probabilities and no post-hoc temperature or amplitude fitting.
The readout uses all 62 nonempty proper subsets of the six fields as correction masks, the same masks for both models of a pair, and weights every image and model equally.
\Cref{tab:supp_private} reports CorrectiveWorst minus CorrectiveMean natively and with both arms placed on the CorrectiveMean starting logits; each arm keeps its own concept forecasts and adds its own correction response, so the fixed-start rows substitute one component.
With the start fixed, the worst-case objective lowers harm by $0.013$ nat but useful gain by $0.048$ nat, more than three times as much, so the partially corrected NLL rises by $0.035$ [$0.016$, $0.056$].
On Private250, the fixed-start corrected NLL rises by $0.034$ [$0.007$, $0.072$] nat, following the pooled direction.
At native starts, the pooled and Private250 corrected-NLL point estimates are also higher, but both intervals include zero.

\paragraph{Reserved case-disjoint cohort.}
A held-out split of 281 images of 146 cases shares no official Case identifier with the 732-case public pool or the 915-case annotation package.
On it we test a single contrast: the calibrated comparison of \cref{sec:supp_clinical_frontier} for the BUS-BRA linear readout at the shared CorrectiveMean start and $f=0.5$, CorrectiveWorst minus CorrectiveMean in corrected NLL and useful gain.
The 48 selected development checkpoints and the source fits on the development calibration roles are used unchanged, and the concept annotations are encoded as in \cref{sec:supp_compute}.
Each case has equal weight, with its images averaged, and the 24 model pairs are averaged.
The intervals combine 1000 case-bootstrap draws of the cohort with the paired source-refit draws and condition on the trained models.
We call the contrast in favor of CorrectiveMean if the NLL interval lies above zero and the useful-gain interval below zero, in favor of CorrectiveWorst in the converse case, and unresolved otherwise.
The result is unresolved: $\Delta$NLL $=+8.3$ [$-1.4$, $+20.6$], $\Delta\Gm=-8.4$ [$-22.5$, $+1.7$] and $\Delta\Hm=-0.1$ [$-6.7$, $+4.2$] ($10^{-3}$ nat), and $\Delta$NLL is positive in four of six seeds.
The estimates have the signs of the development result at $f=0.5$ (\cref{tab:supp_clinical_frontier}), but every interval includes zero.

\section{Code and compute}
\label{sec:supp_compute}

Training, GPU inference and the CUB post-hoc fits used two A800 80\,GB GPUs; BUS-BRA post-hoc fits, independent verification and aggregation used CPU workers.
\paragraph{\arxivor{Released annotations and tools}{Annotations and tools to be released}.}
We \arxivor{release}{will release}, for the first time, the final, adjudicated expert concept annotations of the 647 BUSI647 images, the subset of BUSI~\cite{aldhabyani2020dataset} without its normal class (437 benign, 210 malignant), and of all 1591 images of 915 public BUS-BRA cases.
Each set \arxivor{comes}{will come} with code that encodes the six fields into the 24 concept indicators, a manifest that maps every annotation to its image in the original release, and the correction-audit code of this paper; the images are obtained from the original releases.
The experiments use 452 of the 647 BUSI647 images, the development pool of \cref{sec:supp_clinical_sources}, and one image of each of 732 of the 915 BUS-BRA cases, the public pool of \cref{sec:supp_public_recipe}; we \arxivor{release}{will release} no annotations of the reserved cohort of \cref{sec:supp_sealed}.

\paragraph{Code.}
The \arxivor{supplementary code}{code to be released} contains four standalone implementations built on the official public releases.
The clinical implementation retrains conditions I, B and IB of \cref{sec:supp_clinical_sources} (144 fits) and reproduces the reported fits exactly.
The selection implementation trains the MLP readouts of \cref{sec:supp_selection} and reproduces the BUSI647 and BUS-BRA values of \cref{tab:selection,tab:supp_selection_full,tab:supp_selection_controls,tab:supp_selection_mlp}.
The BUS-BRA implementation is the one described in \cref{sec:supp_public_recipe}, together with the BUS-BRA structure diagnostic of \cref{sec:supp_diag}.
The CUB implementation retrains the 240 jointly trained hybrids from the official dataset and CLIP weights and reruns the path audit, the seed confirmation, the frozen maps, the matched frontiers, the tail comparisons and the CUB structure diagnostic, including the contrasts of \cref{fig:contrasts,tab:supp_bus}, and reproduces the reported values and decisions.
The BUSI647 and BUS-BRA concept annotations \arxivor{are}{will be} released under CC BY 4.0; images and pretrained weights are obtained from their original releases.
No Private250 images, annotations or models \arxivor{are}{will be} released (\cref{sec:supp_selection_p,sec:supp_sealed}).

\end{document}